\documentclass[12pt]{article}

\usepackage[preprint]{colm/colm2026_conference}
\usepackage{microtype}
\usepackage{hyperref}
\usepackage{url}
\usepackage{booktabs}
\usepackage{graphicx}
\usepackage{xcolor}
\usepackage{multirow}
\usepackage{subcaption}
\usepackage{caption}
\usepackage{lineno}
\usepackage{soul}
\usepackage{paralist}
\usepackage{longtable}   
\usepackage{array}  
\usepackage{cleveref}
\usepackage[breakable]{tcolorbox}
\usepackage{alltt}  % verbatim-ish env that survives inside tcolorbox — used for prompt bodies
\usepackage{listingsutf8}
\usepackage[T1]{fontenc}

\usepackage{tabularx, rotating, colortbl, amssymb, pifont, textcomp}

\usepackage{setspace}

\definecolor{darkblue}{rgb}{0, 0, 0.5}
\hypersetup{colorlinks=true, citecolor=darkblue, linkcolor=darkblue, urlcolor=darkblue}

\usepackage{tikz}
\usetikzlibrary{positioning, arrows.meta, fit, backgrounds, calc, shapes.misc}

  \newcommand{\cmark}{\textcolor{green!60!black}{\ding{51}}}
  \newcommand{\xmark}{\textcolor{red}{\ding{55}}}
  \newcommand{\omark}{\textcolor{orange!80!black}{$\circ$}}
  \newcommand{\stagePC}{%
    \tikz[baseline=-0.6ex]\node[fill=blue!12,rounded corners=2pt,
      inner xsep=3pt,inner ysep=1.5pt,font=\scriptsize\sffamily]{%
      Code\,$\neq$\,Paper};}
  
  \newcommand{\stageSA}{%
    \tikz[baseline=-0.6ex]\node[fill=green!15,rounded corners=2pt,
      inner xsep=3pt,inner ysep=1.5pt,font=\scriptsize\sffamily]{%
      Sum.\,$\neq$\,Agent};}

\definecolor{clrStepFill}  {HTML}{D6E4F7}
\definecolor{clrStepBd}    {HTML}{2C6FAC}
\definecolor{clrStepTxt}   {HTML}{1A4A7A}
\definecolor{clrDataFill}  {HTML}{FFF3CD}
\definecolor{clrDataBd}    {HTML}{B08800}
\definecolor{clrInputFill} {HTML}{D9F0DD}
\definecolor{clrInputBd}   {HTML}{3A7D44}
\definecolor{clrGroupBg}   {HTML}{EBF3FB}

\tikzset{
  pplStep/.style={
    rectangle, rounded corners=4pt,
    draw=clrStepBd, fill=clrStepFill, line width=1.1pt,
    text width=2.6cm, align=center,
    font=\small\bfseries\color{clrStepTxt},
    inner sep=7pt, minimum height=1.0cm
  },
  pplStepSm/.style={
    rectangle, rounded corners=4pt,
    draw=clrStepBd, fill=clrStepFill, line width=1.0pt,
    text width=1.4cm, align=center,
    font=\scriptsize\bfseries\color{clrStepTxt},
    inner sep=4pt, minimum height=0.7cm
  },
  pplStep4/.style={
    rectangle, rounded corners=4pt,
    draw=clrStepBd, fill=clrStepFill, line width=1.1pt,
    text width=2.6cm, align=center,
    font=\small\bfseries\color{clrStepTxt},
    inner sep=7pt, minimum height=1.0cm
  },
  pplData/.style={
    rectangle, rounded corners=12pt,
    draw=clrDataBd, fill=clrDataFill, line width=0.9pt,
    text width=2.4cm, align=center,
    font=\scriptsize, inner sep=5pt, minimum height=0.72cm
  },
  pplInput/.style={
    rectangle, rounded corners=12pt,
    draw=clrInputBd, fill=clrInputFill, line width=0.9pt,
    text width=2.0cm, align=center,
    font=\scriptsize, inner sep=5pt, minimum height=0.72cm
  },
  pplShadow/.style={
    rectangle, rounded corners=12pt,
    draw=clrDataBd, fill=clrDataFill, line width=0.6pt,
    text width=2.4cm, minimum height=0.72cm
  },
  arr/.style={
    -{Stealth[length=5pt,width=4pt]},
    line width=0.85pt, color=gray!70!black
  },
  grpbox/.style={
    draw=clrStepBd, fill=clrGroupBg,
    rounded corners=8pt, line width=0.9pt,
  },
}

\newtcolorbox[list inside=prompt,auto counter,number within=section]{prompt}[1][]{
    breakable,
    colbacktitle=black!60,
    fonttitle=\small,
    coltitle=white,
    fontupper=\footnotesize,
    boxsep=3pt,
    left=0pt,
    right=0pt,
    top=0pt,
    bottom=0pt,
    boxrule=1pt,
    #1,
}

\title{Read the Paper, Write the Code:\\Agentic Reproduction of Social-Science Results}

\author{Benjamin Kohler \\
ETH Zurich
\And
David Zollikofer \\
ETH Zurich
\And
Johanna Einsiedler \\
University of Basel
\AND
Alexander Hoyle* \\
TU Wien \\ Complexity Science Hub
\And
Elliott Ash\thanks{Equal supervision. Author contact: Kohler, \url{benjamin.kohler@gess.ethz.ch}; Zollikofer, \url{david.zollikofer@gess.ethz.ch}; Einsiedler, \url{johanna.einsiedler@unibas.ch}; Hoyle, \url{alexander.hoyle@tuwien.ac.at}; Ash, \url{ashe@ethz.ch}. Code at \href{https://github.com/benjamin-kohler/social_science_replicability}
     {github.com/benjamin-kohler/social\_science\_replicability}.} 
\\
ETH Zurich
\And
{\color{white} blank} \\
{\color{white} blank}
}
\begin{document}

% \ifcolmsubmission
% \linenumbers
% \fi

\maketitle

\begin{abstract}

\noindent Recent work has used LLM agents to reproduce empirical social science results with access to both the data and code.  We broaden this scope by asking: Can they reproduce results given only a paper's methods description and original data? We develop an agentic reproduction system that extracts structured methods descriptions from papers, runs reimplementations under strict information isolation---agents never see the original code, results, or paper---and enables deterministic, cell-level comparison of reproduced outputs to the original results. An LLM-based error attribution step traces discrepancies through the system chain to identify root causes. Evaluating four agent scaffolds and four LLMs on 48 papers with human-verified reproducibility, we find that agents can largely recover published results, but performance varies substantially between models, scaffolds, and papers. Root cause analysis suggests that failures stem both from agent errors and from underspecification in the papers themselves.

\end{abstract}

\setstretch{1.1}

\section{Introduction}
\label{sec:introduction}

Agentic systems are increasingly able to autonomously generate, debug, and execute end-to-end research pipelines across a broad range of domains \citep{lu2024aiscientist,si2025can,10.1007/978-981-96-8912-5_1, schmidgall-etal-2025-agent,Sun16102025}.
These capabilities have led to the application of agents to the task of \emph{scientific reproduction}, % (overview in \Cref{tab:prior_work_typology}), 
as well as to novel considerations of replication-oriented publication pipelines \citep{brodeur2025replicationengine,fishman2025editorial}. In the empirical social sciences specifically, recent work uses agents to execute the preexisting code associated with papers and to assess their level of reproducibility \citep{hu2025reprobench,shah2026automating,xu2026scaling}.

Yet publications, not code, communicate scientific research and serve as the source of truth. 
An open question, then, is whether reproduction of research results is contingent on direct access to the analysis code. Or whether, as presumably intended, the paper's methods description is sufficiently detailed to allow reproduction through re-implementation of the code from scratch.

This paper makes progress on this question by building an agentic system to reproduce research results given a paper's method description and associated data, but without the analysis code. Our pipeline extracts structured representations of the research methods and results, where specific numbers are masked. The agent is then tasked with reproducing the masked values by implementing the analysis pipeline from scratch, allowing for deterministic benchmarking. 

We apply the pipeline to a dataset of 48 social science papers that have been human-verified as reproducible \citep{i4replication2024}. Overall, frontier LLM agents are quite successful at re-implementing analysis code of social science papers. % (\Cref{fig:coeff-results}). 
For the three agents with the highest sign agreement, reproduced and original coefficients agree on the sign approximately 85--91\% of the time. For the best-performing agent, reproduced coefficients are within 1.96 original standard errors over 80\% of the time.

Performance varies a lot across LLMs and across agent scaffolds. The best performing agent in our comparisons is GPT-5.4 using the OpenCode scaffold, out-performing Claude Code (Opus 4.6), GPT-5.3 Codex, and GLM-5 using the OpenCode scaffold. The scaffold-model interaction matters a lot, with GPT-5.4 on OpenCode outperforming GPT-5.4 on the Codex CLI or on mini-SWE-Agent. In supporting analysis, we show that OpenCode GPT-5.4's better performance is associated with much greater token usage, coming at the cost of more expensive API calls and longer analysis runs.

Our reproduction pipeline enables a detailed diagnosis of failure modes to help identify the source of incorrect estimates. 
We find that errors mostly come from papers' underspecification of methods and---to a lesser extent---from agents' misunderstanding of or noncompliance with the methods description.

These methods and findings expand on the recent literature on automated scientific reproducibility. While important gaps remain to be closed, LLM agents can increasingly \textit{read the paper and write the code}. These systems could play an important role in addressing the well-recognized reproduction gaps in empirical social science \citep{ioannidis2005,nosek2012,brodeur2024mass}. 

\section{Background on agentic reproducibility}\label{sec:related}

Independent verification is central to the scientific method \citep{Nosek2022}. Recent crowd-sourced reproduction efforts have revealed substantial variation in reproducibility across social science disciplines. Yet such work requires considerable resources---hundreds of trained researchers to collectively assess only a negligible fraction of the published literature \citep{Brodeur2026-hk,Miske2026-dv, Tyner2026-pb}.

A helpful framework from \citet{dreber_replicability} organizes different forms and components of reproducibility and replicability. \textit{Reproducibility} tests whether the results and conclusions of original studies can be reproduced based on the \emph{same data} used in the original studies. \emph{Replicability} concerns the validity of results when similar methods are applied to \emph{new} data. This paper concerns reproducibility.

Reproducibility can be further decomposed into three types: (1) re-running the original authors' code on identical data to regenerate results, (2) reproducing results using only the information reported in the paper and the same data, and (3) testing whether results are robust to reasonable alternative analytical decisions applied to the same data. Our work falls under (2), re-implementation.  

\begin{table}

% Requires: booktabs, xcolor, colortbl, pifont, graphicx (for resizebox)
%\newcommand{\cmark}{\ding{51}}
\newcommand{\con}{\textsuperscript{\textdagger}}

%\begin{table}[t]
\centering
\footnotesize
\setlength{\tabcolsep}{3pt}
\begin{tabular}{@{}lcc c c@{}}
\toprule
  \textbf{System}
  & \textbf{Domain}
  & \textbf{Has Code}
  & \textbf{New Data}
  & \textbf{Evaluation} \\
\midrule

CORE-Bench \citep{siegel2024corebench}
  & Multi & \cmark & & Det. \\ \\

PaperBench \citep{starace2025paperbench}
  & ML & & & LLM \\

Paper2Code \citep{seo2026paper2code}
  & ML & & & LLM \\

AutoReproduce \citep{zhao2025autoreproduce}
  & ML & & & Both \\

FIRE-Bench \citep{wang2026firebench}
  & ML & & & LLM \\ \\ 

REPRO-Bench \citep{hu2025reprobench}
  & Soc.\ sci. & \cmark & & LLM \\

PaperRepro \citep{zhang2026paperrepro}
  & Soc.\ sci. & \cmark & & LLM \\

\citet{shah2026automating}
  & Soc.\ sci. & \cmark & & Det. \\

\citet{xu2026scaling}
  & Soc.\ sci. & \cmark & & Det. \\

ReplicatorBench \citep{nguyen2026replicatorbench}
  & Soc.\ sci. & \cmark & \cmark & LLM \\ \\

\textbf{This work}
  & \textbf{Soc.\ sci.} & & & \textbf{Det.} \\

\bottomrule
\end{tabular}
\caption{
  \textbf{Automated systems for scientific reproducibility}.\textit{ Columns indicate disciplinary domain (multidisciplinary, machine learning, or social science), whether the agent has access to the code, whether the agent replicates on new data, and the evaluation approach (deterministic, LLM-as-judge, or both).}
  %\\[2pt]
  %{\footnotesize \textit{Notes:} Multi = multidisciplinary; ML = machine learning; Soc.\ sci.\ = social science; Det.\ = deterministic evaluation; LLM = LLM-as-judge evaluation; Both = both deterministic and LLM-as-judge.}
}
\label{tab:prior_work_typology}

\end{table}

This paper contributes to ongoing research exploring the deployment of LLM agents for reproduction and replication, summarized in \Cref{tab:prior_work_typology}.\footnote{We exclude benchmarks of code implementation quality rather than reproducibility \citep{huang2023mlagentbench,tian2024scicode,hua2025researchcodebench}.} %The scope varies: CORE-Bench \citep{siegel2024corebench} evaluates agents' ability to re-run existing code and data in multiple domains; PaperBench \citep{starace2025paperbench} creates structured rubrics from machine learning papers and tasks agents' to recreate the code and analyses. 
Existing benchmarks in the ML domain evaluate agents' ability to reproduce code architectures and results. In social science, prior works---\citet{hu2025reprobench}, \citet{zhang2026paperrepro}, \citet{shah2026automating}, and \citet{xu2026scaling}---benchmark reproducibility while granting agents access to the original code. Our contribution differs in that we reproduce social science results from the methods and data alone, without access to any author-provided code. In \citet{nguyen2026replicatorbench}, the agents try to replicate (rather than reproduce) a paper by retrieving new data and testing the same hypothesis.\footnote{In one of Nguyen et al.'s  (\citeyear{nguyen2026replicatorbench}) specifications, the agent does not have access to the analysis code from the replicated paper, but it still has access to the complete original PDF and parsed results---the answers it aims to replicate. We remove such information in our approach because it can cause leakage---in preliminary versions of our setup where models had access to results, agents would often copy them directly.}

Unlike most prior work, which relies on LLM-as-judge evaluation, our evaluation is deterministic, comparing reproduced outputs directly to original values with adjustments for statistical significance based on ground-truth standard errors. This approach may be more reliable, since LLMs are not well-validated for evaluation tasks \citep{li2024llms}.

% The last column of \Cref{tab:prior_work_typology} indicates whether performance is evaluated deterministically based on numerical outputs or whether it is done by prompting an LLM judge. Our evaluation is deterministic, with adjustments for statistical significance of coefficients based on ground-truth standard errors. Deterministic evaluations may be more reliable than LLM-as-judge evaluations, since LLMs are not well-validated for this task \citep{li2024llms}.

\section{A pipeline for paper-derived reproduction}% (Ben) }
\label{sec:methods}

\begin{figure}%[t]
\centering
\includegraphics[width=\textwidth]{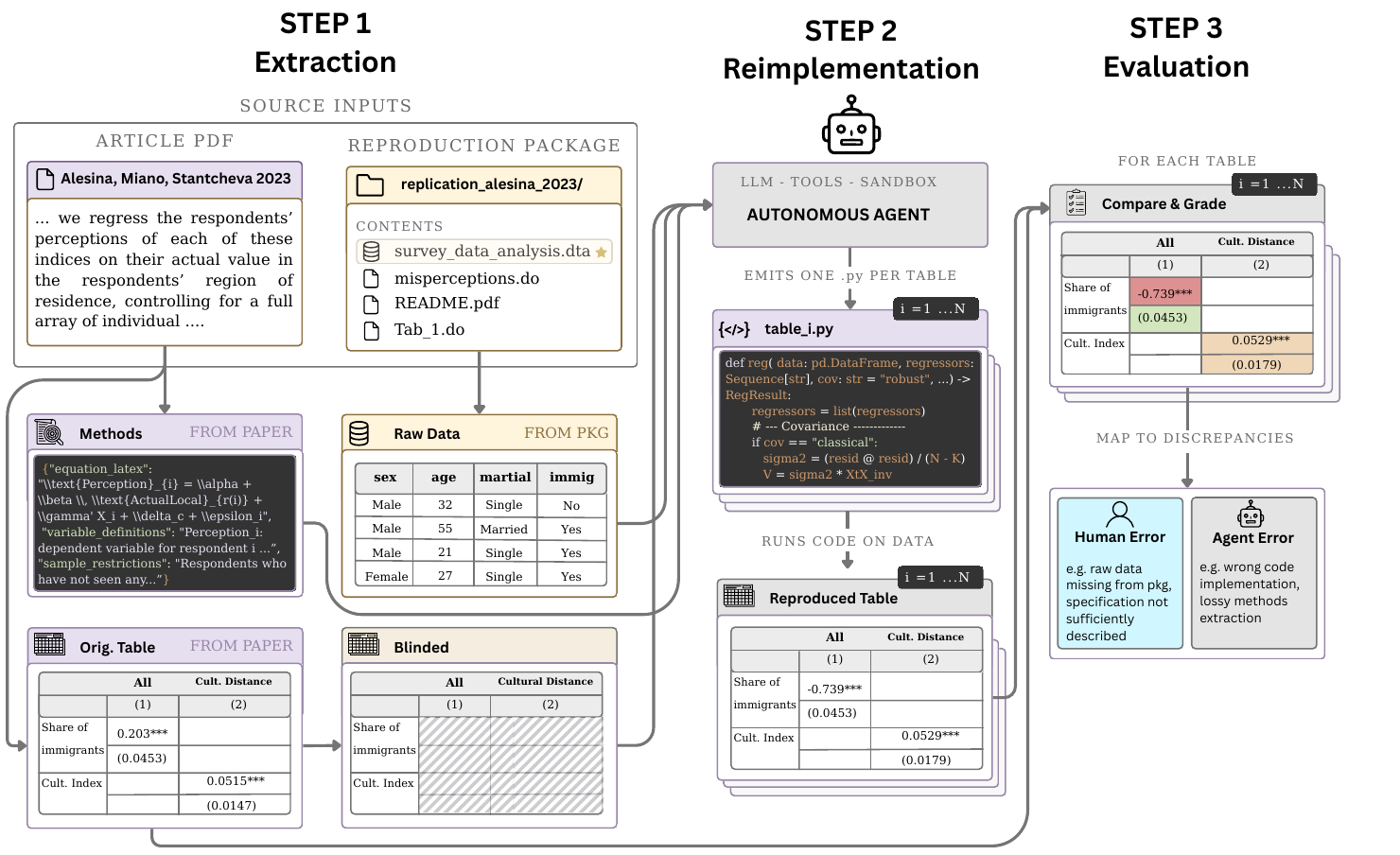}
\caption{\textbf{Overview of the pipeline for replicating empirical results. } \textit{Step 1 (Extraction): Paper and replication package are parsed to extract the methods, original data, and a blinded version of the original results table. Step 2 (Reimplementation): An autonomous LLM agent, equipped with tools and a sandboxed execution environment, emits one Python script per table and runs it on the original data to produce a reproduced table. Step 3 (Evaluation): Each reproduced table is compared cell-by-cell against the original, and divergences are traced back to human errors (e.g., missing data, underspecified methods in the original package) and agent errors (e.g., incorrect code, lossy extraction of the methods). Code at \href{https://github.com/benjamin-kohler/social_science_replicability}
     {github.com/benjamin-kohler/social\_science\_replicability}.
}}
\label{fig:pipeline}
\end{figure}

We develop a multi-step pipeline to evaluate whether LLM agents can reproduce social science results. The pipeline tasks agents with re-implementing code from methods descriptions extracted from the paper, using the associated data (see \Cref{fig:pipeline}).
Separating the pipeline into distinct steps minimizes information leakage of both existing numerical results and original code to the agents. The four steps are: (1) extracting data, methods, and results from the paper and associated files; (2) re-implementing the code based on the data, methods, and results table templates; (3) evaluating estimated results against the original results; and (4) diagnosing and explaining deviations.

\subsection{Step 1: Extraction}\label{sec:methods:extraction}
The first step extracts structured information from the paper and its reproduction package in three sub-processes (for the exact prompt see Appendix \ref{prompt:extraction}).

\paragraph{Method extraction.} We provide an LLM with the complete PDF of the paper and prompt it to extract a complete methods description, similar to the approach taken by \citet{seo2026paper2code} for machine learning papers. The methods description includes the main research question, the study setting (e.g., country and time period), data descriptions (e.g., data sources and a broad summary of each dataset), general data manipulations (e.g., filtering steps or variable construction), and detailed descriptions of each results table without the numerical contents (e.g., name, caption, row and column labels, table-specific data filters, and model specifications such as regression formulas). The model is explicitly instructed not to include any numerical results or conclusions, verified by checking for any numerals in the table description.

This step produces a structured representation of the paper's method without leaking results or code to the reimplementation agent (\Cref{sec:methods:reimplementation}).
Compared to providing the original paper directly (as is done in \citealt{nguyen2026replicatorbench}), this step helps prevent the agent from hard-coding results or re-iterating code until exact results are reached.

The quality of the methods extraction influences downstream results. An incomplete or erroneous extraction can result in a failed replication. In the error source diagnosis below, we show that incomplete extraction is occasionally responsible for failed reproductions. While we found that overall, the extracted methods descriptions were of high quality, there is likely scope for further improvements in the quality of method extraction. An alternative extraction prompt produces a small decline in paper grades on a ten-paper, two-agent subset (Appendix~\ref{sec:appendix-extractor-variant}).

\paragraph{Results extraction and blinding.} 

The next step is to extract the original results from tables.\footnote{Empirical social science results can take the form of either tables or figures. Our pipeline handles both tables and figures, but we restrict our core analysis to tables because the results can be compared deterministically. Figure-level comparisons require vision-language models whose reliability for this task remains uncertain \cite[][\emph{inter alia}]{wang2024charxiv,tang2025chartmuseum,hou2026vlm}. Supporting figure-level results are reported in Appendix~\ref{sec:appendix-figures}.} A dedicated LLM call retrieves the numerical values from all tables in the paper.\footnote{We include all tables from the paper's published journal PDF. Tables in separate online appendices are not included. In the results below, about 8\% of tables (from 18 papers) are not included in the analysis due to extraction issues.} The model returns a structured representation of the table in which cells are identified by both row and column indices as well as their label names.
For each cell, the model records the content as a string. It then decomposes it into several components: a numerical value, a classification of the metric type (e.g., coefficient or number of observations), the number of significance stars, and---for statistics such as standard errors or p-values---a reference to the corresponding coefficient.

To construct a template for the reproduction agents to fill in, we ``blind'' a copy of this structured table by setting all cell values and significance stars to null.

Table parsing uses GPT-5-mini applied to the PDF rendered as images, which we found to outperform extraction from machine-readable PDF text.
We manually validate table extraction on a random sample of five papers (24 tables in total).
The model extracted information perfectly for all portrait-oriented tables and produced occasional minor numerical errors in the second or third decimal place for landscape-oriented or multi-page tables.

\paragraph{Data extraction.} We instruct an LLM ({GPT-5-mini}) to identify the data files required for replication from the full directory tree and README of each provided reproduction package. The model is instructed to limit the datasets to the least pre-processed version available, i.e., the minimum viable set to perform reproduction. The model also classifies whether the package contains sufficient data to fully reproduce  the paper, or whether data were partially or entirely missing---for example, because they are proprietary, subject to privacy restrictions, or require independent collection. Papers with no usable data are skipped. 

\subsection{Step 2: Reimplementation}\label{sec:methods:reimplementation}

The extracted method, table templates, and original data serve as the sole inputs for the reproduction agents, following \cite{starace2025paperbench} and \cite{seo2026paper2code}.
Agents are instructed to read the extracted methods and table templates, explore the data, and write a separate Python script for each table that fills in the corresponding template (see Appendix \ref{prompt:replication} for the prompt used).

The agents are prohibited from accessing any file paths outside their workspace or retrieving the original reproduction package or paper via web tools. Each agent operates in a designated workspace containing only the task description, the extracted methods, the output templates, and a symlink to the data. The original reproduction package and paper PDF remain on the same machine but outside this boundary. These restrictions are necessary because agents could otherwise simplify the task by directly accessing the original code or results.

 We implement a two-stage audit pipeline to verify compliance with the guardrail (Appendix~\ref{app:guardrails_and_model_trace_audits}). The guardrail audit includes a deterministic regex scan, which classifies all accessed file paths and URLs as allowed or forbidden, as well as a GPT-5.4-mini review that rates each run from clean to severe violation. Suspicious runs are manually inspected and rerun if leakage is detected. In parallel, a hardcoding audit checks whether agents output statistical results as numeric literals without a computation path from the data. These two audits provide complementary signals: the guardrail audit identifies how forbidden information may have entered the pipeline, while the hardcoding audit detects outputs not derived from genuine computation. Together, they indicate that the reported results reflect genuine reimplementation rather than retrieval or memorization of pre-existing outputs.

\subsection{Step 3: Evaluation}\label{sec:methods:validation}

Since both the agent-produced results and the original results share the same structured template, we can directly compare outputs at the cell level.
For all numerical values, we check the sign agreement and compute the relative percentage difference. For regression coefficients, we additionally compute the difference scaled by the ground-truth standard error.

Each numerical value receives a letter grade based on the percent absolute deviation from the original value. If the sign matches, we assign A (0\%-2\%), B (2\%-20\%), C (20\%-40\%), D (40\%-60\%), or E (60+\%). If the sign doesn't match, we assign E. If the estimate is not produced or is not a number, we assign F (see Appendix~\ref{app:grading} for a few more details on grades).

The letter grades have two advantages over the statistical scores. First, assigning E for 60+\% discrepancies is more robust to outlier estimates. Second, multiple criteria like the sign match can be combined into one grading scheme.
%Second, assigning F to undefined results allows us to include all table cells in our evaluation, as non-numerical outputs have to be dropped when computing the statistical scores.
%

Results are aggregated first by computing the average grade across all cells within a table, and then by computing the average table grade within a paper.
Binning into grades before aggregation improves interpretability and reduces the influence of outliers. Unlike some of the prior work, our approach avoids the ambiguity and opacity of LLM-judges. In this regard, it is closest to \cite{xu2026scaling}, who also measure exact reproduction of coefficients.

\subsection{Step 4: Explanation}\label{sec:methods:explanation}

The explanation step produces interpretable diagnostics of discrepancies between the original and reproduced results. An LLM agent (GPT-5.4 with Codex CLI) is prompted to perform the analysis as follows. The input is the set of observed cell-level numerical failures (all cells with a score of B or below). For each failing output, the agent is tasked to locate the relevant code in both the original reproduction package and the agent-generated Python code. It should then identify and describe the discrepancy causing the failure. 

For each reported discrepancy, an LLM auditor runs consistency checks comparing the various input and output files. The auditor can identify four main types of errors. In the bucket of \textit{Human Error}, we have \textit{Missing Data} and  \textit{Paper vs. Code} (the original paper did not properly or completely describe the code or contradicts it). Under \textit{Agent Error}, we have \textit{Paper vs. Methods Extraction} (the methods extraction pipeline did not properly or completely describe the methods) or \textit{Method Extraction vs. Agent} (the agent did not properly follow the extracted methods description). 

%by exploring beginning by checking whether the data required for correct implementation are missing from the reproduction package. %A single trace call can resolve multiple related outputs simultaneously via an \texttt{also\_explains} field. 
%For discrepancies not attributable to missing data, the agent performs an error source diagnosis and summarizes. 

\section{Application to I4Replication data}

\subsection{Data}

Our analysis is based on papers with verified reproducibility, enabling meaningful comparisons across reproduction results.
We draw on the work of \textit{I4Replication}, an institute coordinating a large-scale research community effort to reproduce social science papers \citep{i4replication2024}.
Their process combines systematic review of reproduction packages, step-by-step code execution, and direct exchange with the original authors, resulting in a detailed reproducibility report and classification for each paper.
We collect the PDFs and reproduction packages of papers that I4Replication has classified as fully reproducible and retain those with sufficient or partial data availability and extractable tables, leading to a total dataset of 48 papers (see Appendix Table~\ref{tab:app-sample-funnel}).

% \begin{table}[h]
% \centering

% \caption{Papers by journal and discipline (final sample, $N=48$).}
% \begin{tabular}{llr}
% \toprule
% Journal & Discipline & N papers \\
% \midrule
% AJPS & Political Science & 11 \\
% EJ & Economics & 9 \\
% AER & Economics & 8 \\
% AEJ:Pol & Economics & 5 \\
% JOP & Political Science & 5 \\
% AEJ:AP & Economics & 3 \\
% APSR & Political Science & 3 \\
% REStud & Economics & 2 \\
% AEJ:Mac & Economics & 1 \\
% QJE & Economics & 1 \\
% \midrule
% Total &  & 48 \\
% \bottomrule
% \end{tabular}

% \label{tab:journal_discipline}
% \end{table}

Appendix Table~\ref{tab:sample_papers} lists all the papers with associated journal, title, main programming language, and number of lines of code. Table~\ref{tab:summary_stats_panels} Panel A tabulates the counts by journal. The papers are from a mix of well-regarded economics and political science journals. Summary stats on the code are shown in Appendix Table~\ref{tab:overview-lang-loc}. The original analysis code is mostly programmed in Stata (56.3\%) or R (27.1\%), with zero papers in Python. The packages have an average of 5,599 lines of code.

\begin{table}%h]
\centering
\small

\textbf{Panel A: Papers by journal and discipline (final sample, $N=48$)}\\[3pt]
\begin{tabular}{llr}
\toprule
Journal & Discipline & N papers \\
\midrule
\textit{American Journal of Political Science (AJPS)} & Political Science & 11 \\
\textit{Economic Journal (EJ)} & Economics & 9 \\
\textit{American Economic Review (AER)} & Economics & 8 \\
\textit{American Economic Journal: Economic Policy (AEJ:Pol)} & Economics & 5 \\
\textit{Journal of Politics (JOP)} & Political Science & 5 \\
\textit{American Economic Journal: Applied Economics (AEJ:AE)} & Economics & 3 \\
\textit{American Political Science Review (APSR)} & Political Science & 3 \\
\textit{Review of Economic Studies (REStud)} & Economics & 2 \\
\textit{American Economic Journal: Macroeconomics (AEJ:Macro)} & Economics & 1 \\
\textit{Quarterly Journal of Economics (QJE)} & Economics & 1 \\
\midrule
Total &  & 48 \\
\bottomrule
\end{tabular}

\vspace{1.5em}

\textbf{Panel B: Extracted Elements in Original Papers (per paper, $N=48$)}\\[3pt]
\begin{tabular}{l>{\bfseries}r@{\hspace{16pt}}rrrr}
\toprule
 & Total & Mean & SD & Min & Max \\
\midrule
Tables & 222 & 4.6 & 2.8 & 1 & 11 \\
Cells (present) & 14,214 & 296.1 & 261.4 & 20 & 1433 \\
Coefficients & 5,149 & 107.3 & 93.5 & 8 & 384 \\
Standard errors & 4,253 & 88.6 & 85.3 & 0 & 319 \\
p-values & 779 & 16.2 & 32.5 & 0 & 127 \\
t-statistics & 10 & 0.2 & 1.4 & 0 & 10 \\
Confidence intervals & 112 & 2.3 & 16.2 & 0 & 112 \\
R-squared & 590 & 12.3 & 17.9 & 0 & 58 \\
N observations & 1,701 & 35.4 & 86.2 & 0 & 602 \\
F-statistics & 121 & 2.5 & 6.2 & 0 & 30 \\
Other numeric & 1,607 & 33.5 & 53.6 & 0 & 253 \\
\bottomrule
\end{tabular}

\vspace{1.5em}

\textbf{Panel C: Completion Counts and Rates (\%) by Paper Element and Agent System (Confidence Intervals Excluded)}\\[3pt]
\resizebox{\textwidth}{!}{%
\begin{tabular}{lrrrrrrr@{\hspace{12pt}}>{\bfseries}c}
\toprule
\textbf{Scaffold}
& \multicolumn{1}{c}{\textbf{Claude Code}}
& \multicolumn{2}{c}{\textbf{Codex CLI}}
& \multicolumn{2}{c}{\textbf{SWE-Agent}}
& \multicolumn{2}{c}{\textbf{OpenCode}}
& \multicolumn{1}{c}{\textbf{Average}} \\
\textbf{LLM}
& \multicolumn{1}{c}{\textbf{Opus 4.6}}
& \multicolumn{1}{c}{\textbf{GPT-5.3 Codex}}
& \multicolumn{1}{c}{\textbf{GPT-5.4}}
& \multicolumn{1}{c}{\textbf{GPT-5.4}}
& \multicolumn{1}{c}{\textbf{GLM-5}}
& \multicolumn{1}{c}{\textbf{GPT-5.4}}
& \multicolumn{1}{c}{\textbf{GLM-5}}
& \\
\midrule
Papers (non-F) & 48 (100\%) & 46 (96\%) & 46 (96\%) & 47 (98\%) & 46 (96\%) & 46 (96\%) & 44 (92\%) & 96\% \\
Tables & 202 (95\%) & 210 (95\%) & 194 (91\%) & 190 (93\%) & 173 (95\%) & 184 (88\%) & 184 (86\%) & 92\% \\
\midrule
\textbf{Cells (Overall)} & \textbf{10,037 (90\%)} & \textbf{9,835 (81\%)} & \textbf{9,938 (85\%)} & \textbf{8,261 (76\%)} & \textbf{7,501 (77\%)} & \textbf{9,416 (86\%)} & \textbf{7,670 (66\%)} & \textbf{80\%} \\
\midrule
Coefficients & 3,509 (88\%) & 3,236 (77\%) & 3,446 (84\%) & 3,025 (79\%) & 3,050 (84\%) & 3,224 (86\%) & 2,736 (67\%) & 81\% \\
Standard errors & 2,752 (90\%) & 2,481 (73\%) & 2,612 (80\%) & 1,716 (57\%) & 1,730 (64\%) & 2,491 (84\%) & 1,763 (55\%) & 72\% \\
p-values & 551 (85\%) & 620 (92\%) & 594 (91\%) & 483 (84\%) & 416 (67\%) & 491 (84\%) & 350 (56\%) & 80\% \\
t-statistics & 10 (100\%) & 10 (100\%) & 10 (100\%) & 10 (100\%) & 10 (100\%) & 10 (100\%) & 10 (100\%) & 100\% \\
R-squared & 428 (86\%) & 464 (89\%) & 395 (81\%) & 415 (82\%) & 349 (82\%) & 380 (89\%) & 371 (74\%) & 83\% \\
N observations & 1,220 (97\%) & 1,392 (97\%) & 1,283 (92\%) & 1,251 (96\%) & 710 (83\%) & 1,244 (91\%) & 1,200 (87\%) & 92\% \\
F-statistics & 97 (90\%) & 93 (85\%) & 94 (90\%) & 104 (94\%) & 48 (77\%) & 100 (90\%) & 68 (61\%) & 84\% \\
Other numeric & 1,046 (91\%) & 1,123 (85\%) & 1,078 (89\%) & 1,091 (89\%) & 905 (88\%) & 1,082 (85\%) & 874 (73\%) & 86\% \\
\midrule
\bottomrule
\end{tabular}
}
\caption{\textbf{Descriptive overview } \textit{Panel A: number of included papers by journal. Panel B: per-paper summary statistics on the original outputs. Panel C: reproductions, counts are completed outputs, with percentages relative to each agent system's assessed targets in parentheses. Confidence intervals are excluded from cell and table denominators. Completion measures usable numerical presence, not correctness; a non-F paper need not have every table completed. Average is the mean of the shares across the 7 approach--model combinations.}}
\label{tab:summary_stats_panels}
\end{table}

Summary statistics on the extracted paper elements are reported in Table~\ref{tab:summary_stats_panels} Panel B. The system extracted 222 ground-truth tables containing 14,214 table cells. Of these, about one-third (5,149) contain regression coefficients. About four-fifths (4,253) of coefficients are accompanied by a standard error, while one-sixth (779) have p-values. Other frequently observed statistics are the number of observations and R-squared.  

A potential issue with this application is data leakage: tested LLMs may have been exposed to the manuscripts or code during pre-training. To probe the relevance of leakage, we provide an auxiliary analysis reproducing results from papers published after the model knowledge cutoff, which is August 2025 for both GPT 5.3-Codex/5.4 \citep{openai2026gpt54} and Claude Opus 4.6  \citep{anthropic2026modelsoverview}. Specifically, we sample ten papers (with reproduction packages) from the \emph{Economic Journal} (EJ), with five papers published \emph{before} and five papers published \emph{after} August 2025. While these papers have not yet been verified as reproducible by I4Replication, they have been verified by EJ's data editor team, which carefully vets reproduction packages and verifies that they produce the same outputs as the paper. 

\subsection{Agent systems}

The LLM agents tested are GPT-5.4, GPT-5.3 Codex,
 Claude Opus 4.6, and the open-weights GLM-5 \citep{glm5team2026glm5}. We evaluate two proprietary agent scaffolds---Claude Code and Codex---and two open-source alternatives---mini-SWE-agent \citep{yang2024sweagent} and OpenCode. %\footnote{\url{https://opencode.ai/}} 
 For the open-source scaffolds, we compare a proprietary model (GPT-5.4) and an open-weights alternative (GLM-5). Figure \ref{fig:app-screenshots} illustrates some sample code segments generated by the agents.

Before evaluating success in reproduction, a preliminary question is whether these agent systems can even generate any results at all. Table~\ref{tab:summary_stats_panels} Panel C summarizes the completion rates according to various features of the paper results. The agents deliver usable results for at least 92\% (and up to 100\%) of papers, 86\% (up to 95\%) of tables, and 66\% (up to 90\%) of cells. The bottom part of the table shows completion across various types of cells, with the most important ones being coefficients (81\% completion on average) and standard errors (72\% completion on average). In terms of overall cell completion, the best agent system is Claude Opus 4.6 and the worst is OpenCode GLM-5.
 
 %\footnote{We acknowledge that results from AI agents are non-deterministic. A systematic analysis of the within-agent consistency between different runs would be valuable, but is unfortunately cost-prohibitive.}

% \subsection{Failed runs and evaluation}
% The agent independently reads and executes a response to the task. As a result, we do not have control over the agent leaving out cells, tables or failing to fill in values for particular tables.
% %
% We also do not check which tables are affected by partial data availabilty that could lead to some non-producable results.
% %
% We exclude missing values from the evaluation, missing cells are ignored for tables, and missing tables omitted for paper grades.
% %
% We restart agents that fail on all tables, e.g., by not creating any code.

\section{Results: How well do AI agents reproduce papers without their code?}
\label{sec:results}

% \subsection{Summary Statistics}

% Summary Statistics Table
% number of papers
% number of tables
% number of each kind of statistic
% positive coefficients
% number of missing values

% Summary Histograms
% distributions of true and estimated coefficients
% distributions of true and estimated SE's
% distributions of true and estimated p-values
% distributions of true and estimated T-stats
% etc

% separate summary stats on originals and reproductions

\subsection{Main results}
This section presents the main results on reproducing the I4Replication papers. We start at the cell (estimate) level and proceed to table- and paper-level results.

\begin{figure}%[ht]
    \centering
    \begin{subfigure}[t]{0.48\textwidth}
        \centering
        \includegraphics[width=\textwidth]{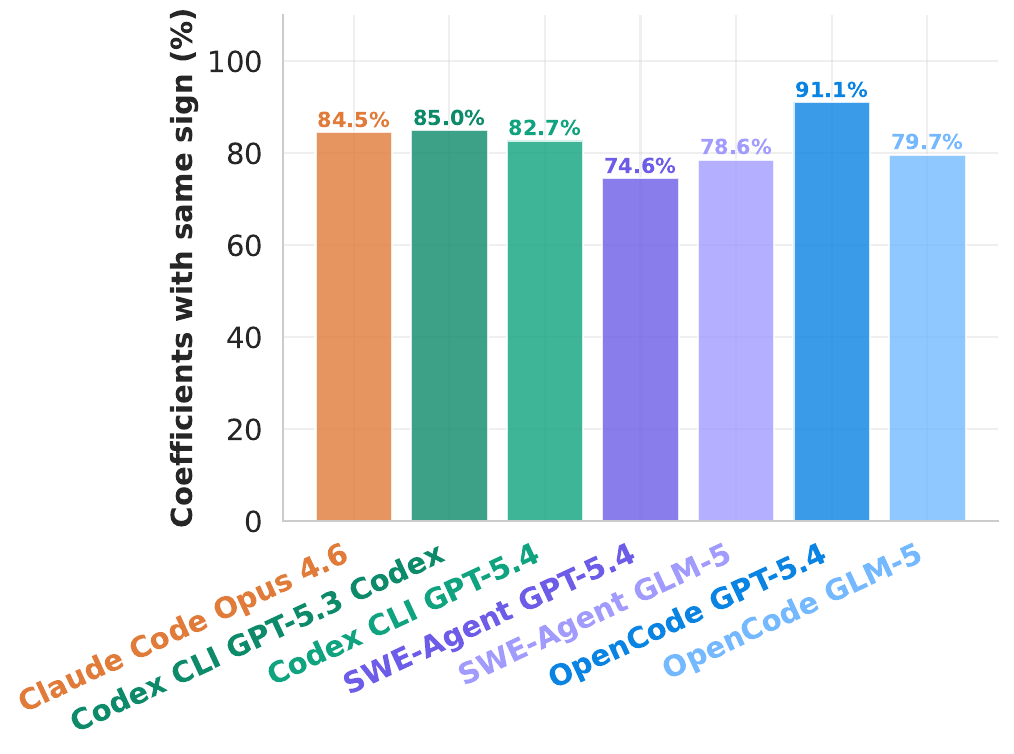}
        \caption{Share of reproduced coefficients with the same sign as the original paper.}\label{fig:coeff-results:sign}
    \end{subfigure}
    \hfill
    \begin{subfigure}[t]{0.48\textwidth}
        \centering
        \includegraphics[width=\textwidth]{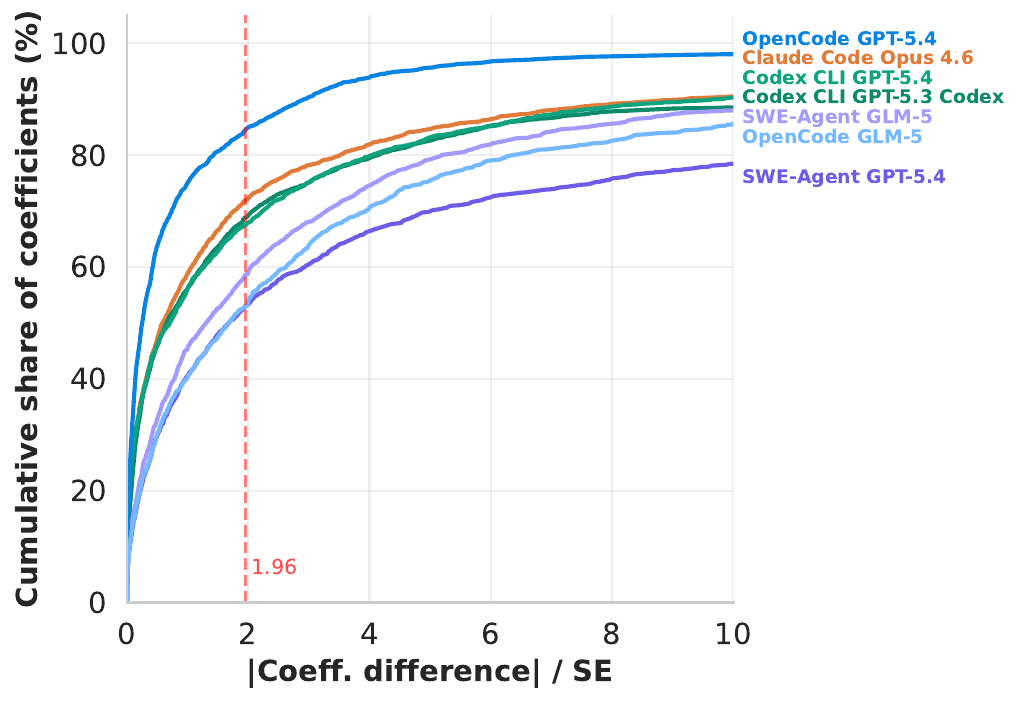}
        \caption{Cumulative distribution of $|$coefficient difference$|$ / SE across approaches. The standard error is the original standard error of the coefficient reported in the paper. The dashed red line marks 1.96 (95\% CI threshold).}\label{fig:coeff-results:cumdist}
    \end{subfigure}
    \caption{\textbf{Performance metrics of agent-reproduced coefficients}. \textit{Coefficients are identified via an LLM table extraction pipeline that classifies all numerical values into one of multiple statistic types. Agentic systems can reliably recover a large share of the directional findings from social science publications given the publication text and data alone. All metrics are based on reproduced results excluding missing coefficients. Panel (a): The best performing agent recovers the correct sign over 90\% of the time. Panel (b): The reproduced coefficients of the best performing agent are within the 95\% confidence interval (based on the ground-truth standard error) over 80\% of the time.}}
    \label{fig:coeff-results}
\end{figure}

\paragraph{Cell-level.} Figure~\ref{fig:coeff-results} reports cell-level results for regression coefficients, which are the most prevalent statistic in tables. First, \Cref{fig:coeff-results:sign} reports the share of reproduced estimates that match the sign of the original estimates. Across all models and scaffolds, reproduced coefficients match the sign for the large majority of observations, ranging from 75\% (SWE-Agent GPT-5.4) to 91\% (OpenCode GPT-5.4). In all cases, that exceeds the naive ``guess positive'', which would give 68\% accuracy. Note that the numbers in \Cref{fig:coeff-results:sign} ignore ``missing'' coefficients that were not generated. For comparison, Appendix Figure \ref{fig:coeff-results-with-missing} includes missing coefficients in the denominator. % guess positive = 68\%

Next, \Cref{fig:coeff-results:cumdist} presents cumulative distributions (CDFs) of the absolute difference of the original and reproduced coefficients, divided by the ground-truth standard error. This normalized difference measures the discrepancy between original and reproduced coefficients in statistically meaningful units (same units as the Wald t-statistic). For these curves, better performance is indicated by hugging the top-left corner. The value of the CDF at x = 1.96 indicates the share of reproduced estimates which lie within the 95\% confidence interval (CI), using 1.96 original standard errors as the discrepancy threshold. Here, we see that the reproduced coefficient falls within the 95\% CI for more than half of all reproduced values, even in the worst-performing case. For the best model (OpenCode GPT 5.4), over 80\% of reproduced values are within the 95\% CI.

The CDFs further reveal that a non-negligible share of reproduced coefficients deviate substantially from the originals. An informal inspection of these cases suggests that while some are attributable to agent behavior (e.g., selecting the wrong dataset from among several similar candidates), the majority stem from unit scaling issues not accounted for by our comparison system (e.g., the original authors report values in dollars, whereas the dataset is denominated in cents).

\begin{figure}%[ht]
    \centering
    \includegraphics[width=0.9\textwidth]{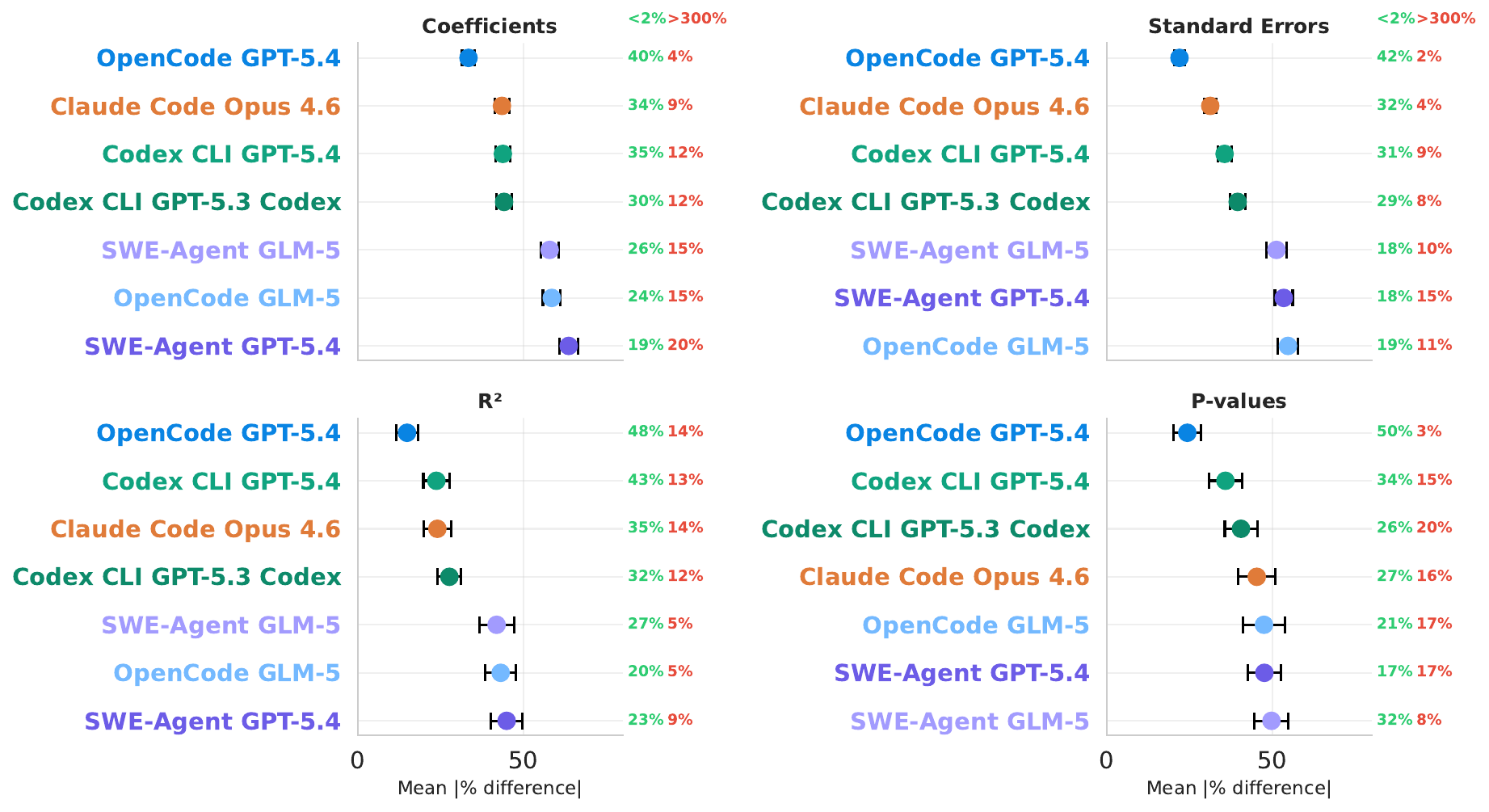}
    \caption{\textbf{Mean percentage differences between original and reproduced, by statistic type}. \textit{The strongest models can reproduce more than 40\% of produced coefficients and standard errors exactly after rounding. Note that these statistics are only produced when the original table contains them, and therefore the number of observations per table varies. We drop observations beyond a 300\% difference, as these outliers frequently are due to scaling issues. Green and red numbers next to the plot indicate both the share of these large outliers (red) and the share that are (close to) perfectly reproduced (green).}}
    \label{fig:pct-diff-cell-type}
    
\end{figure}

Figure~\ref{fig:pct-diff-cell-type} measures reproduction error as the absolute percentage difference between the original and reproduced statistic, with additional heterogeneity reported  by  type of statistics (coefficient, standard error, R-squared, and p-value). Across all statistic types, the best agent is OpenCode GPT 5.4. The worst agent is SWE-Agent GPT-5.4 (3/4 stats) or OpenCode GLM-5 (1/4 stats). Codex GPT-5.4, Codex GPT-5.3, and Claude Code Opus 4.6 perform similarly, forming a second tier behind OpenCode GPT-5.4.  SWE-Agent performs the worst among the scaffolds, even with GPT-5.4 as the underlying model.

\begin{figure}
    \centering
    \begin{subfigure}[t]{0.48\textwidth}
        \centering
        \includegraphics[width=\textwidth]{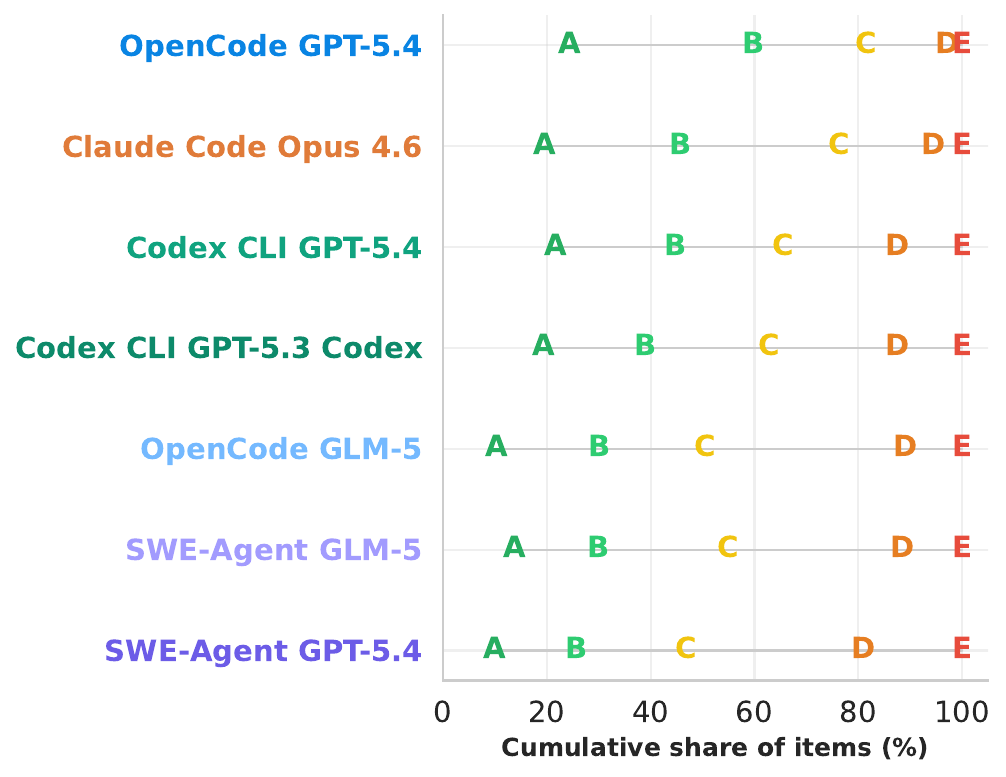}
        \caption{Table-level grades}
    \end{subfigure}
    \hfill
    \begin{subfigure}[t]{0.48\textwidth}
        \centering
        \includegraphics[width=\textwidth]{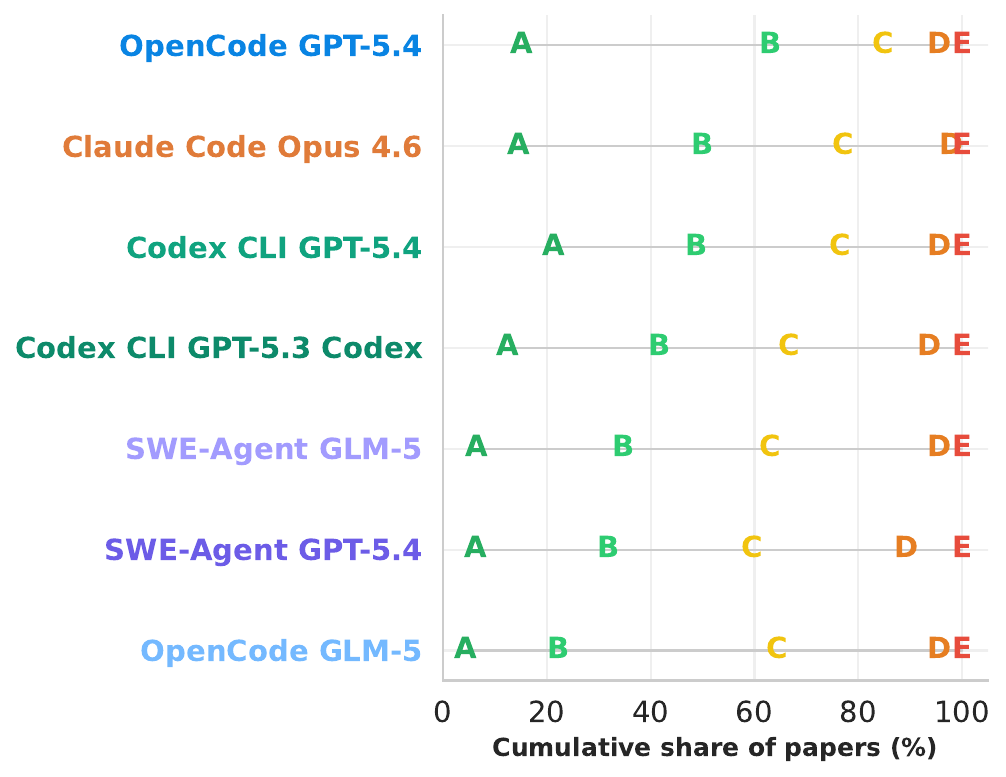}
        \caption{Paper-level grades}
    \end{subfigure}
    \caption{\textbf{Grade comparison by aggregation level and agent}. \textit{Grades are assigned to percentage differences to reproduced and original results following the rubric in Table~\ref{tab:grading_detailed} and then aggregated with equal weights. All reproduced numbers are rounded to the number of digits reported in the original papers. At table level, the highest shares are about a quarter A and 60\% A or B; at paper level, the corresponding maxima are about a fifth and 63\%. Observations are ranked by their share of A and B grades. Equivalent distributions including the F-grades can be found in Appendix Figure~\ref{fig:app-agg-grades-with-f}.} }
    \label{fig:agg-grades}
\end{figure}

\paragraph{Table- and paper-level.} Figure \ref{fig:agg-grades} reports results with letter grades, aggregated by table (panel a) and paper (panel b). Converting numerical values to grades mitigates the influence of outliers when comparing and aggregating results. Overall, the results are consistent when looking at tables and papers. The table-level aggregation highlights performance differences across agents: the best-performing model--scaffold combination reproduces twice as many tables near-perfectly (grade A) compared to the worst-performing one. 

Both the table- and paper-level aggregations indicate that achieving accurate and reliable reproduction beyond individual cells remains challenging for all agents. Agent rankings become less differentiated at the paper level, suggesting that some agents tend to perform consistently well across all tables once they successfully reproduce one (e.g., Codex with GPT-5.4), whereas others solve individual tables more independently, leading to greater within-paper performance variance.

Appendix Figure \ref{fig:app-table-grade-by-type} shows table results separately by the empirical function of the table -- main results, mechanism analysis, robustness checks, or descriptive statistics. The results show that descriptive statistics are easier to reproduce than the results tables, which demonstrate similar performance across main, mechanisms, and robustness. Appendix \ref{sec:appendix-determinants} provides a number of other supporting results on correlates of reproduction performance.

\subsection{Variation in agentic effort}

\begin{figure}%[ht]
    \centering
    
    \begin{subfigure}[t]{0.32\textwidth}
        \centering
        \includegraphics[width=\textwidth]{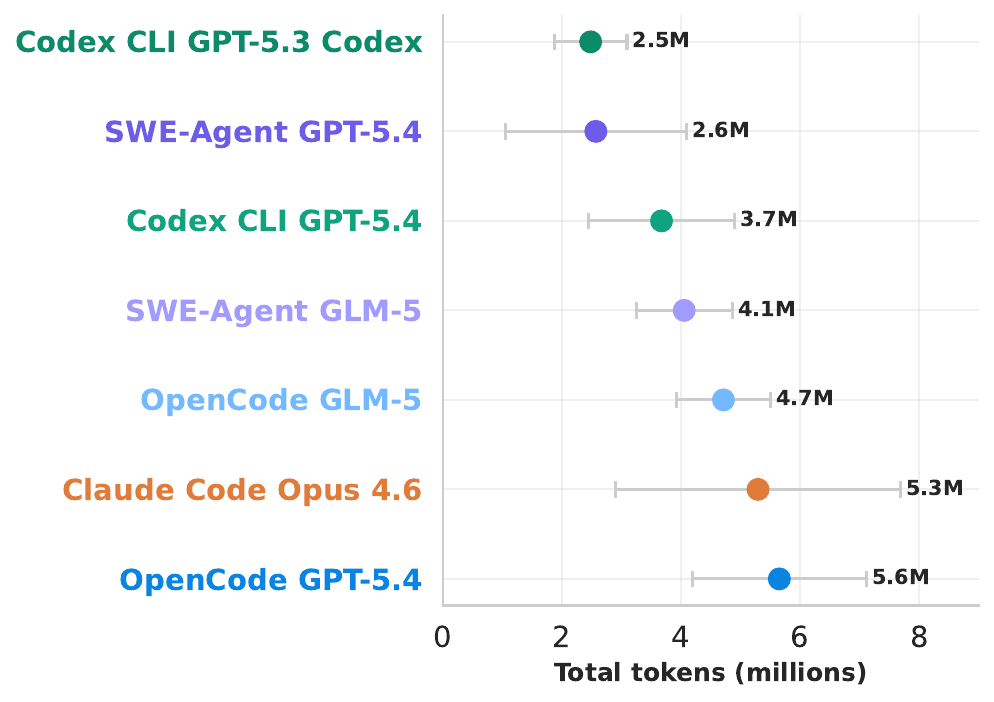}
        \caption{Token usage by run}
        \label{fig:app-token-usage}
        \end{subfigure}
    \hfill
    \begin{subfigure}[t]{0.32\textwidth}
        \centering
        \includegraphics[width=\textwidth]{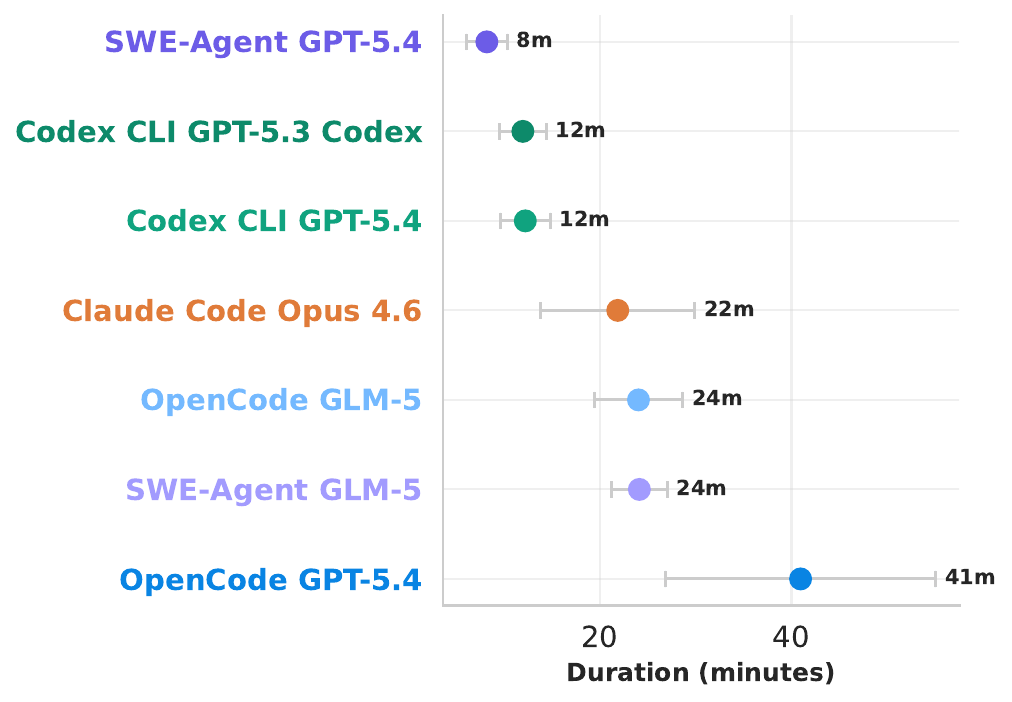}
    \caption{Duration of runs}
        \label{fig:app-duration}
    \end{subfigure} 
   \hfill
    \begin{subfigure}[t]{0.32\textwidth}
    \centering
    \includegraphics[width=\textwidth]{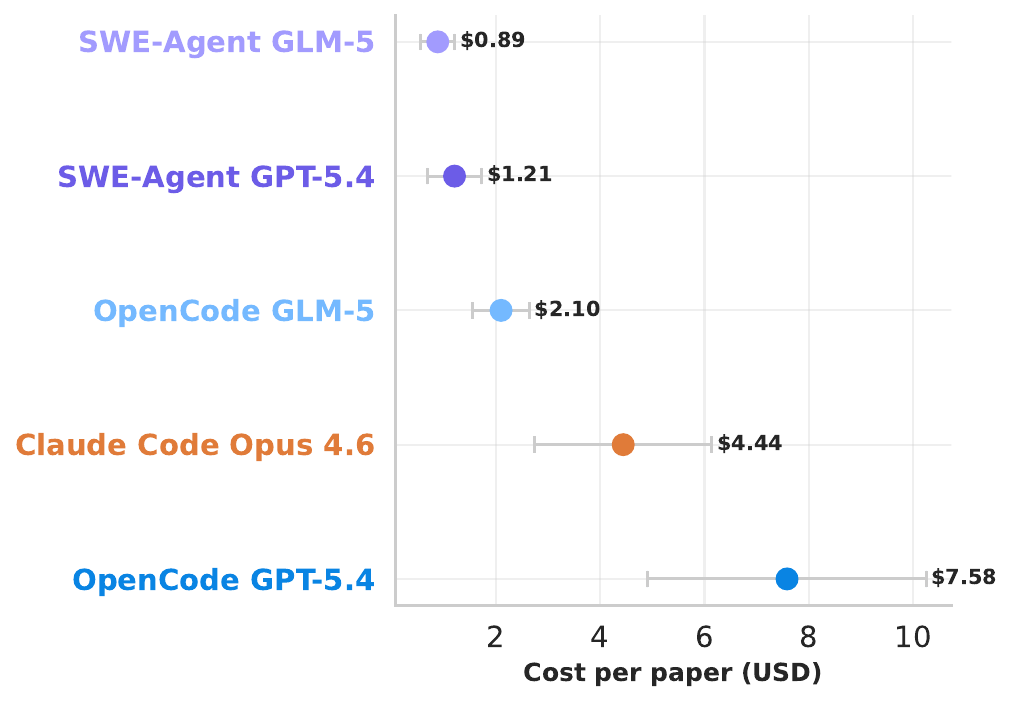}
    \caption{Costs of runs}
        \label{fig:app-costs}
            \end{subfigure}
    
    \caption{\textbf{Token usage, run durations, and run costs, by agent}. \textit{Error bars indicate 95\% confidence intervals. Token usage includes cached and uncached input and output tokens. Codex CLI (through subscription) did not report usage costs. OpenCode in combination with GPT 5.4 consumes more tokens, takes longer, and costs more.}}    \label{fig:app-tokens-durations-cost}
\end{figure}

To better understand performance differences across agents, we examine variation in computational effort, measured by token usage, runtime, and cost. Figure~\ref{fig:app-tokens-durations-cost} shows substantial heterogeneity along all three dimensions.

The best-performing agent, OpenCode GPT-5.4, consistently operates at a higher level of effort. It consumes many more tokens and takes much more time per run than competing approaches, and it is also the most costly configuration. This pattern suggests that its superior performance is associated with greater computational investment: more extensive exploration, longer reasoning chains, or more iterations over the data and code. By contrast, lower-performing agents tend to operate under tighter computational budgets, which may limit their ability to resolve ambiguities or recover from intermediate errors.

Token usage statistics give us an indication of the overall effort on a task but not how it is spent. To better understand the composition of this effort, we use an LLM to classify each agent action into one of six categories (execution, reading, navigation, search, writing, or other) and count both the number of actions and the volume of text (in characters) produced in each action (details in Appendix~\ref{app:trace-analysis}).

Figure \ref{fig:actions-per-run} shows the breakdown. For most models, the main tool call is execution, followed by a mix of reading and writing, and then other tools. However, OpenCode GPT-5.4 makes more than twice as many tool calls as other setups, spending more calls on execution and many more calls on reading files. When weighting the tool calls by volume of text (right panel), however, the picture looks very different. Most tokens are spent on writing, with almost all of the remaining tokens spent on execution. As a share of tokens, reading, searching, navigation, and other are negligible tasks. Qualitative analysis reveals that while GPT-5.4 mostly favors inline Python execution via heredocs, GLM-5 writes scripts to disk before execution, explaining the larger fraction of characters spent on writing files.

\begin{figure}%[ht]
    \centering
    \begin{subfigure}[t]{0.48\textwidth}
        \centering
        \includegraphics[width=\textwidth]{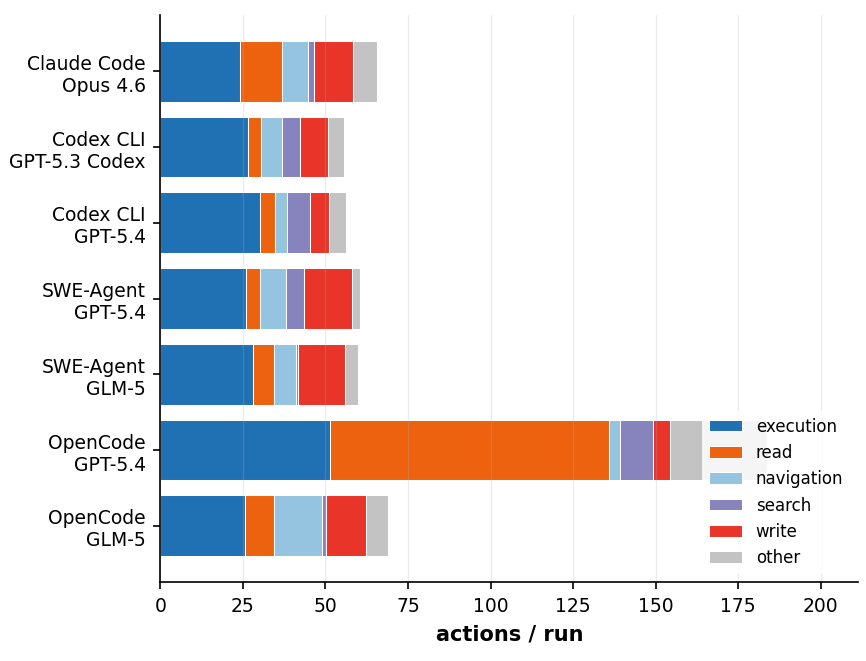}
\caption{}
\label{fig:actions-per-run}
    \end{subfigure}
    \hfill
    \begin{subfigure}[t]{0.48\textwidth}
        \centering
        \includegraphics[width=\textwidth]  {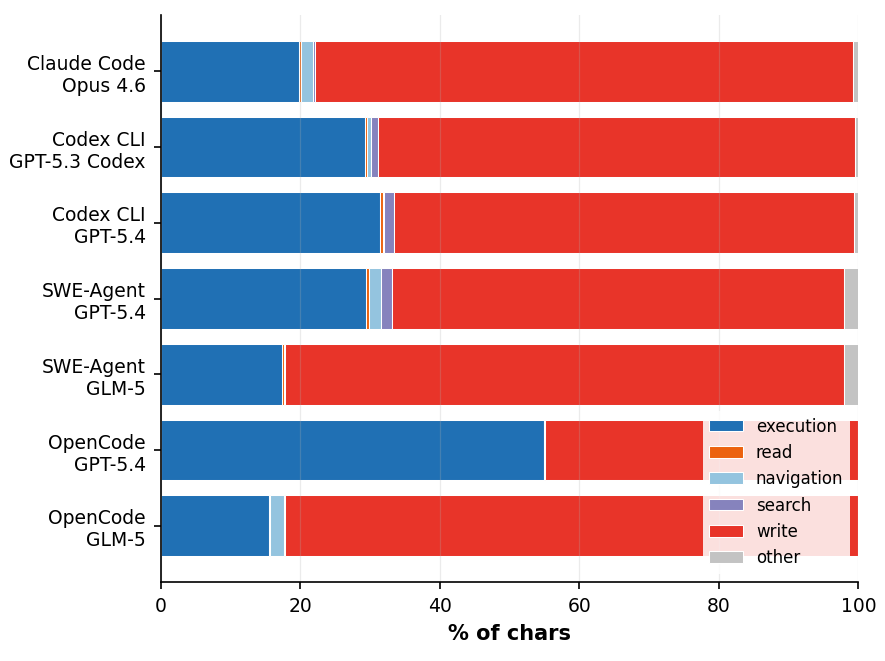}
        \caption{}
        \label{fig:chars-for-actions-per-run}
    \end{subfigure}
    \caption{\textbf{Effort allocation across tasks.} The figures show the number of tool-call actions taken by the agents (left) and the relative volume of text (number of characters) emitted (right), colored by function category.}
    \label{fig:trace-analysis}
\end{figure}

These results highlight a key trade-off in agentic reproduction. Performance gains are a function of the effort (tokens) expended during execution. In this sense, the strong performance of OpenCode GPT-5.4 reflects both effective scaffold–model alignment and a willingness to spend more tokens and time on each task. This finding is consistent with prior work emphasizing the importance of scaffold–model interaction for agentic performance \citep{cohn2026evidencedecisionfeedbacktheorydrivenadaptivescaffolding}, and suggests that differences in observed accuracy may partly reflect differences in implicit compute budgets rather than purely differences in capability.

Appendix Table~\ref{tab:papergrade_tokens_contrast} reports the within-paper association between completion tokens and reproduction performance.

\subsection{Exploration of error sources} 
%The failure---of either a machine or human---to reproduce a scientific result can be attributed to issues with the original result or with the replication attempt \citep{freese2017replication}. Understanding these discrepancies is inherently complex: the agent must reconcile code across different programming languages, locate relevant sections spread across multiple files, and disregard ancillary code (e.g., for figures). This also makes validating the error attribution pipeline difficult. The results, based on runs over ten randomly selected papers, should therefore be treated as qualitative signals rather than precise estimates.

\begin{figure}%[ht]
    \centering
    \includegraphics[width=0.72\textwidth]{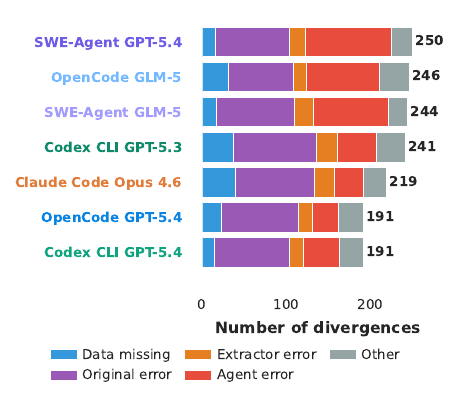}
    \caption{\textbf{Error analysis pipeline:  Number of divergences by error source.} \textit{The error pipeline identifies divergences between the original and reproduced code for tables that score worse than overall grade B. These divergences are described and classified into categories. The agent then traces the root cause of the divergence. \textbf{Agent error:} the agent makes an error in mapping the methods to reproduced code. \textbf{Extractor error:} the initial extraction from paper to methods description is faulty. \textbf{Original error}: the paper under-specifies or mis-specifies the original code; this does not imply that the original findings are erroneous. \textbf{Data missing}: data is missing in the reproduction package. Appendix~\ref{sec:app-error-source} has more detail.}
    % Mismatches between agent and summary, paper and summary, and paper and code respectively are further decomposed into more granular categories (see Appendix  ~\ref{}\needscite{}) 
   }
    \label{fig:disc_root_cause_agg}
\end{figure}

Where do these errors originate? Figure~\ref{fig:disc_root_cause_agg} summarizes the root causes identified by our discrepancy pipeline across model-scaffold combinations, grouped into five categories: \textit{agent error}, \textit{extractor error}, \textit{original error}, \textit{data missing}, and \textit{other/unknown}. As shown, the large majority of divergences—over three-fourths—can be traced to a specific, interpretable source.

Because these attributions rely on LLM judgments, we treat their aggregate shares as indicative rather than definitive. A targeted human audit confirms 49 of 53 flagged divergences (92\%) as meaningful, but confirms only 33 of 49 error-source labels (67\%); 10 further labels are judged unsure rather than incorrect. Appendix~\ref{sec:app-error-source-human-audit} reports the audit design, category-level results, and inter-annotator agreement. A matched comparison with a Claude attribution judge also changes the agent-error share (Appendix~\ref{sec:app-error-source-judge-sensitivity}).

A substantial share of discrepancies are attributed to sources outside the reproducing agent's control. First, the largest share of divergences is attributed to \textit{original errors}, indicating mismatches between the paper and the underlying code. In other words, the methods descriptions in the paper are often insufficiently precise to enable faithful reimplementation. Another substantial share of errors comes from missing data. Finally, a number of discrepancies arise from errors in the extraction pipeline -- done in our shared data construction step, before the agent systems start their work.

Agent errors constitute the second largest category of discrepancies. For the strongest agents, the share of agent errors declines markedly, becoming a relatively minor component of total discrepancies. A larger relative fraction of divergences is attributed to human sources --- either to underspecification in the original paper or to missing data. This pattern suggests that limitations in how methods are documented, alongside agent capability, are an important barrier to automated reproducibility.

There is substantial variation across agents in the number of discrepancies attributed to human sources, such as paper underspecification. This variation does not imply that some agents encounter fewer human-caused issues, but rather that they handle these ambiguities differently. Even when the underlying discrepancy originates from incomplete or unclear reporting in the paper, agents may still reproduce the correct result if their implicit assumptions happen to align with the original implementation. As a result, the observed number of human-attributed discrepancies reflects both the presence of underspecification and the agent’s ability to resolve it correctly.

% Custom colors
\definecolor{cellred}{RGB}{255, 200, 200}
\definecolor{cellgreen}{RGB}{180, 230, 180}
\definecolor{cellgray}{RGB}{235, 235, 235}
\definecolor{headergray}{RGB}{100, 100, 100}

% \thispagestyle{empty}

% \begin{center}

% \vspace{14pt}

% Column layout:
% col1: Agent (2.5cm)
% cols 2-3: Paper 1 behavior + grade (5.5cm + 0.8cm)
% cols 4-5: Paper 2 behavior + grade (5.5cm + 0.8cm)
% cols 6-7: Paper 3 behavior + grade (5.5cm + 0.8cm)
\begin{table}

\resizebox{\textwidth}{!}{%
\begin{tabular}{
  >{\bfseries}p{2.5cm}
  p{5.5cm}
  >{\centering\arraybackslash\bfseries}p{0.8cm}
  p{5.5cm}
  >{\centering\arraybackslash\bfseries}p{0.8cm}
  p{5.5cm}
  >{\centering\arraybackslash\bfseries}p{0.8cm}
}

\toprule
% Top header: paper identifiers spanning behavior+grade columns
%\rowcolor{headergray}
\color{black} &
\multicolumn{2}{c}{\color{black}\bfseries\small 10.1093/ej/ueab069 — Table 3a} &
\multicolumn{2}{c}{\color{black}\bfseries\small 10.1257/pol.20200559 — Table 2} &
\multicolumn{2}{c}{\color{black}\bfseries\small 10.1093/ej/ueab102 — Table 3}\\[-6pt]

% Sub-header: behavior / grade
%\rowcolor{headergray}
\color{black} &
\color{black}\centering\small\bfseries Behavior &
\color{black}\small\bfseries Gr. &
\color{black}\centering\small\bfseries Behavior &
\color{black}\small\bfseries Gr. &
\color{black}\centering\small\bfseries Behavior &
\color{black}\small\bfseries Gr. \\

\midrule

% claude-code / opus-4-6
Claude code \newline behavior
& \cellcolor{cellred}\small Agent assumes id for democrats is party=100
& \cellcolor{cellred} C
& \cellcolor{cellred}\small Agent assumes that codes A1 and A2 indicate democrat support
& \cellcolor{cellred} D
& \cellcolor{cellred}\small Uses \texttt{linearmodels}' \texttt{diagnostics.f.stat}
& \cellcolor{cellred} D \\

\midrule

% Opencode 5.4
OpenCode 5.4\newline behavior
& \cellcolor{cellgreen}\small Agent assumes id for democrats is party=200
& \cellcolor{cellgreen} B
& \cellcolor{cellgreen}\small Agent assumes that codes A4 and A5 indicate democrat support
& \cellcolor{cellgreen} B
& \cellcolor{cellgreen}\small Manually reconstructs the (Kleibergen-Paap) Wald F-statistic
& \cellcolor{cellgreen} B \\

\midrule

% Correct behavior
\cellcolor{cellgray} Correct\newline behavior
& \cellcolor{cellgray}\small Democrats id is party=200
& \cellcolor{cellgray}
& \cellcolor{cellgray}\small A4 and A5 indicates democrat support
& \cellcolor{cellgray}
& \cellcolor{cellgray}\small Stata uses \texttt{ivreg2} (Kleibergen-Paap) \texttt{widstat} F-statistic
& \cellcolor{cellgray} \\

\midrule

% Error attribution
\cellcolor{cellgray} Error\newline Attribution
& \multicolumn{2}{p{6.35cm}}{\cellcolor{cellgray}\small Paper underspecifies code by not disclosing how party is coded in data}
& \multicolumn{2}{p{6.35cm}}{\cellcolor{cellgray}\small Paper underspecifies code by not disclosing how party is coded in data}
& \multicolumn{2}{p{6.35cm}}{\cellcolor{cellgray}\small Paper underspecifies F-statistic type} \\

\bottomrule
\end{tabular}
}
\caption{\textbf{Examples of discrepancies of paper underspecification.} \textit{The examples illustrate how agent behavior can vary with an underspecified task. }}

\label{tab:discrepancy-underspecified-examples}

\end{table}
% \end{center}  % orphan — matching \begin{center} is commented out above

Table~\ref{tab:discrepancy-underspecified-examples} illustrates this mechanism. In all three cases, the paper omits key details—such as how party affiliation is coded or which F-statistic is used—forcing the agent to infer the correct implementation. Different agents make different assumptions: for instance, one agent incorrectly maps party codes or relies on a default statistical routine, while another reconstructs the intended specification more accurately and recovers results closer to the original. These examples show that the same human-caused ambiguity can either lead to a discrepancy or be silently resolved, depending on the agent. Consequently, higher-performing agents tend to exhibit fewer human-attributed discrepancies, not because the underlying issues are absent, but because they are more often resolved correctly during reimplementation.

\subsection{Robustness to repeated runs, agent agreement, extraction, and pre-training leakage}

This section evaluates robustness to repeated runs, inter-agent agreement, method extraction, and pre-training leakage. Since reasoning-based agentic systems are stochastic, we evaluate stability of performance across multiple runs. Some papers may have been included in pre-training, potentially leading to memorization, and so we also compare performance for a selection of papers published before and after the model knowledge cutoffs.

\paragraph{Re-run stability.}

\begin{figure}%[ht]
    \centering
    \begin{subfigure}[t]{0.48\textwidth}
       \centering
        \includegraphics[width=\textwidth]{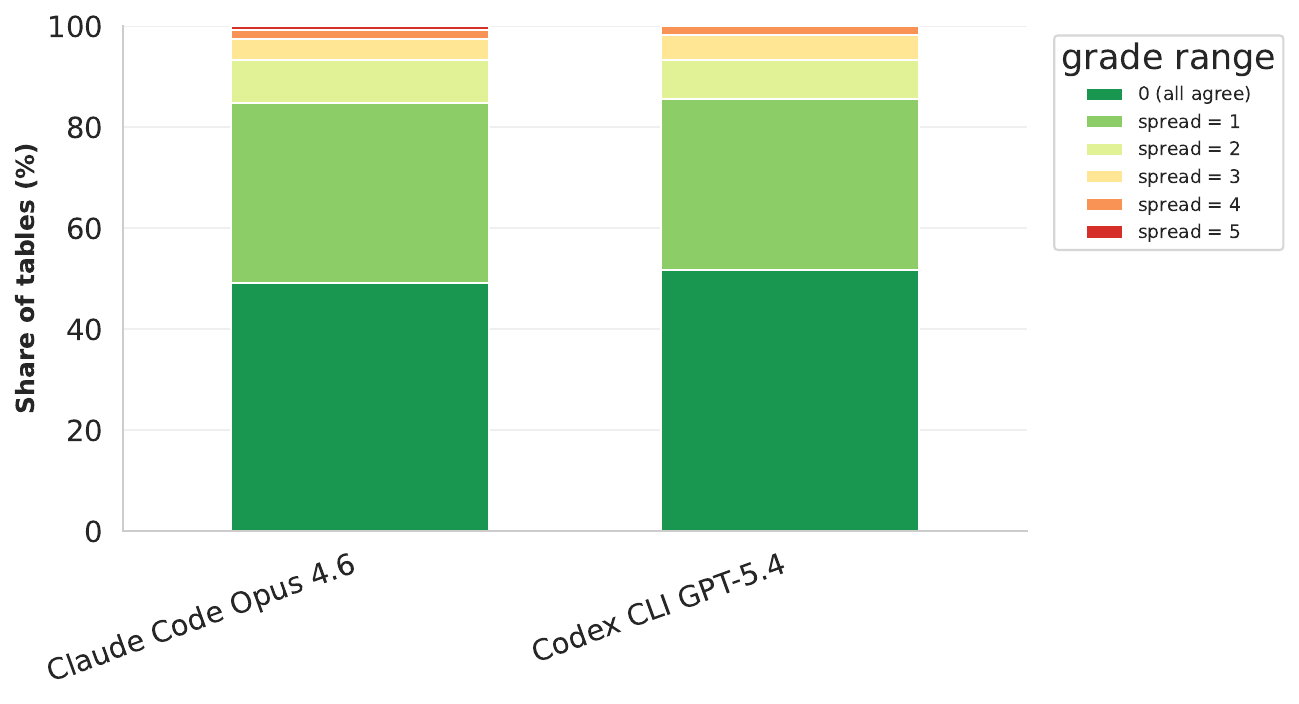}
    \caption{Distribution of table-level grade differences}
    \label{fig:app-run-stability-table-diff}
        \end{subfigure}
    \hfill
    \begin{subfigure}[t]{0.48\textwidth}
          \centering
    \includegraphics[width=\textwidth]{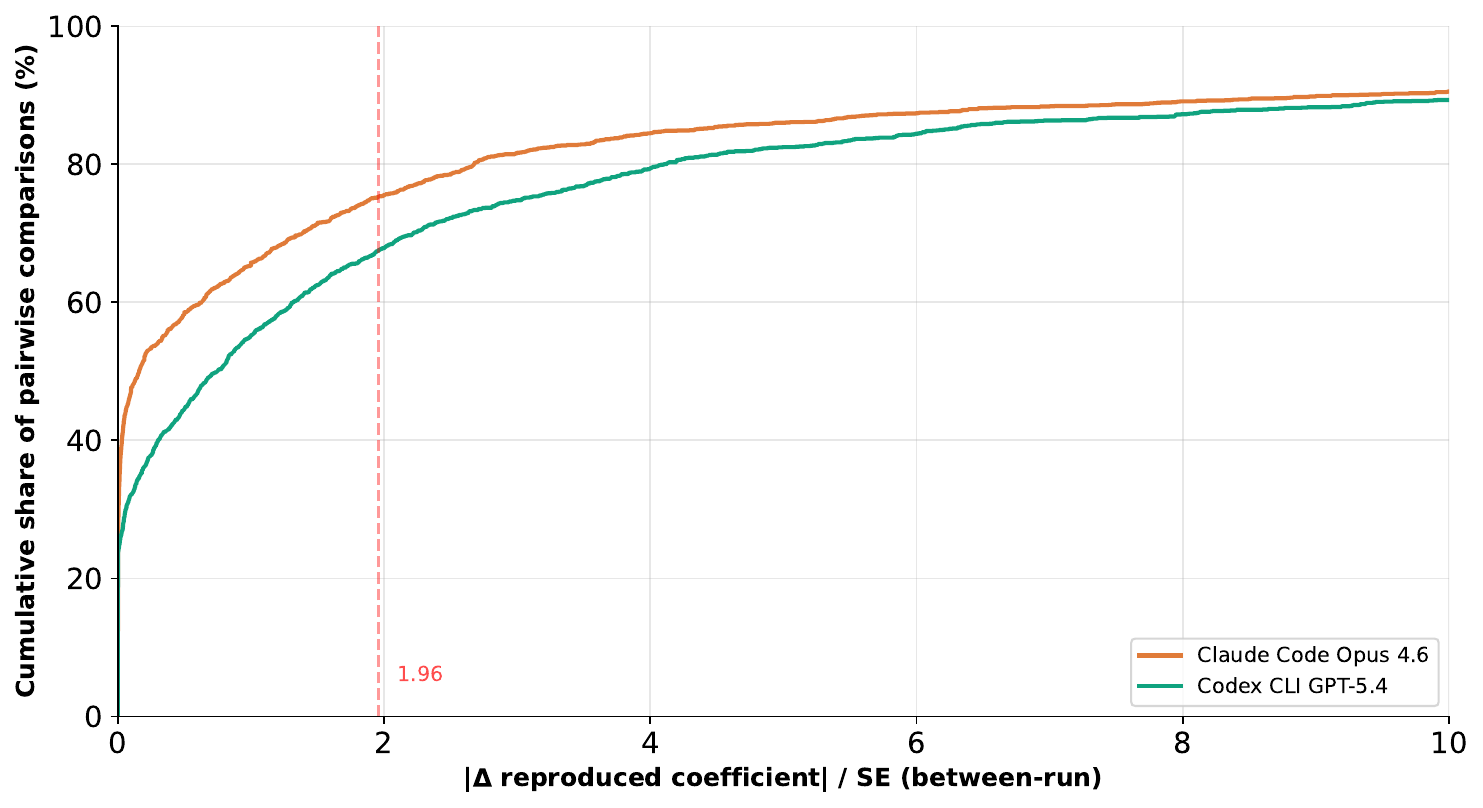}
    \caption{Cumulative distribution of absolute coefficient differences across independent reproduction runs scaled by original standard error.}
    \label{fig:run-stability-t-stat-cdf}
    \end{subfigure}
\caption{\textbf{Stability of results between multiple reproduction runs.} \textit{Both agents reproduce 20 random papers two additional times to a total of three runs. The grade range is computed as the difference between highest and lowest grade between runs excluding F grades. For the Panel B each coefficient cell that an agent replicated in two or more runs, the pairwise absolute difference between the replicated values, normalized by the original paper's standard error is computed. The CDF reports the share of pairwise comparisons at or below each threshold, using matched cells with a positive published standard error and consistent original values; ambiguous matches are excluded. Panel B contains 2,393 Claude and 2,185 Codex comparisons from 19 papers per agent. Additional figures can be found in Appendix~\ref{sec:appendix-stability}.}}
\label{fig:run-stability-coef}
\end{figure}

Because agentic systems are stochastic, we assess how sensitive reproduction outcomes are to repeated runs of the same task. We rerun the full pipeline twice for 20 randomly selected papers using Claude Code and Codex GPT-5.4, yielding three runs per paper (Figure~\ref{fig:run-stability-coef}). Overall, results are stable: more than 80\% of tables exhibit a grade spread of at most one across runs. Exact agreement is common, and large deviations are rare, indicating that most reproduction outcomes are robust to randomness in the agent’s execution.\looseness=-1

At the same time, stability at the table level masks meaningful variation at the coefficient level. Figure~\ref{fig:run-stability-t-stat-cdf} shows the distribution of within-agent-cell across-run changes in coefficient estimates, in units of the original paper's standard error. This chart suggests significant noise within agent systems, as 25--32\% of matched cross-run coefficient comparisons differ by more than 1.96 original standard errors. This suggests that while overall conclusions are stable, fine-grained comparisons between agents should be interpreted with caution, especially when differences are small.

\begin{figure}%[ht]
    \centering
    \includegraphics[width=0.6\textwidth]{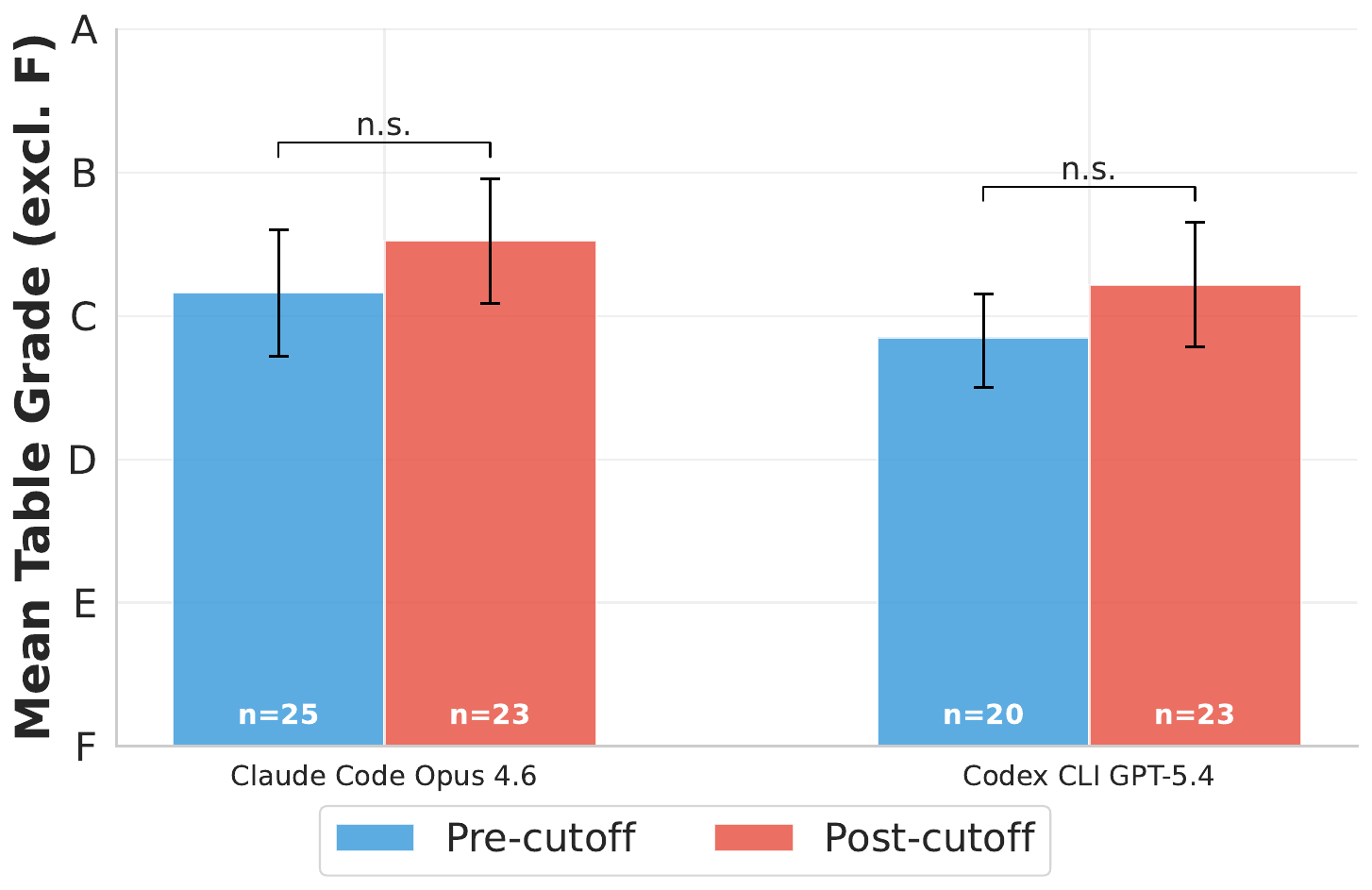}
    \caption{\textbf{Pre-training leakage evaluation. } \textit{Average table grades for sample of papers published before and after the model knowledge cutoff. No statistically significant difference is detected in this small sample; this does not rule out pre-training leakage.}}
    \label{fig:pre-post-comparison}
\end{figure}

\paragraph{Inter-agent agreement.}
Figure~\ref{fig:app-aggreement} reports paper-level grade agreement across agents. Exact agreement ranges from 33\% to 55\% across agent pairs, and 68--87\% of papers receive grades within one step. These comparisons indicate consistency of reproduction grades, but do not establish their accuracy or the quality of the original research.

\paragraph{Method extraction.}
On a ten-paper subset of the stability sample, we replace the structured method extraction with a free-form Markdown description for Claude Code (Opus 4.6) and Codex (GPT-5.4). The coarser description lowers paper grades by roughly one-third of a grade step on average, within the spread observed across repeated runs (Appendix~\ref{sec:appendix-extractor-variant}). This comparison tests one alternative prompt and does not establish robustness to all extraction designs.

\paragraph{Leakage analysis.}

To assess the potential role of pre-training leakage—where models may have been exposed to results during training—we compare performance on papers published before and after the model knowledge cutoff. Specifically, we reproduce five \textit{EJ} papers published before and five published after the cutoff using Claude Code (Opus 4.6) and Codex (GPT-5.4), and test for differences in performance between the two groups.

Figure~\ref{fig:pre-post-comparison} reports the results for mean table grades with bootstrapped confidence intervals. For both Claude and GPT, we find no statistically significant difference in performance between pre- and post-cutoff papers. If anything, post-cutoff performance is slightly higher. The absence of a pre/post difference suggests that pre-training leakage is unlikely to be the primary driver of performance.

Mechanistically, such leakage would also be difficult to exploit: any memorized results would need to be translated into executable code, and likely across programming languages, since the agents generate Python while the original replication packages use other languages. Nonetheless, these results should be interpreted with caution, as the sample size is small, and the selected papers have not undergone independent reproducibility verification. Moreover, this analysis does not rule out leakage arising during post-training (e.g., through fine-tuning or reinforcement learning on evaluation-relevant data).  Still, overall, this analysis supports the interpretation that the observed performance reflects genuine reimplementation rather than retrieval of memorized outputs.

\section{Conclusion}
\label{sec:discussion}

This paper establishes that AI agents can, in many cases, recover social science results from a paper's text and data alone. Across a diverse set of empirical papers, agents are able to reconstruct a large share of published results without access to the original code, demonstrating substantial progress toward automated reproducibility. While this first generation of agents delivers imperfect results, the diagnosed failure modes suggest a pattern. Going forward, an important bottleneck alongside model capability may be the way social science methods are documented. In many cases, discrepancies arise because key implementation details are underspecified or omitted entirely, forcing agents to infer choices (just as a human researcher would under the same constraints). As a result, the ability to reproduce results reflects a combination of both the agent's capabilities and the paper's clarity and completeness.

We began this paper by articulating the common view that the scientific paper is the source of truth for scientific claims. However, many of the discrepancies we observe arise from missing or ambiguous specifications, such as variable coding choices, data filters, or estimation procedures. In this light, it is worth reconsidering what we mean by the ``source of truth.'' For the purpose of reproducing results, that role is already fulfilled by the code, which provides a complete, machine-readable specification of the analysis pipeline and resolves ambiguities that the paper often leaves implicit. This suggests a clearer division of roles: code serves as the authoritative representation of \emph{what was done}, while the paper explains \emph{why} those choices were made, defining variables, motivating identification strategies, and interpreting results in context.% Failures of reproducibility in our setting frequently arise when these roles are conflated---when papers implicitly stand in for code without providing the level of precision that code requires.

This distinction has practical implications for how reproducibility is evaluated. Rather than expecting papers to fully encode implementation details, a more robust approach is to require explicit alignment between the narrative and the underlying code. Agentic systems provide a natural mechanism for this: they can translate papers into executable code, compare outputs, and surface inconsistencies between the two representations. Thus, automated reproduction can serve as both a tool for verification and a diagnostic, identifying where methods are underspecified, where implementations diverge, and where assumptions must be inferred. Aligning research practices with the requirements of both human and automated interpretation---through more explicit, structured, and complete method descriptions---may therefore play an important role in improving reproducibility at scale.

%In terms of the literature, our results contribute to the growing work on reproducibility and automated scientific evaluation. Prior work in both machine learning and the social sciences has largely focused on settings in which agents are given access to existing replication code \citep{hu2025reprobench,shah2026automating,xu2026scaling}. By contrast, we study reimplementation from methods descriptions and raw data alone, requiring agents to reconstruct the analysis pipeline without access to author-provided code. This distinction is both conceptual and methodological: by combining structured method extraction, deterministic cell-level evaluation, and systematic error attribution, our approach enables precise diagnosis of failures within the reproduction process. In doing so, it shifts the focus from whether results can be reproduced to understanding why they do or do not reproduce.

This perspective extends naturally to a broader spectrum of automated scientific tasks, each removing additional structure from the problem. Moving beyond reproduction with shared data raises a series of increasingly demanding questions. What if the data are unavailable, requiring agents to recover or reconstruct them? What if only the research question or hypothesis is given, and methods must be inferred rather than followed? What if agents are tasked not just with reproducing results, but with refining the analysis through specification checks, falsification of identification assumptions, or exploration of underlying mechanisms? And what if the goal shifts from reproduction to replication --- asking questions or applying methods with new data?  

While tackling these extensions could help address ongoing concerns about reproducibility in empirical research \citep[e.g.][]{brodeur2024mass,Brodeur2026-hk}, such progress would also involve pushing agentic systems further toward active participation in the research process. This shift requires us to define new criteria for scientific validity, including how to assess identification, robustness, and the reliability of conclusions generated without direct human oversight.

\clearpage

\paragraph{Code and data availability.}
This paper is an extended version for social scientists, derived from a version in computer science format forthcoming at EMNLP. The agentic pipeline and analysis code are available at \url{https://github.com/benjamin-kohler/social_science_replicability}; generated artifacts are archived on Zenodo \citep{kohler_zenodo_2026}. We provide pointers to the original papers and reproduction packages rather than redistributing them because their republication terms vary.

\bibliography{colm/colm2026_conference}
\bibliographystyle{colm/colm2026_conference}

% \section*{LLM usage}
% The conception and argumentation for this paper came entirely from the authors. All sections in the main text were originally drafted by humans, with occasional light revision for grammar and length by LLMs in some cases. Most figures and some tables were created with LLM assistance; the guardrails appendix also includes some AI-generated descriptions of code. On the side of coding, AI agents were used to assist in the creation of the pipeline and the analysis.

\clearpage 

\setcounter{table}{0}
\setcounter{figure}{0}
\setcounter{section}{0}
\renewcommand{\thetable}{A\arabic{table}}
\renewcommand{\thefigure}{A\arabic{figure}}
\renewcommand{\thesection}{\Alph{section}}

\appendix

\section{Main Appendix}

\subsection{Data overview}
\label{sec:appendix-data}

\begin{table}[htbp]
\centering
%\label{tab:sample_funnel}
\begin{tabular}{lr}
\toprule
Step & N \\
\midrule
I4R Replicate universe (unique papers) & 109 \\
Perfect computational reproduction & 59 \\
Sufficient or partial data availability & 54 \\
Has extractable tables & 48 \\
Final sample & 48 \\
\bottomrule
\end{tabular}
\caption{\textbf{Sample Construction.} Decomposition of papers from I4R.}
\label{tab:app-sample-funnel}
\end{table}

\begin{table}[htbp]
\centering
%\label{tab:language_distribution}
\begin{tabular}{lrrrr}
\toprule
Language & N papers & \% papers & Total LOC & Mean LOC \\
\midrule
Stata & 27 & 56.3\% & 217,123 & 8,042 \\
R & 13 & 27.1\% & 24,462 & 1,881 \\
MATLAB & 1 & 2.1\% & 5,520 & 5,520 \\
mixed & 7 & 14.6\% & 21,637 & 3,091 \\
\midrule
All & 48 & 100.0\% & 268,742 & 5,599 \\
\bottomrule
\end{tabular}
\caption{\textbf{Primary language of replication packages in the final sample.} LOC is a line-count across all code files.} \label{tab:overview-lang-loc}
\end{table}

\begin{footnotesize}
\begin{longtable}{p{1.8cm}p{8cm}rp{1.3cm}r}
\caption{Papers in Final Sample}
 \\
\toprule
Journal & Title & Year & Language & LOC \\
\midrule
\endfirsthead
\toprule
Journal & Title & Year & Language & LOC \\
\midrule
\endhead
APSR & Flight to Safety: COVID-Induced Changes in the Intensity of Status Quo Preference and Voting Behavior & 2021 & R & 2,372 \\
APSR & Campaign Contributions and Roll-Call Voting in the U.S. House of Representatives: The Case of the Sugar Industry & 2022 & Stata & 93 \\
APSR & Can't We All Just Get Along? How Women MPs Can Ameliorate Affective Polarization in Western Publics & 2022 & R & 716 \\
JOP & Not All Elections Are Created Equal: Election Quality and Civil Conflict & 2021 & Stata & 148 \\
JOP & Antinormative Messaging, Group Cues, and the Nuclear Ban Treaty & 2021 & R & 1,544 \\
JOP & Changing Tides: Public Attitudes on Climate Migration & 2021 & R & 3,720 \\
JOP & What Makes Anticorruption Punishment Popular? Individual-Level Evidence from China & 2021 & R & 1,328 \\
JOP & Black Workers in White Places: Daytime Racial Diversity and White Public Opinion & 2021 & R & 1,297 \\
EJ & Non-Linearities, State-Dependent Prices and the Transmission Mechanism of Monetary Policy & 2021 & MATLAB & 5,520 \\
EJ & Understanding Ethnolinguistic Differences: The Roles of Geography and Trade & 2021 & Stata & 1,362 \\
EJ & Gender Differences in Cooperative Environments? Evidence from the U.S. Congress & 2021 & Stata & 16,514 \\
EJ & Pre-Colonial Warfare and Long-Run Development in India & 2021 & Stata & 4,758 \\
EJ & Spillover Effects of Intellectual Property Protection in the Interwar Aircraft Industry & 2021 & Stata & 2,595 \\
EJ & Why Don't Firms Hire Young Workers During Recessions? & 2021 & Stata & 1,127 \\
EJ & The Wheels of Change: Technology Adoption, Millwrights and the Persistence in Britain's Industrialisation & 2022 & Stata & 1,240 \\
EJ & Peer Effects in Academic Research: Senders and Receivers & 2022 & mixed & 5,606 \\
 & The Power of Hydroelectric Dams: Historical Evidence from the United States over the Twentieth Century & 2022 & Stata & 7,598 \\
QJE & War, Socialism, and the Rise of Fascism: an Empirical Exploration & 2022 & Stata & 2,415 \\
REStud & Who Chooses Commitment? Evidence and Welfare Implications & 2021 & mixed & 4,823 \\
 & Exposure and Preferences: Evidence from Indian Slums & 2020 & R & 2,399 \\
AJPS & Re-Assessing Elite-Public Gaps in Political Behavior & 2020 & R & 1,731 \\
AJPS & Public Infrastructure and Economic Development: Evidence from Postal Systems & 2021 & mixed & 1,074 \\
AJPS & Hate Crimes and Gender Imbalances: Fears over Mate Competition and Violence against Refugees & 2021 & R & 3,068 \\
AJPS & Ascriptive Characteristics and Perceptions of Impropriety in the Rule of Law: Race, Gender, and Public Assessments of Whether Judges Can Be Impartial & 2021 & R & 2,118 \\
AJPS & Decentralization Can Increase Cooperation Among Public Officials & 2021 & mixed & 1,028 \\
AJPS & The Geography of Repression and Opposition to Autocracy & 2021 & Stata & 8,345 \\
AJPS & Talking Shops: The Effects of Caucus Discussion on Policy Coalitions & 2021 & R & 890 \\
AJPS & Parties as Disciplinarians: Charisma and Commitment Problems in Programmatic Campaigning & 2021 & mixed & 1,445 \\
AJPS & Indecent Disclosures: Anticorruption Reforms and Political Selection & 2021 & R & 1,075 \\
REStud & Immigration and Redistribution & 2023 & mixed & 5,287 \\
AER & Interaction, Stereotypes, and Performance: Evidence from South Africa & 2022 & Stata & 12,428 \\
AER & Vulnerability and Clientelism & 2022 & Stata & 5,412 \\
AER & Enabling or Limiting Cognitive Flexibility? Evidence of Demand for Moral Commitment & 2023 & Stata & 4,656 \\
AER & Market Access and Quality Upgrading: Evidence from Four Field Experiments & 2022 & Stata & 3,828 \\
AER & Evaluating Deliberative Competence: A Simple Method with an Application to Financial Choice & 2022 & Stata & 3,178 \\
AER & When a Doctor Falls from the Sky: The Impact of Easing Doctor Supply Constraints on Mortality & 2023 & Stata & 5,230 \\
AER & Can Technology Solve the Principal-Agent Problem? Evidence from China's War on Air Pollution & 2022 & Stata & 1,864 \\
AER: Insights & Wage Cyclicality and Labor Market Sorting & 2022 & Stata & 16,880 \\
AEJ: Applied & Assortative Matching at the Top of the Distribution: Evidence from the World's Most Exclusive Marriage Market & 2022 & Stata & 12,957 \\
AEJ: Applied & Historical Lynchings and the Contemporary Voting Behavior of Blacks & 2022 & Stata & 1,432 \\
 & How Effective Are Monetary Incentives to Vote? Evidence from a Nationwide Policy & 2021 & Stata & 7,523 \\
AEJ: Macro & Declining Worker Turnover: The Role of Short-Duration Employment Spells & 2021 & mixed & 2,374 \\
AEJ: Policy & Mortality, Temperature, and Public Health Provision: Evidence from Mexico & 2022 & Stata & 13,170 \\
AEJ: Policy & Multinationals' Sales and Profit Shifting in Tax Havens & 2022 & Stata & 7,081 \\
AEJ: Policy & School Spending and Student Outcomes: Evidence from Revenue Limit Elections in Wisconsin & 2022 & Stata & 10,813 \\
AEJ: Policy & The Long-Run Effects of Sports Club Vouchers for Primary School Children & 2022 & Stata & 29,492 \\
AEJ: Policy & How Do Beliefs about the Gender Wage Gap Affect the Demand for Public Policy? & 2022 & Stata & 34,984 \\
AJPS & Entertaining Beliefs in Economic Mobility & 2022 & R & 2,204 \\
\bottomrule
\end{longtable}

\label{tab:sample_papers} 

\end{footnotesize}

\begin{table}[ht]
\centering
\resizebox{\columnwidth}{!}{
\begin{tabular}{llll}
\toprule
\textbf{Step} & \textbf{Name} & \textbf{Inputs} & \textbf{Outputs} \\
\midrule
1 & Extraction &
  \begin{tabular}[t]{@{}l@{}}
    Paper PDF \\
    Reproduction Package
  \end{tabular} &
  \begin{tabular}[t]{@{}l@{}}
    Raw Data \\
    Extracted Methods \\
    Original Table \\
    Table Template
  \end{tabular} \\
\midrule
2 & Reimplementation &
  \begin{tabular}[t]{@{}l@{}}
    Raw Data \\
    Extracted Methods \\
    Table Template
  \end{tabular} &
  \begin{tabular}[t]{@{}l@{}}
    Script\textsubscript{\textit{i}}.py \\
    Table\textsubscript{\textit{i}}
  \end{tabular} \\
\midrule
3 & Evaluation &
  \begin{tabular}[t]{@{}l@{}}
    Table\textsubscript{\textit{i}} \\
    Original Table
  \end{tabular} &
  \begin{tabular}[t]{@{}l@{}}
    Comparison
  \end{tabular} \\
\midrule
4 & Explanation &
  \begin{tabular}[t]{@{}l@{}}
    Paper PDF \\
    Reproduction Package \\
    Extracted Methods \\
    Script\textsubscript{\textit{i}}.py \\
    Comparison
  \end{tabular} &
  \begin{tabular}[t]{@{}l@{}}
    Discrepancies \& \\
    \quad Error Sources
  \end{tabular} \\
\bottomrule
\end{tabular}
}
\caption{Inputs and outputs for each step of the reproduction pipeline.}
\label{tab:pipeline_in_out}
\end{table}

\clearpage 

\subsection{Reproduction of Figures}
\label{sec:appendix-figures}

Our pipeline handles both tables and figures as reproduced outputs. For figures, the extractor generates a detailed description of the axes and plot structure without referencing any underlying data or results, and produces a simple Python template that recreates the plot and its styling from a data object (e.g., a dataframe). This design allows the reimplementation agent to focus on reconstructing the data while avoiding exposure to the original plot and minimizing the risk of result leakage, while maintaining stylistic consistency across runs.

\begin{figure}[ht]
    \centering
    \begin{subfigure}[t]{0.48\textwidth}
        \centering
        \includegraphics[width=\textwidth, height=10cm, keepaspectratio]{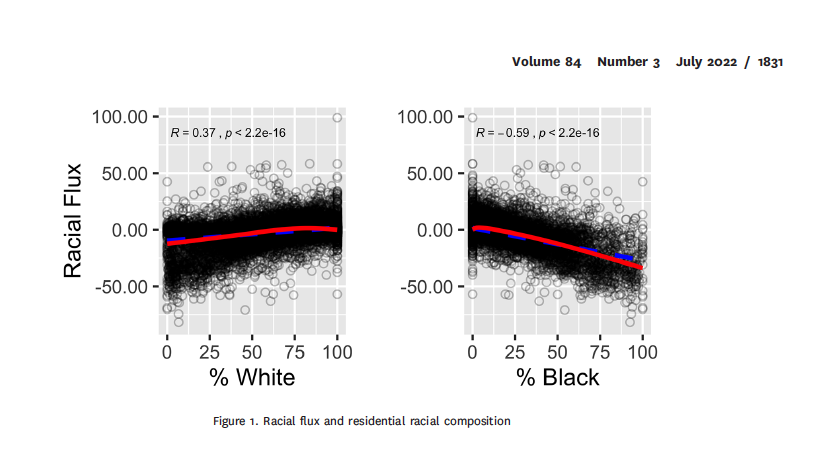}
        \caption{Original Figure 1}
    \end{subfigure}
    \hfill
    \begin{subfigure}[t]{0.48\textwidth}
        \centering
        \includegraphics[width=\textwidth, height=10cm, keepaspectratio]{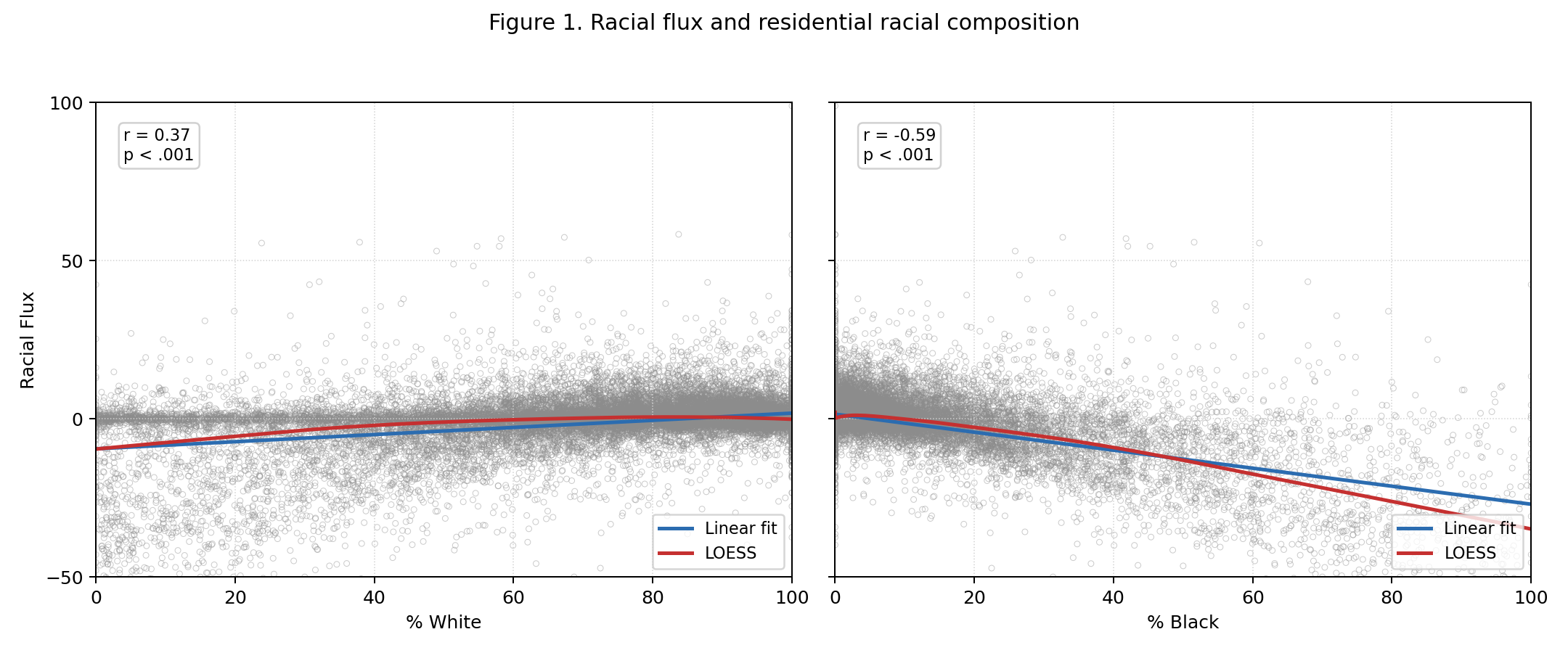}
        \caption{Reproduced Figure 1}
    \end{subfigure}
    
    \vspace{0.5cm}
    
    \begin{subfigure}[t]{0.48\textwidth}
        \centering
        \includegraphics[width=\textwidth, height=6cm, keepaspectratio]{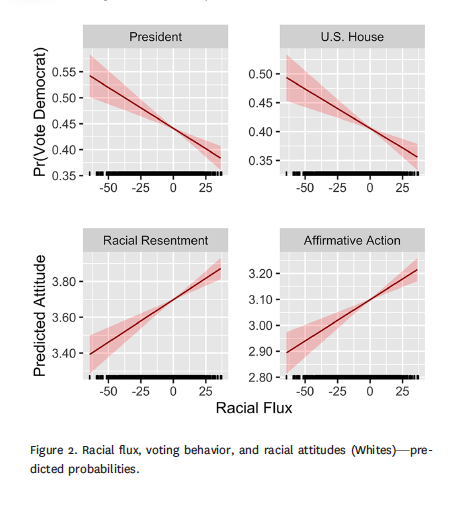}
        \caption{Original Figure 2}
    \end{subfigure}
    \hfill
    \begin{subfigure}[t]{0.48\textwidth}
        \centering
        \includegraphics[width=\textwidth, height=6cm, keepaspectratio]{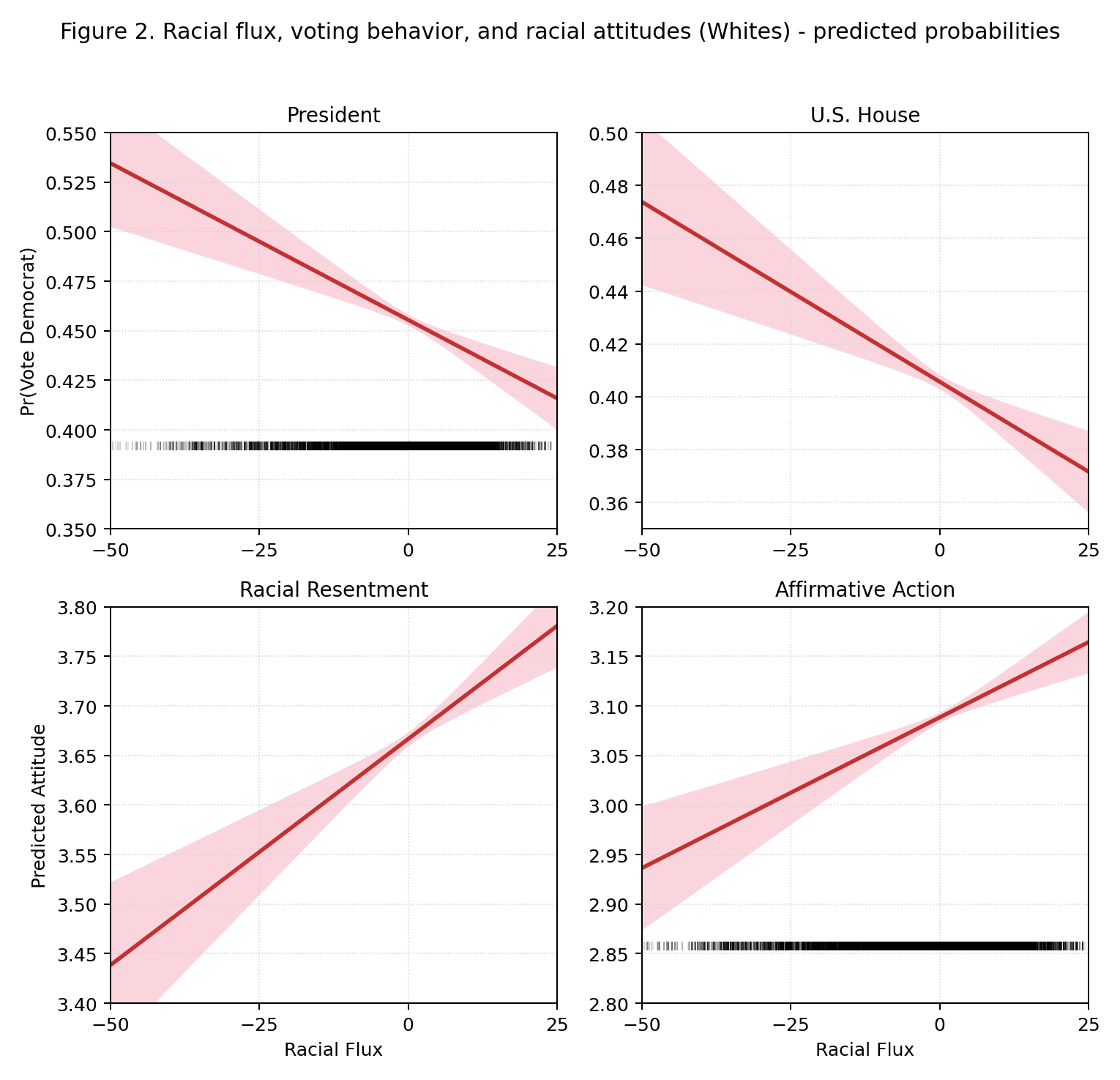}
        \caption{Reproduced Figure 2}
    \end{subfigure}
    \caption{\textbf{Agentic reproduction of figure results. }Comparison between original and reproduced figures from the paper "Black Workers in White Places: Daytime Racial Diversity and White Public Opinion." The reproduction was conducted by OpenCode GPT-5.4.}%The reproducing agent received a Python snippet to recreate the plot's styling without any actual data.}
    \label{fig:app-figures}
\end{figure}

Evaluating reproduced figures is inherently more complex than comparing tables, as it requires assessing overall trends, individual data points, axes, and other visual features. Automating this process therefore relies on visual language models. However, whether such models can provide fair and consistent evaluations remains an open question. For this reason, we do not report quantitative results for figures. Figure~\ref{fig:app-figures} illustrates a representative example from GPT-5.4 in OpenCode, demonstrating the feasibility of the approach: the reproduced figures capture the main trends of the originals, though some visual discrepancies remain.

\clearpage 

\subsection{Details on A-F Grading}
\label{app:grading}

Tables are evaluated through a comparison of aligned original and replicated outputs. First, cells are aligned across the two tables and replicated values may be rescaled when the comparison detects an approximate power-of-ten mismatch. Grades are assigned from the resulting cell-level differences. Because percentage differences are unstable when the original value is close to zero, the grading rule distinguishes between near-zero and non-near-zero cases. The grading rubric is outlined in Table~\ref{tab:grading_detailed} 

\begin{table}[htbp]
\centering
\small
\begin{tabularx}{\textwidth}{>{\raggedright\arraybackslash}p{1.6cm} >{\raggedright\arraybackslash}X >{\raggedright\arraybackslash}X}
\toprule
Grade & Near-zero original ($|x| < 0.001$) & Otherwise \\
\midrule
A & Absolute difference $< 0.002$; also assigned when both values are exactly zero & Percentage difference $< 2\%$ \\
B & Absolute difference $< 0.02$ & Percentage difference $< 20\%$ \\
C & Absolute difference $< 0.05$ & Percentage difference $< 40\%$ \\
D & Absolute difference $< 0.1$ & Percentage difference $< 60\%$ \\
E & Absolute difference $\geq 0.1$, or signs differ & Percentage difference $\geq 60\%$, or signs differ \\
F & \multicolumn{2}{l}{Assigned if either the original or replicated value is missing} \\
\bottomrule
\end{tabularx}
\caption{Cell-level grading rules}
\label{tab:grading_detailed}
\end{table}

Table-level grades are computed by converting cell grades to numeric values ($A=5$, $B=4$, $C=3$, $D=2$, $E=1$, $F=0$), averaging over non-$F$ cells only, and then mapping the average back to a letter grade using fixed thresholds: $[4.5,5] \rightarrow A$, $[3.5,4.5) \rightarrow B$, $[2.5,3.5) \rightarrow C$, $[1.5,2.5) \rightarrow D$, and $[0.5,1.5) \rightarrow E$. If all cells are graded $F$, the table grade is $F$. Paper-level grades are computed analogously, averaging across item grades using the same numeric mapping and the same thresholds, while excluding items marked unverifiable, items flagged as judge errors, and items graded $F$. 

Figures are evaluated separately through a vision-capable language model and use the grading rubric presented in Figure~\ref{tab:figure_grading_rubric}.

\begin{table}[htbp]
\centering
\begin{tabular}{ll}
\toprule
Grade & Criterion \\
\midrule
A & Visually indistinguishable \\
B & Patterns match, visible differences \\
C & Recognizable, noticeable gaps \\
D & Substantially different \\
E & Fundamentally different \\
F & Missing, blank, or not verifiable \\
\bottomrule
\end{tabular}
\caption{Figure grading rubric}
\label{tab:figure_grading_rubric}
\end{table}

%\subsubsection{Aggregated plots including F-graded items}

\begin{figure}
    \centering
    \begin{subfigure}[t]{0.48\textwidth}
        \centering
        \includegraphics[width=\textwidth]{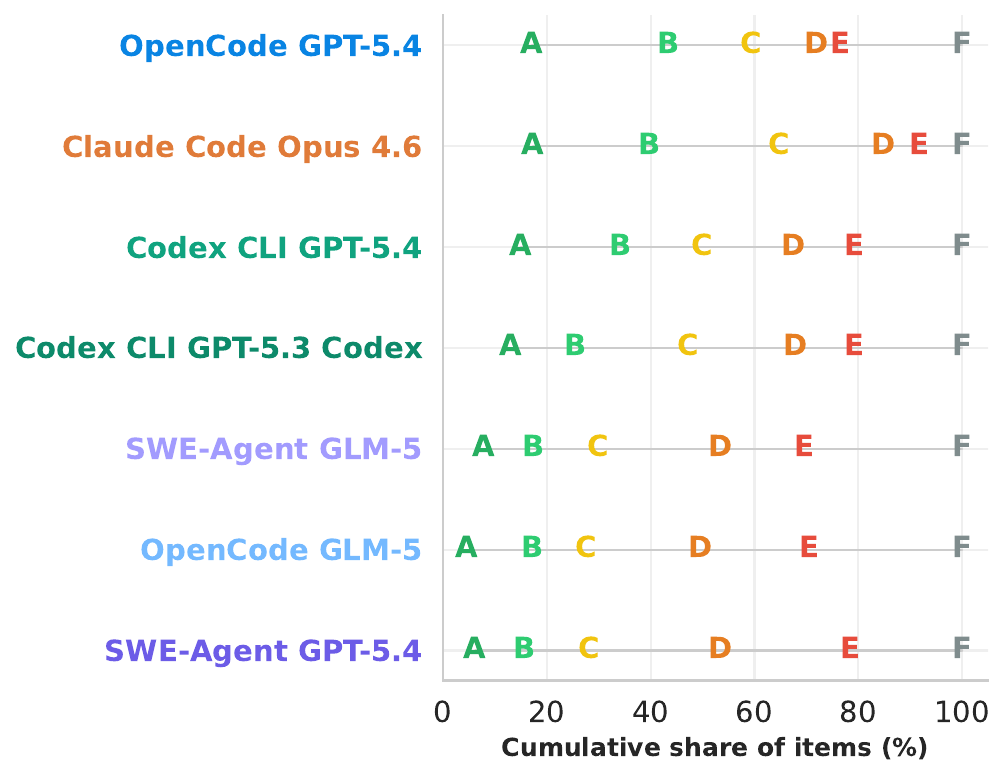}
        \caption{Table-level grades including all F-graded cells.}
    \end{subfigure}
    \hfill
    \begin{subfigure}[t]{0.48\textwidth}
        \centering
        \includegraphics[width=\textwidth]{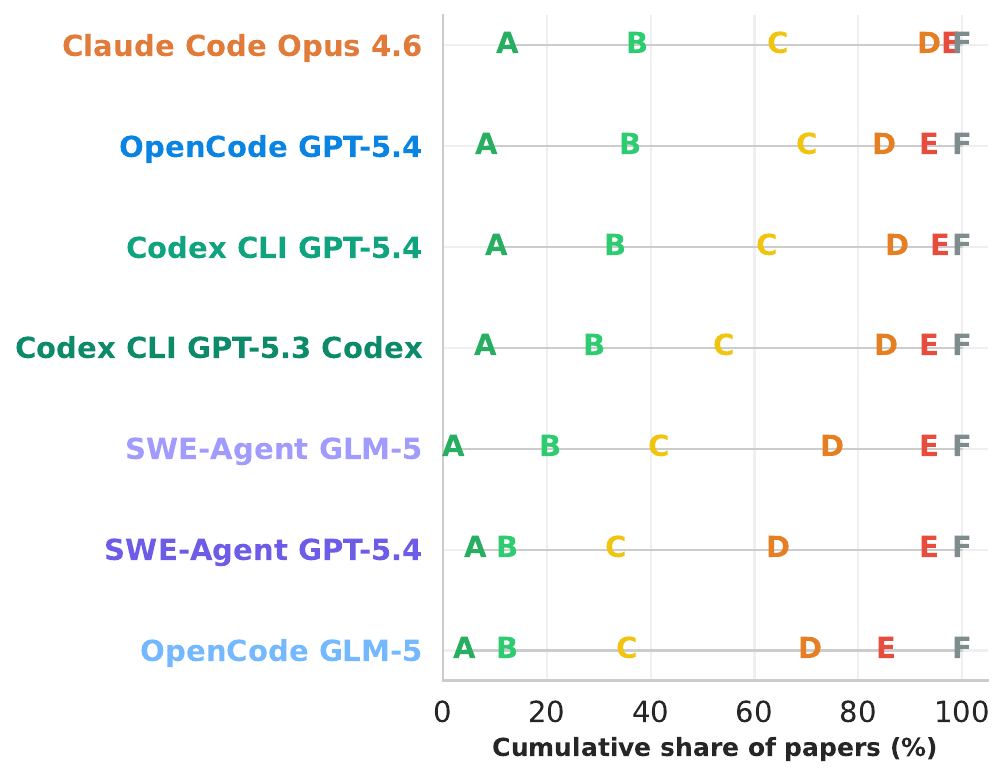}
        \caption{Paper-level grades including all F-graded tables.}
    \end{subfigure}
    \vspace{12pt}
    \begin{subfigure}[t]{0.48\textwidth}
        \centering
        \includegraphics[width=\textwidth]{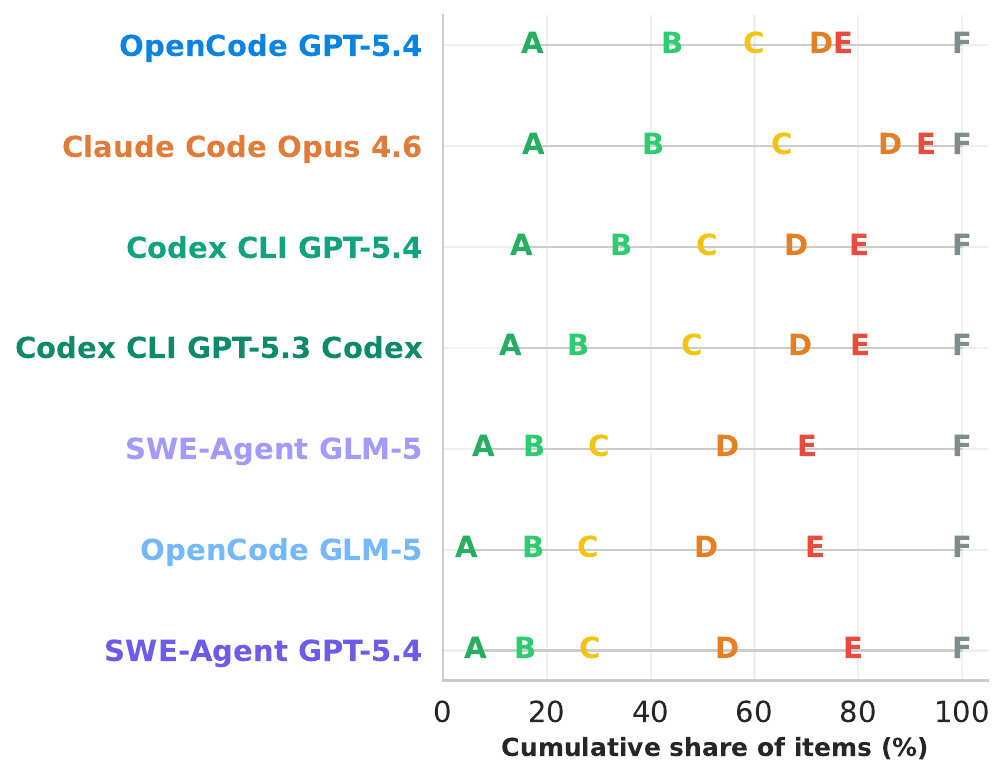}
        \caption{Table-level grades including F-grades for cells with at least one agent with a non-F result.}
    \end{subfigure}
    \hfill
    \begin{subfigure}[t]{0.48\textwidth}
        \centering
        \includegraphics[width=\textwidth]{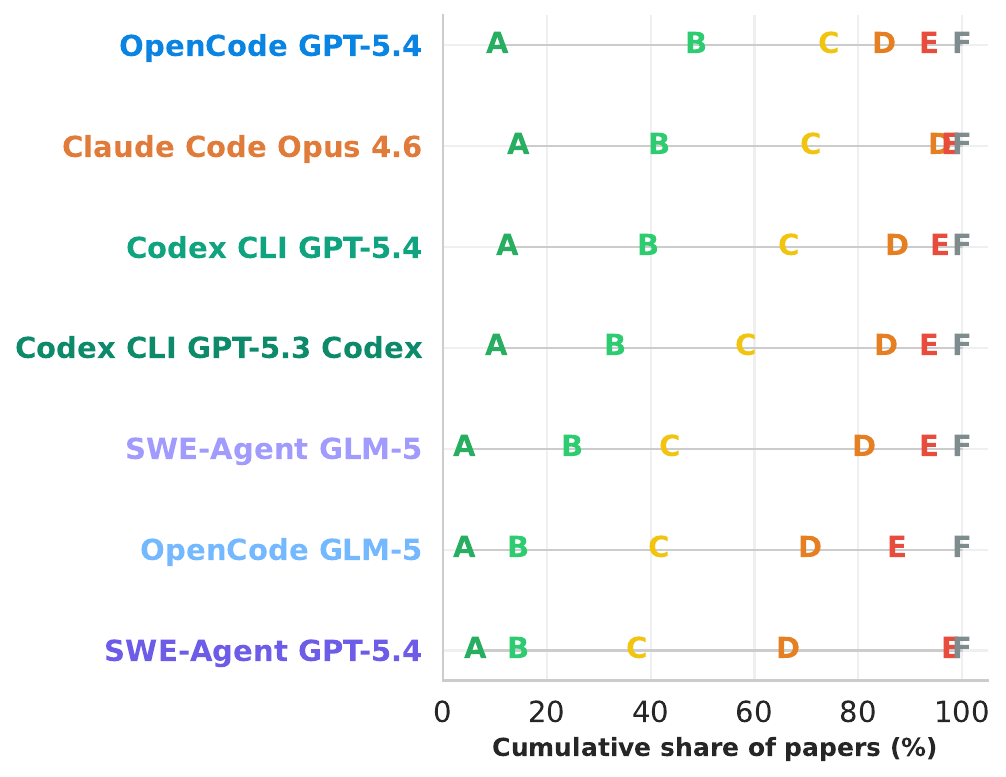}
        \caption{Paper-level grades including F-grades for tables with at least one agent with a non-F result.}
    \end{subfigure}
    \caption{\textbf{Table-level grades including F-grades.} Aggregated performance naturally shrinks when including empty/not-reproduced tables/items. Notably the best-performing model/scaffold also switches on the paper level.}
    \label{fig:app-agg-grades-with-f}
\end{figure}

\clearpage 

\subsection{Additional Figures}

\begin{figure}[ht]
    \centering
        \includegraphics[width=.7\textwidth]{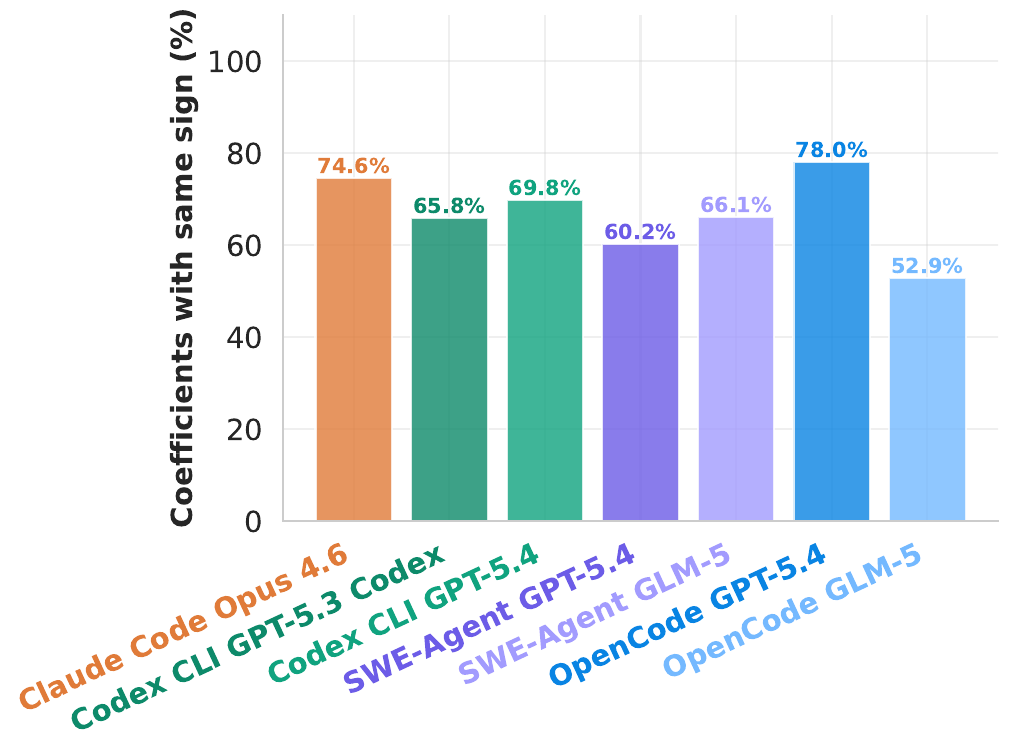}
        
        \caption{\textbf{Share with correct sign, including missings}. Share of reproduced coefficients with the same sign as the original paper including coefficients that are not being replicated in the denominator.}

    \label{fig:coeff-results-with-missing}
\end{figure}

\begin{figure}[ht]
    
        \centering
        \includegraphics[width=.7\textwidth]{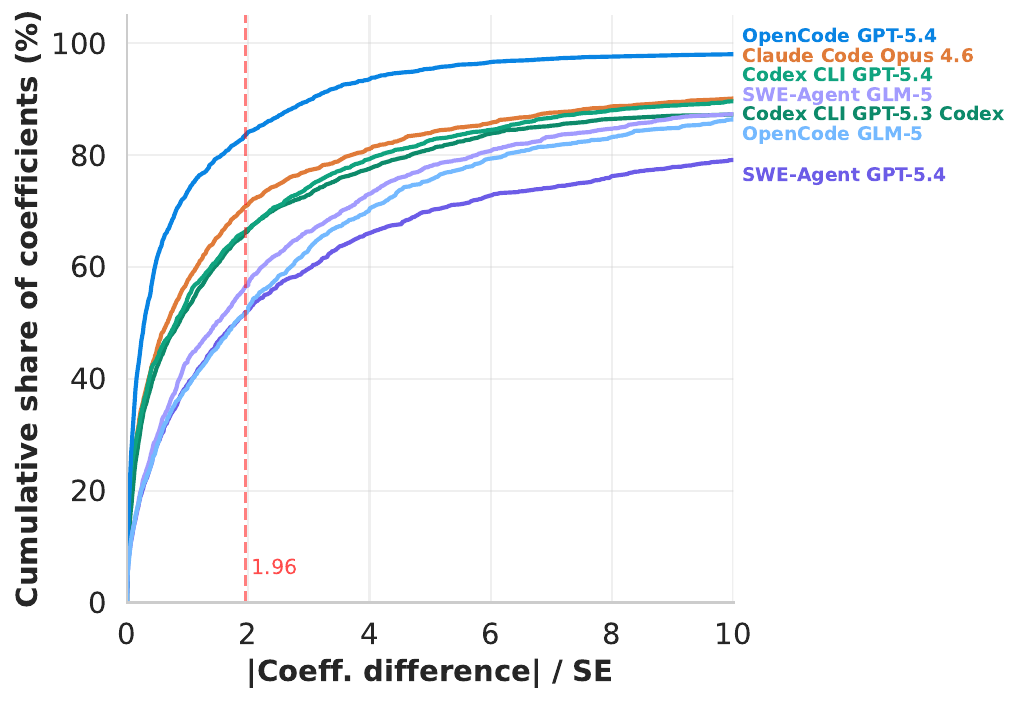}
        \caption{\textbf{CDF of SE-adjusted coeff difference, excluding descriptive stats tables}. Cumulative distribution of $|$coefficient difference$|$ / SE across approaches when removing descriptive tables. The dashed red line marks 1.96 (95\% CI threshold. The magnitudes and broad ranking are similar to the unrestricted comparison).}

    \label{fig:app-cdf-wo-descriptives}
\end{figure}

\begin{figure}[ht]
    \centering
    \includegraphics[width=0.95\textwidth]{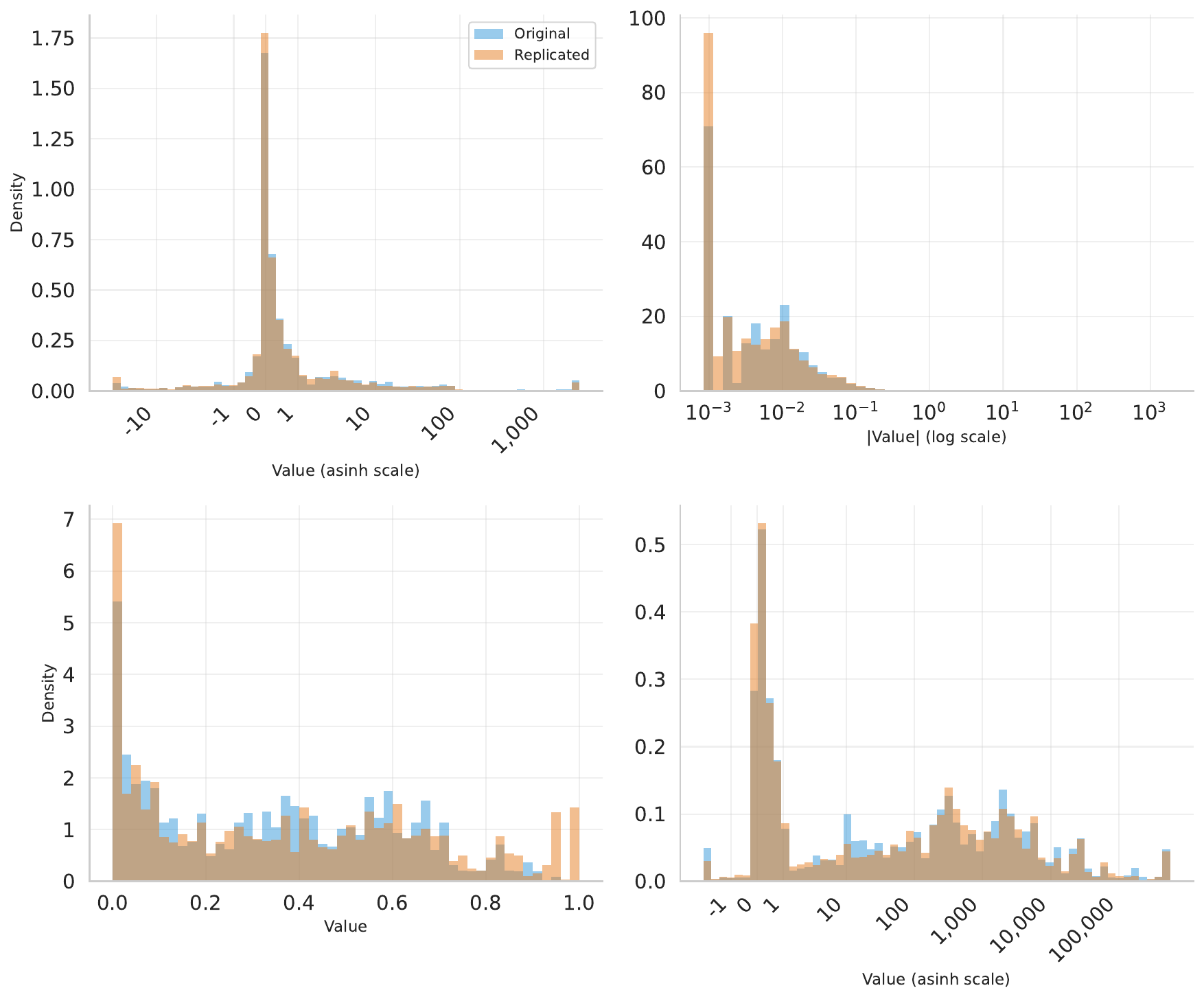}
    \caption{\textbf{Histograms of original and reproduced statistics.} Across various transformations of the statistics, the distributions of the original and reproduced statistics are quite similar.}
    \label{fig:app-hist-agg}
\end{figure}

\begin{figure}[ht]
    \centering
    \includegraphics[width=\textwidth]{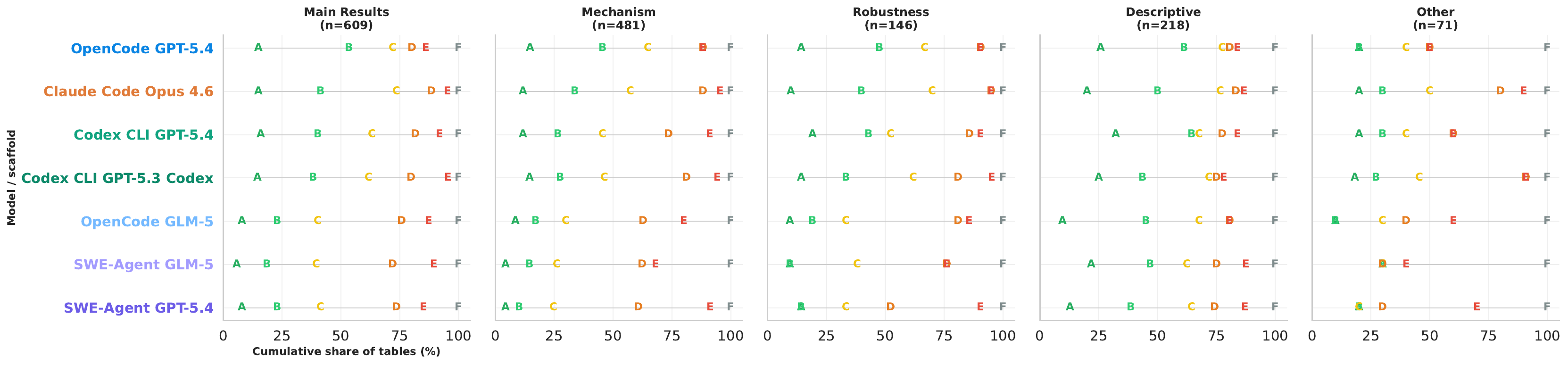}
    \caption{\textbf{Grade distributions by table type.} Including all F-grades.}
    \label{fig:app-table-grade-by-type}
\end{figure}

\clearpage 

\subsection{Computational effort and reproduction performance}
\label{sec:appendix-effort}

\begin{table}[h]
\centering
\caption{Paper grade and completion tokens, within paper (run level). OLS; outcome $=$ the run's overall paper grade on a 0--5 scale (A=5, \ldots, F=0; all\_f grading), one observation per paper $\times$ scaffold run. Column (1): tokens and paper fixed effects only — the pooled within-paper association. Column (2) adds scaffold/model dummies (reference: Claude Code Opus 4.6). Tokens in $\log_2$: coefficients read as grade points per doubling. Common telemetry sample. Effort is agent-chosen: coefficients are descriptive, not causal. SEs clustered by paper in parentheses. $^{*}p<0.1$, $^{**}p<0.05$, $^{***}p<0.01$.}
\label{tab:papergrade_tokens_contrast}
\begin{tabular}{lcc}
\toprule
 & (1) & (2) \\
\midrule
log$_2$(completion tokens) & 0.17*** & 0.10* \\
 & (0.06) & (0.06) \\
\addlinespace
Codex CLI GPT-5.3 Codex &  & -0.24* \\
 &  & (0.14) \\
Codex CLI GPT-5.4 &  & -0.06 \\
 &  & (0.14) \\
SWE-Agent GPT-5.4 &  & -0.77*** \\
 &  & (0.16) \\
SWE-Agent GLM-5 &  & -0.63*** \\
 &  & (0.16) \\
OpenCode GPT-5.4 &  & -0.20 \\
 &  & (0.13) \\
OpenCode GLM-5 &  & -0.97*** \\
 &  & (0.14) \\
\midrule
Paper FE & Yes & Yes \\
Scaffold/model FE & No & Yes \\
Observations & 333 & 333 \\
R$^2$ & 0.698 & 0.771 \\
Mean dep.\ var.\ & 2.54 & 2.54 \\
\bottomrule
\end{tabular}
\end{table}

Table~\ref{tab:papergrade_tokens_contrast} relates reproduction performance to token usage within papers. We regress the run-level paper grade (0--5) on the log$_2$ of completion tokens, with one observation per paper~$\times$~agent run and standard errors clustered by paper; column (1) includes paper fixed effects only, while column (2) adds scaffold/model fixed effects. Conditional on the paper, token usage correlates positively with performance: a doubling of completion tokens is associated with a 0.17 grade-point higher paper grade. Conditional on both paper and scaffold/model, however, the association is smaller, at 0.10 grade points, and is marginally significant ($p=0.099$). This leaves weak evidence of a within-agent association; much of the pooled correlation still reflects \emph{which} agent is spending rather than the spending itself. We stress that this evidence is purely correlational, since token usage is endogenous to the model and scaffold. A causal investigation would need to vary token usage experimentally within agents, e.g., by testing different hyperparameter settings---such as reasoning-effort levels or explicit token budgets---against each other.

\subsection{Multi-run Stability}
\label{sec:appendix-stability}

Based on 20 random papers re-run two times (three runs in total), in two non-overlapping cohorts of ten papers. The method-extractor and alternate-judge analyses use the first ten-paper cohort.

\begin{figure}[ht]
    \centering
    \begin{subfigure}[t]{0.48\textwidth}
       \centering
    \includegraphics[width=\textwidth]{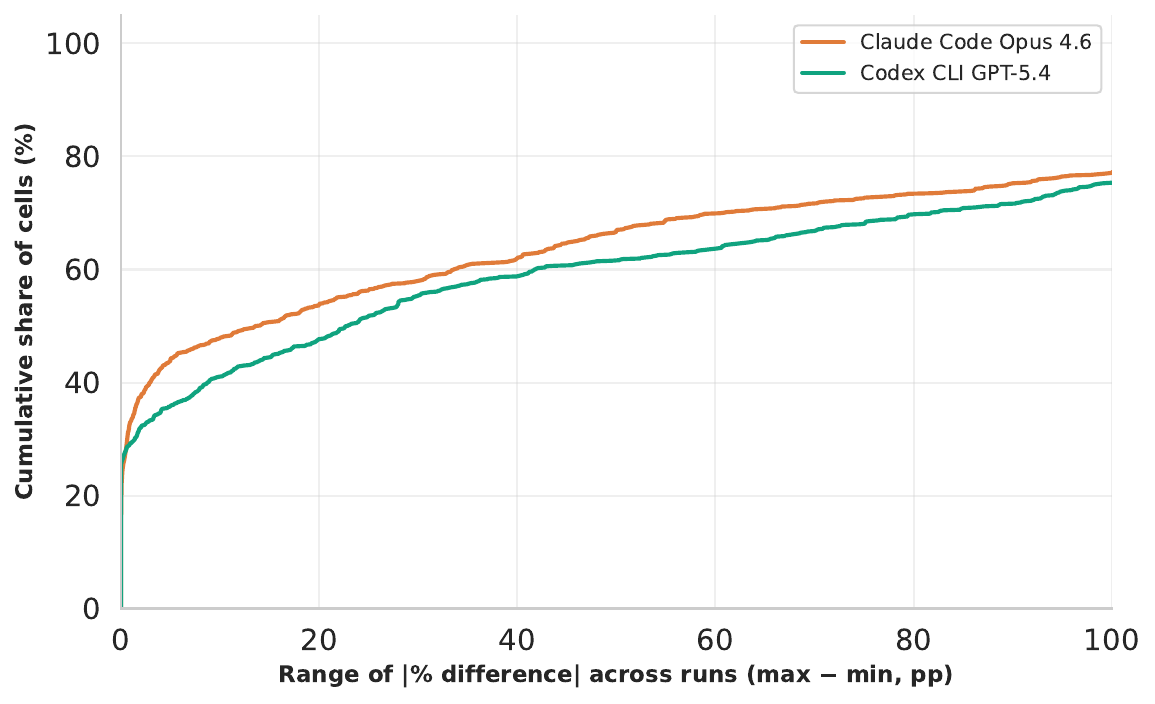}
    \caption{This plot shows how the percentage difference to the original value varies between runs for each cell. For the three runs the percentage difference to the original is computed. The plot shows the cumulative distribution of the range of these percentage distributions. It can be interpreted as "For y\% of cells the difference between the \textit{best} and the \textit{worst} reproduction attempt has been at most x\%"}
    \label{fig:app-run-stability-cell-pct-diff}
        \end{subfigure}
    \hfill
    \begin{subfigure}[t]{0.48\textwidth}
          \centering
    \includegraphics[width=\textwidth]{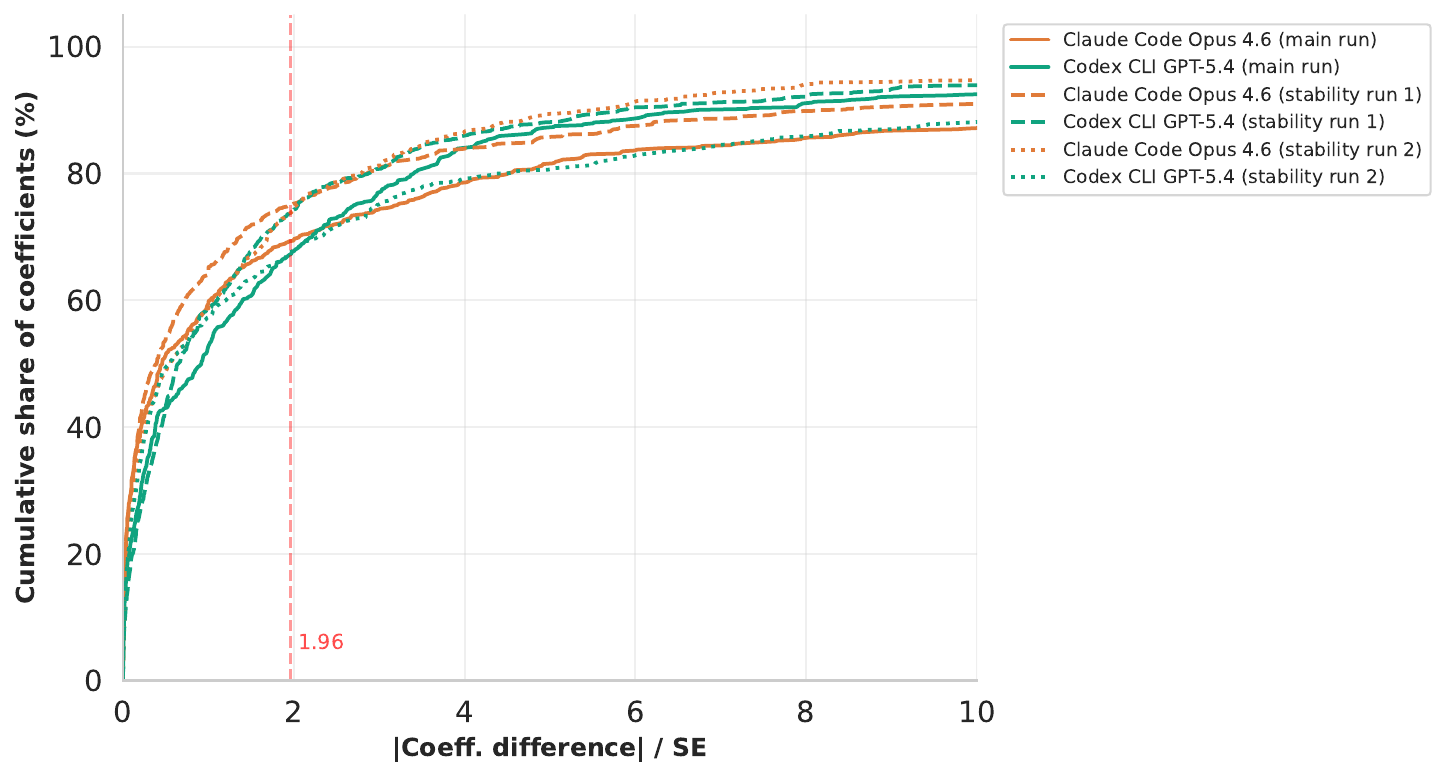}
    \caption{Distribution of t-statistics between runs.}
    \label{fig:app-run-stability-t-stat-cdf}
    \end{subfigure}

    % \begin{subfigure}[t]{0.48\textwidth}
    %     \centering
    %         \includegraphics[width=\textwidth]{results/run_stability/cell_pct_diff_range_cdf.pdf}
    % \caption{This plots shows for how many papers the percentage difference to the original value varies between run. For the three runs the percentage difference to the original is computed. The plot shows the cumulative distribution of the range of these percentage distributions. It can be interpreted as "For y\% of cells the difference between the \textit{best} and the \textit{worst} reproduction attempt has been at most x\%"}

    % \end{subfigure}
    \hfill
    \begin{subfigure}[t]{0.48\textwidth}
  \centering
    \includegraphics[width=\textwidth]{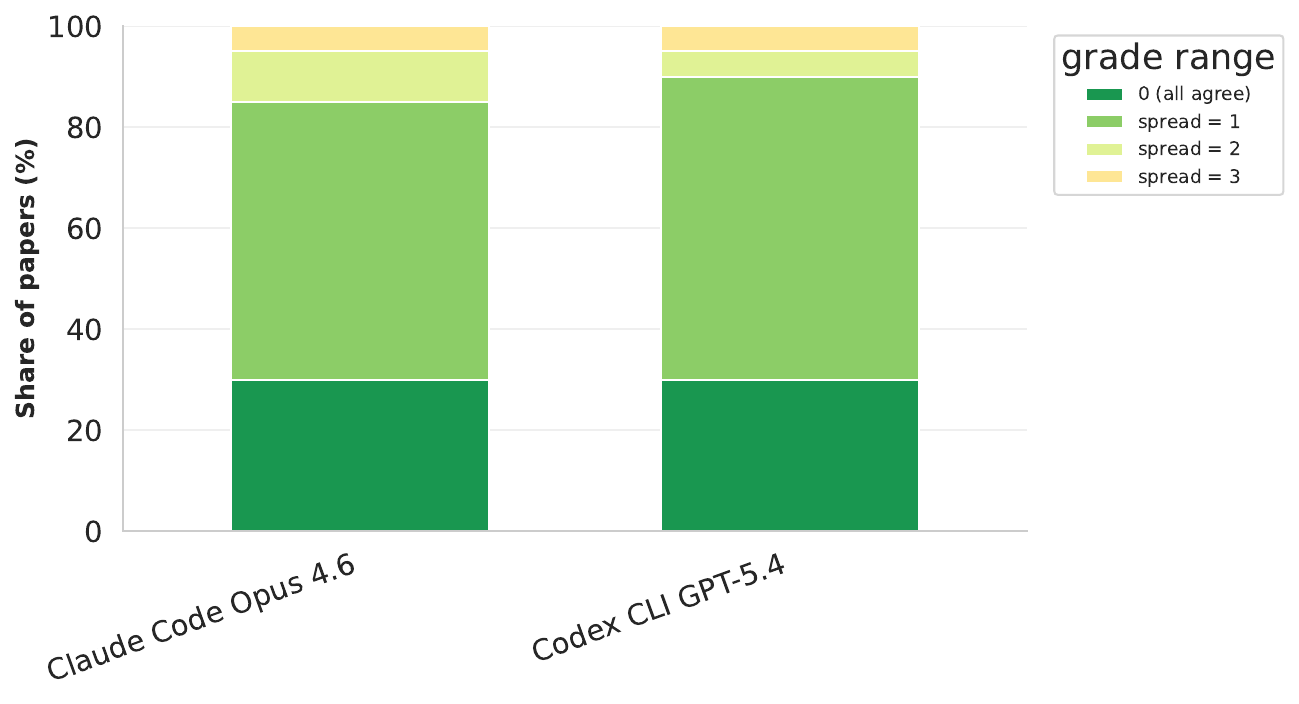}
    \caption{Distribution of paper grade differences.}
    \label{fig:app-run-stability-paper-diff}
    \end{subfigure}

        \caption{Additional statistics on stability between runs}

\end{figure}

\clearpage

\begin{figure}%[ht]
    \centering
    \includegraphics[width=0.95\textwidth]{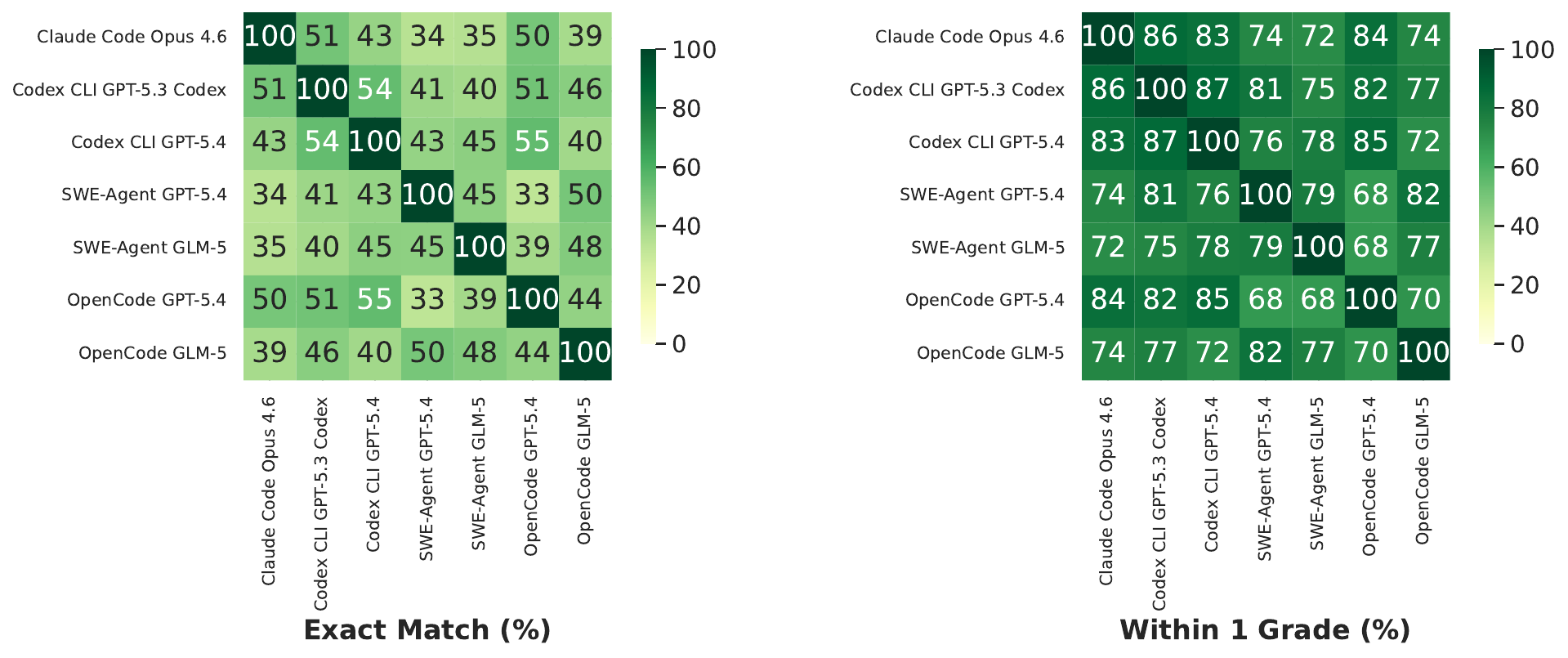}
    \caption{Inter-agent agreement at the paper level. Left: share of papers for which two agents assign the same grade. Right: share of papers for which grades are within one step.}
    \label{fig:app-aggreement}
\end{figure}

\subsection{Method extractor variant}
\label{sec:appendix-extractor-variant}

\begin{figure}[t]
    \centering
    \begin{subfigure}[t]{0.52\textwidth}
       \centering
       \includegraphics[width=\linewidth]{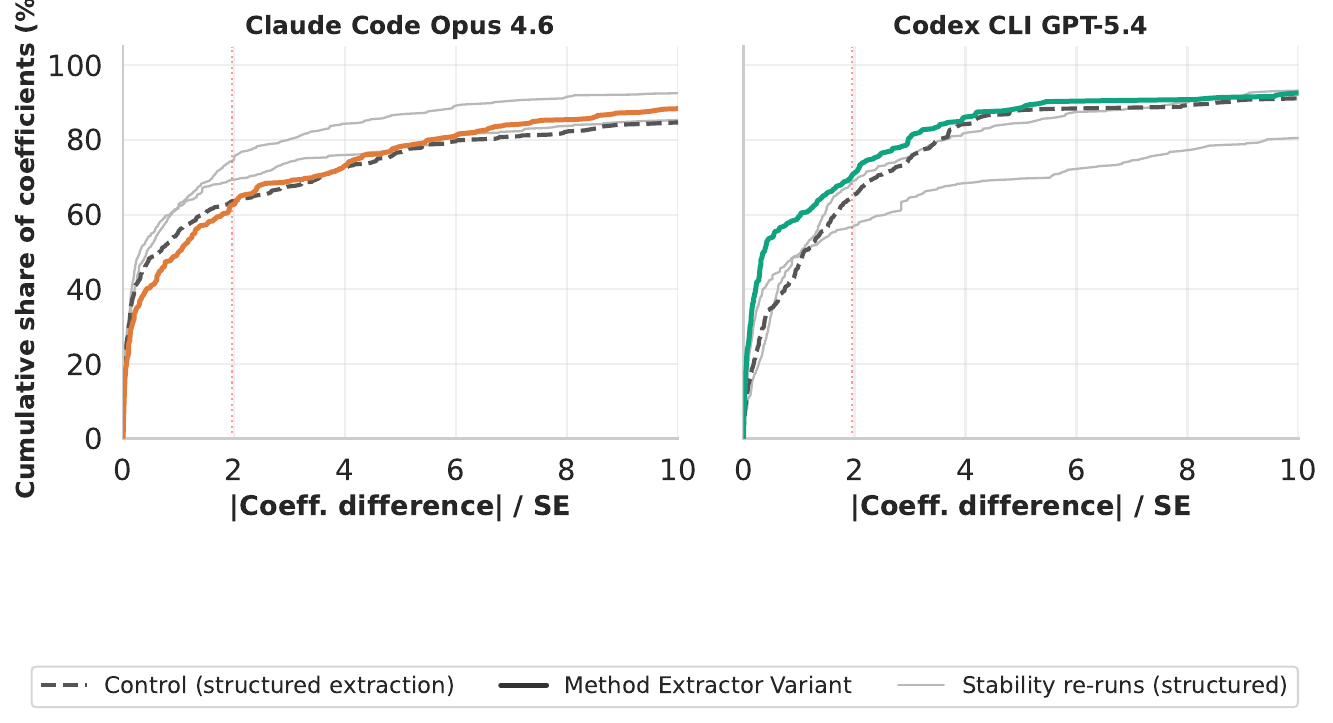}
       \caption{Coefficient-level agreement: cumulative distribution of absolute coefficient differences normalized by the original standard error, for the variant, the structured control, and the stability re-runs.}
       \label{fig:extractor-variant-cdf}
    \end{subfigure}
    \hfill
    \begin{subfigure}[t]{0.46\textwidth}
       \centering
       \includegraphics[width=\linewidth]{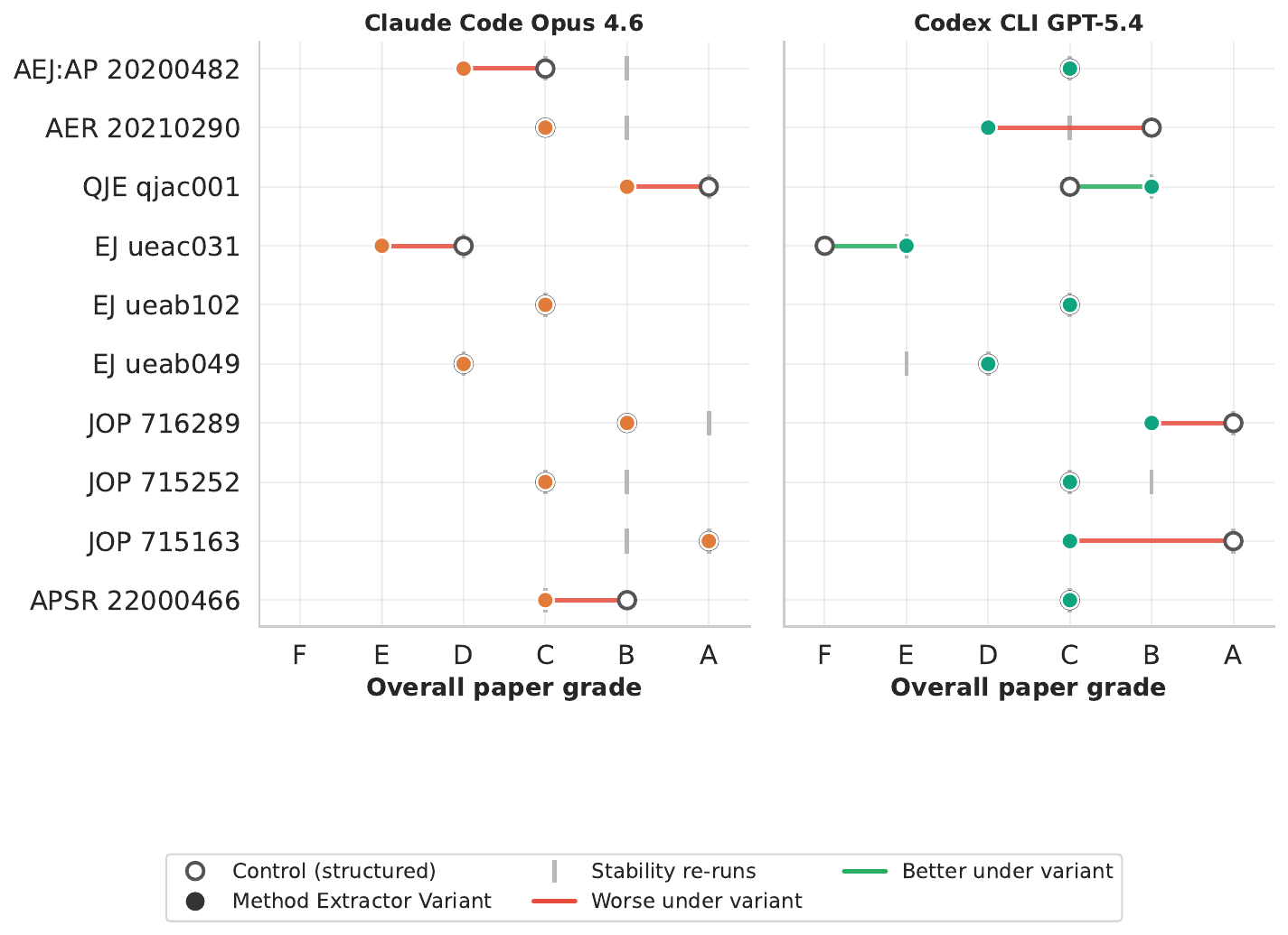}
       \caption{Paper-level grades: control (structured) vs.\ variant (markdown) grade per paper, with tick marks showing the range across stability re-runs.}
       \label{fig:extractor-variant-dumbbell}
    \end{subfigure}
\caption{\textbf{Sensitivity of reproduction to the method extractor.} \textit{The full pipeline is re-run on a ten-paper subset of the multi-run stability sample, holding everything fixed except the method-extraction prompt: the structured specification (control, Prompt~\ref{prompt:extraction}) is replaced with a free-form Markdown methods document (variant, Prompt~\ref{prompt:extraction-markdown}). The stability re-runs of the structured extractor provide the run-to-run noise reference. Coefficient-level agreement (panel a) is essentially unchanged, while paper grades (panel b) shift slightly downward but within re-run noise.}}
\label{fig:extractor-variant}
\end{figure}

To probe how sensitive reproduction outcomes are to the upstream method extraction, we re-ran the full pipeline on a ten-paper subset of our 20-paper multi-run stability sample (Appendix~\ref{sec:appendix-stability}), using Claude Code (Opus 4.6) and Codex (GPT-5.4). Every component of the pipeline---the reproduction agents, the blinded table templates, the judge, and the deterministic grader---was held fixed; only the method-extraction prompt was changed. Whereas the main extractor emits a structured specification with exact row and column labels, per-column regression specifications, and explicit estimating equations (Prompt~\ref{prompt:extraction}), this variant instead produces a free-form Markdown methods document that describes the data preparation and analyses in prose at the granularity of a well-written methods section, deliberately omitting exact table structure and per-column specifications (Prompt~\ref{prompt:extraction-markdown}). At the coefficient level, the variant is largely indistinguishable from the structured extractor and stays within the envelope of re-run noise (Figure~\ref{fig:extractor-variant}, panel~a). At the paper-grade level, the coarser description costs on average roughly a third of a grade step---a small, directionally downward shift that remains within the spread observed across repeated runs (Figure~\ref{fig:extractor-variant}, panel~b).

\subsection{Correlates of reproduction performance}
\label{sec:appendix-determinants}
We analyze potential sources of performance variation across papers and agents. Figure~\ref{fig:app-aggreement}  shows how frequently two reproducing agents assign the same grade (left panel) or grades within one step of each other (right panel) for a given paper. This indicates that agents---particularly those developed by the same organization---tend to agree more than would be expected by chance, though between-agent performance nonetheless varies substantially.

\begin{figure}%[ht]
    \centering
    \includegraphics[width=0.6\textwidth]{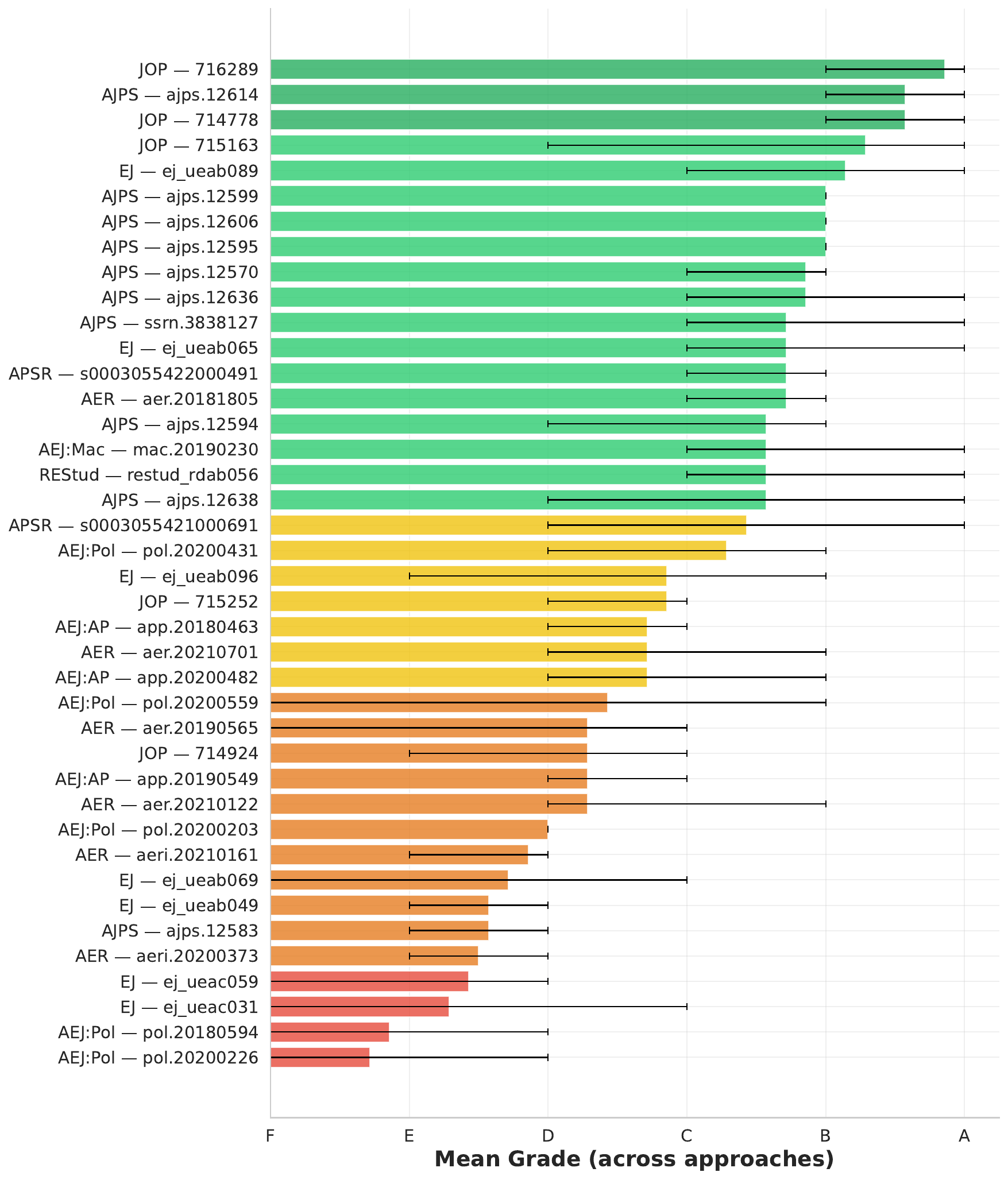}
    \caption{Within-paper performance range across all agent--model combinations, sorted by mean grade. Each bar spans the worst to best grade assigned by any agent.}
    \label{fig:app-paper-difficulty}
\end{figure}

Figure~\ref{fig:app-paper-difficulty} offers a complementary perspective by displaying the within-paper performance range, which can also be interpreted as a measure of reproduction difficulty. The range varies across papers but remains within two grade steps for the majority, highlighting that some papers are inherently harder to reproduce than others, while differences in agentic performance remain meaningful throughout.

\begin{figure}[ht]
    \centering
    \begin{subfigure}[t]{0.48\textwidth}
        \centering
        \includegraphics[width=\textwidth]{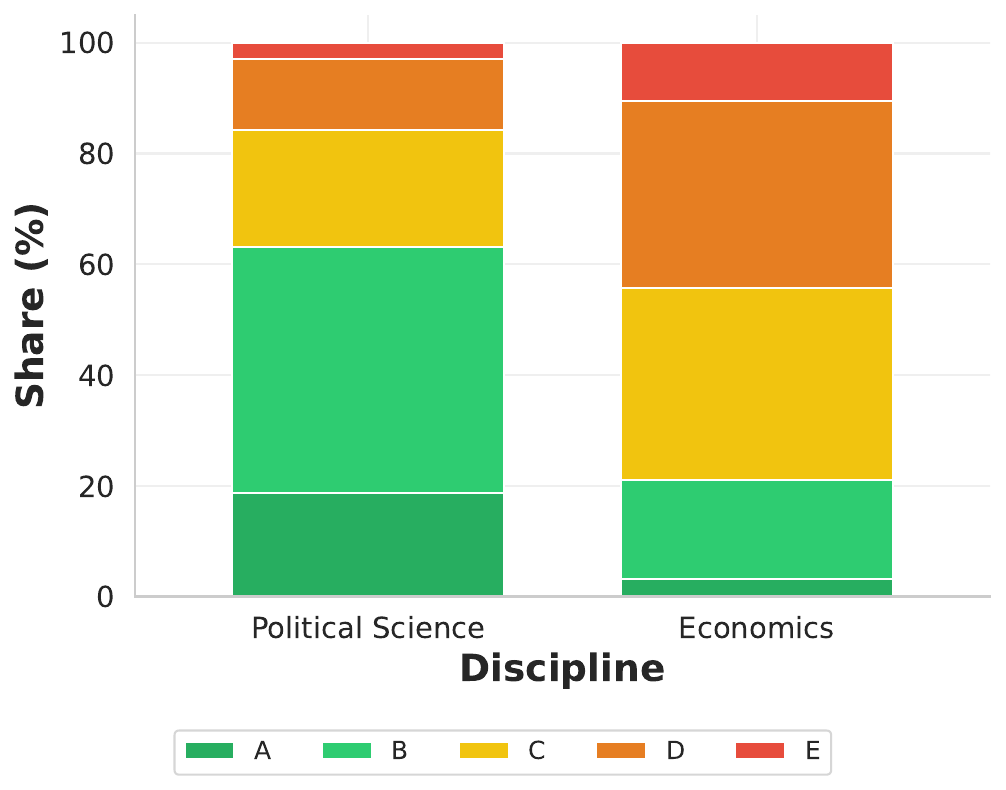}
        \caption{}
        \label{fig:app-grade_by_discipline}
        \end{subfigure}
    \hfill
    \begin{subfigure}[t]{0.48\textwidth}
        \centering
        \includegraphics[width=\textwidth]{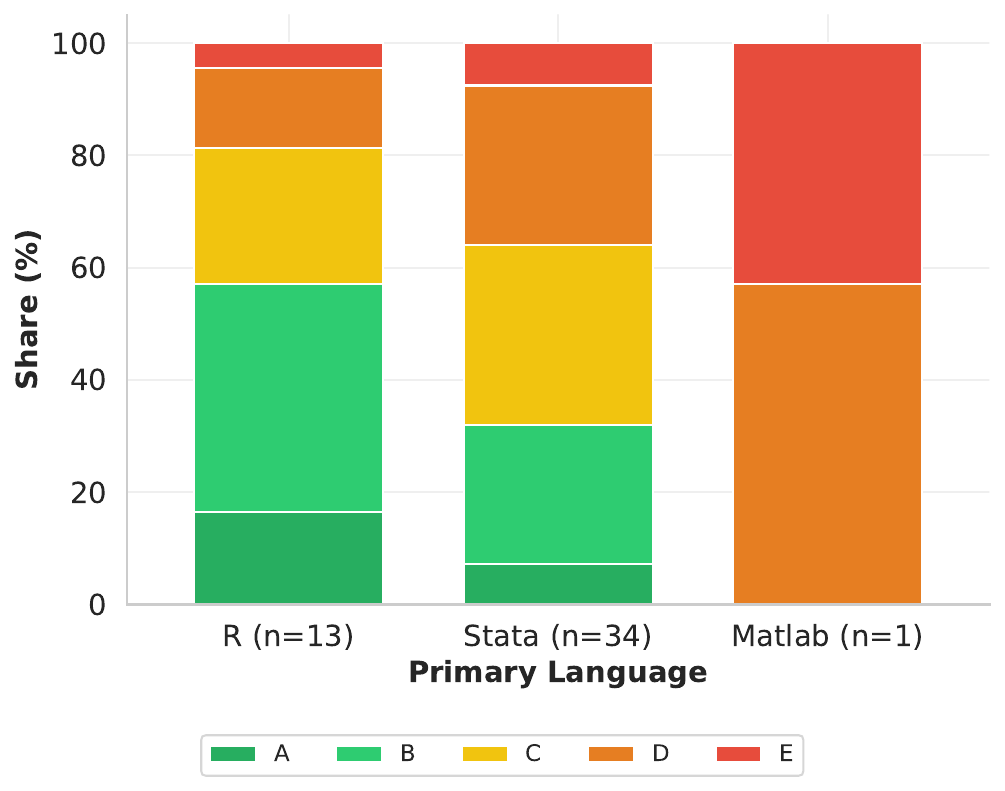}
    \caption{}
        \label{fig:app-language-grade}
    \end{subfigure}
    \caption{Paper-level grades by social science discipline and programming language used in the original reproduction package. Only papers with a single programming language considered.}
    \label{fig:app-discipline-lang}
\end{figure}

Performance also varies across disciplines (Figure~\ref{fig:app-grade_by_discipline}): economics papers are generally reproduced less accurately than political science papers. However, since our sample is not random---having been selected by I4Replication based on study reproducibility---this pattern should not be interpreted as a disciplinary characteristic. A similar selection caveat applies to the comparison between performance and the original authors' code language (Figure~\ref{fig:app-language-grade}): the finding that papers originally written in R tend to be reproduced more accurately than those in Stata or MATLAB should be interpreted with corresponding caution.

Finally, we examine whether overall dataset size or the volume of generated code are associated with reproduction performance. Both associations are modestly negative: the run-level Spearman correlations are $-0.247$ for data size and $-0.252$ for generated-code length. In approach-adjusted rank regressions on 323 non-F runs, with standard errors clustered by paper, both differ from zero ($p=0.030$ and $p=0.009$, respectively). These associations are descriptive, not causal.

\begin{figure}[ht]
    \centering
    \begin{subfigure}[t]{0.48\textwidth}
        \centering
        \includegraphics[width=\textwidth]{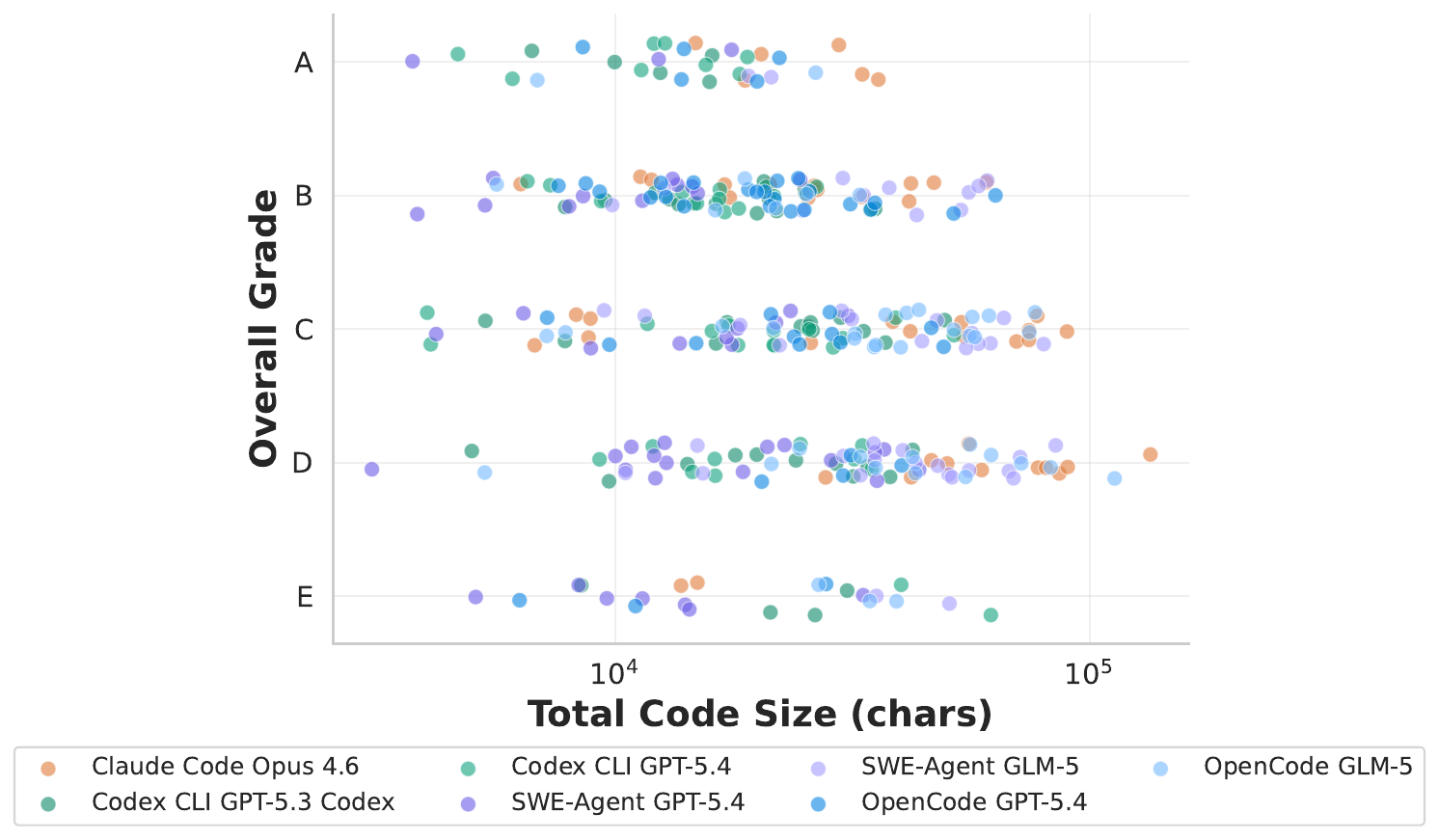}
    \end{subfigure}
    \hfill
    \begin{subfigure}[t]{0.48\textwidth}
        \centering
        \includegraphics[width=\textwidth]{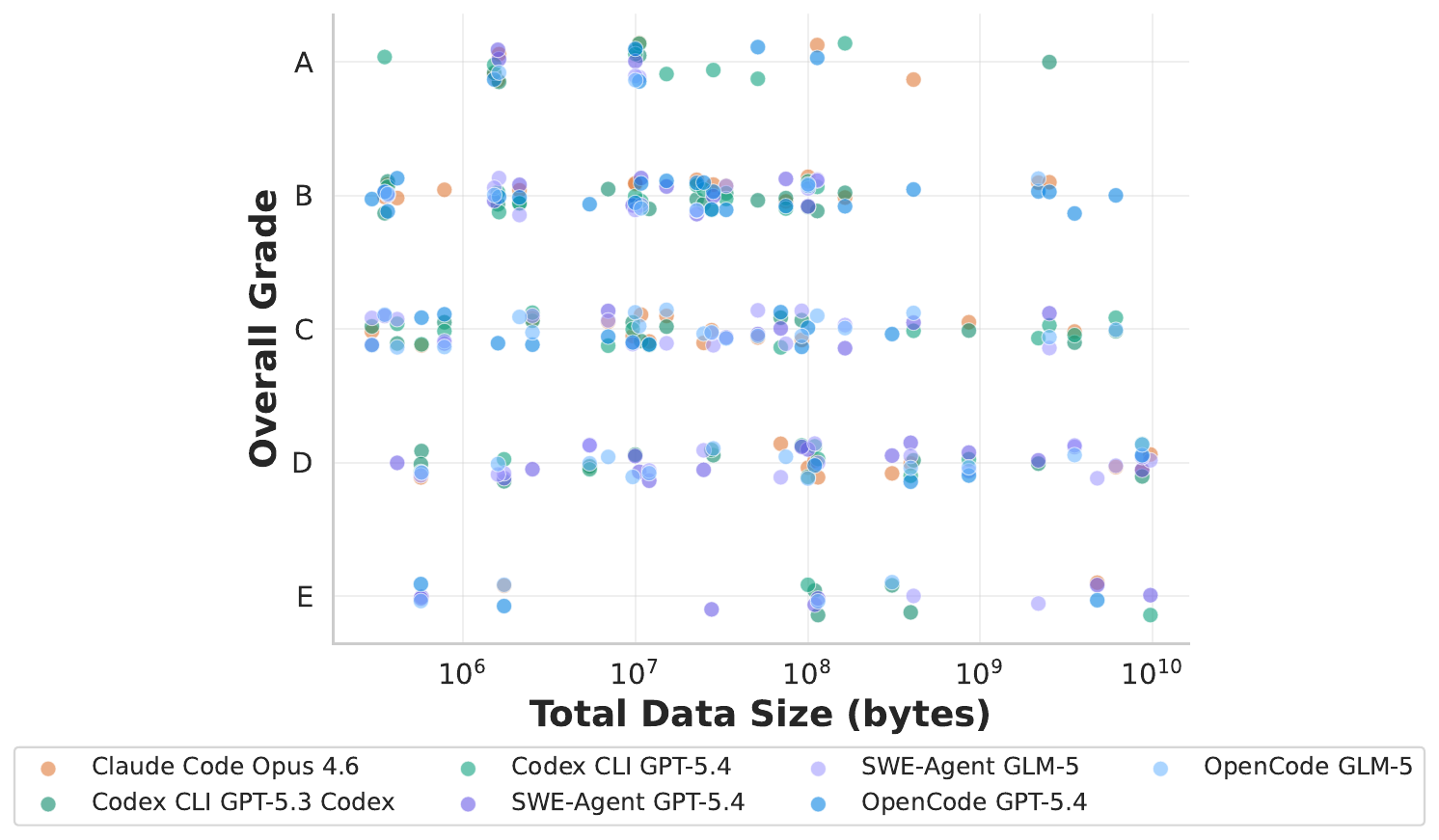}

    \end{subfigure}
    \caption{Paper-level grade by total code characters produced by the reproducing agent and the size of all datasets required for reproduction.}
    \label{fig:app-dataset-code-grades}

\end{figure}

% \begin{figure}[ht]
%     \centering
%     \includegraphics[width=0.6\textwidth]{results/pre_post_cutoff/bootstrap_grade_comparison.pdf}
%     \caption{Comparison of mean table grades for papers published before and after the model knowledge cutoff.}
%     \label{fig:pre-post-comparison}
% \end{figure}

\begin{figure}[ht]
    \centering
    \includegraphics[width=\textwidth]{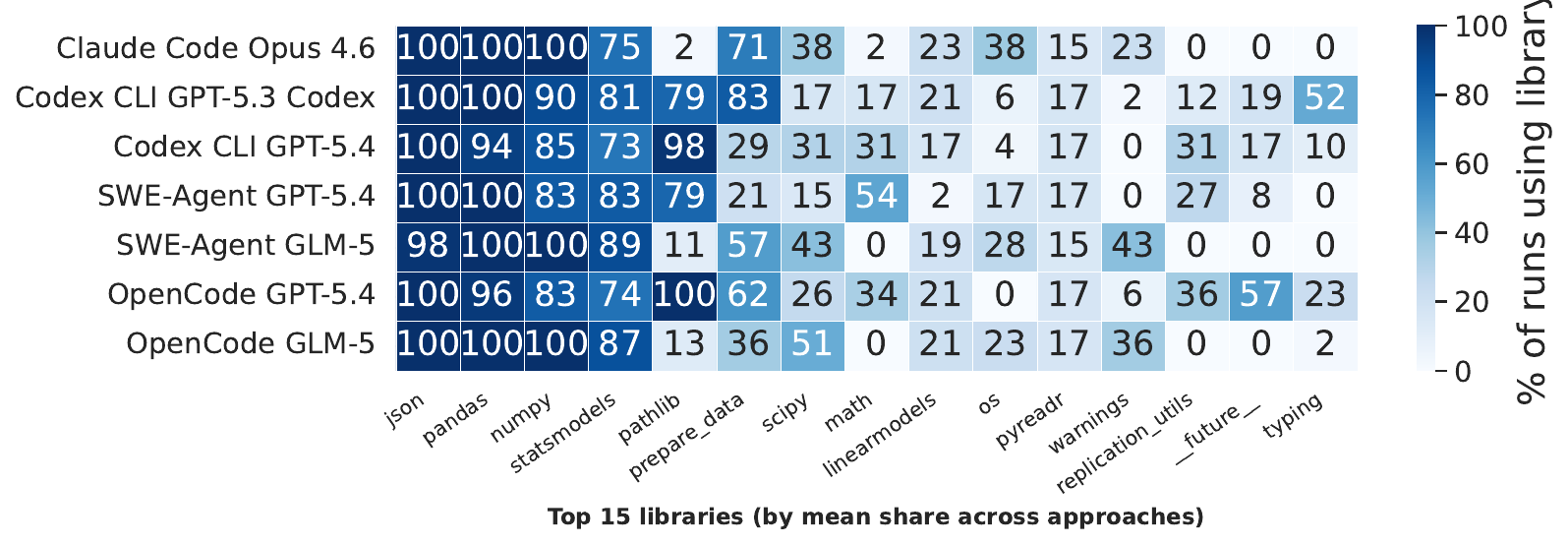}
    \caption{\textbf{Incidence of Python imports across runs by agent}. Note that \textit{prepare\_data} and \textit{replication\_utils} are python scripts that the agent created}
    \label{fig:app-languages-agents}
\end{figure}

\begin{figure}[ht]
    \centering
    \includegraphics[width=\textwidth]{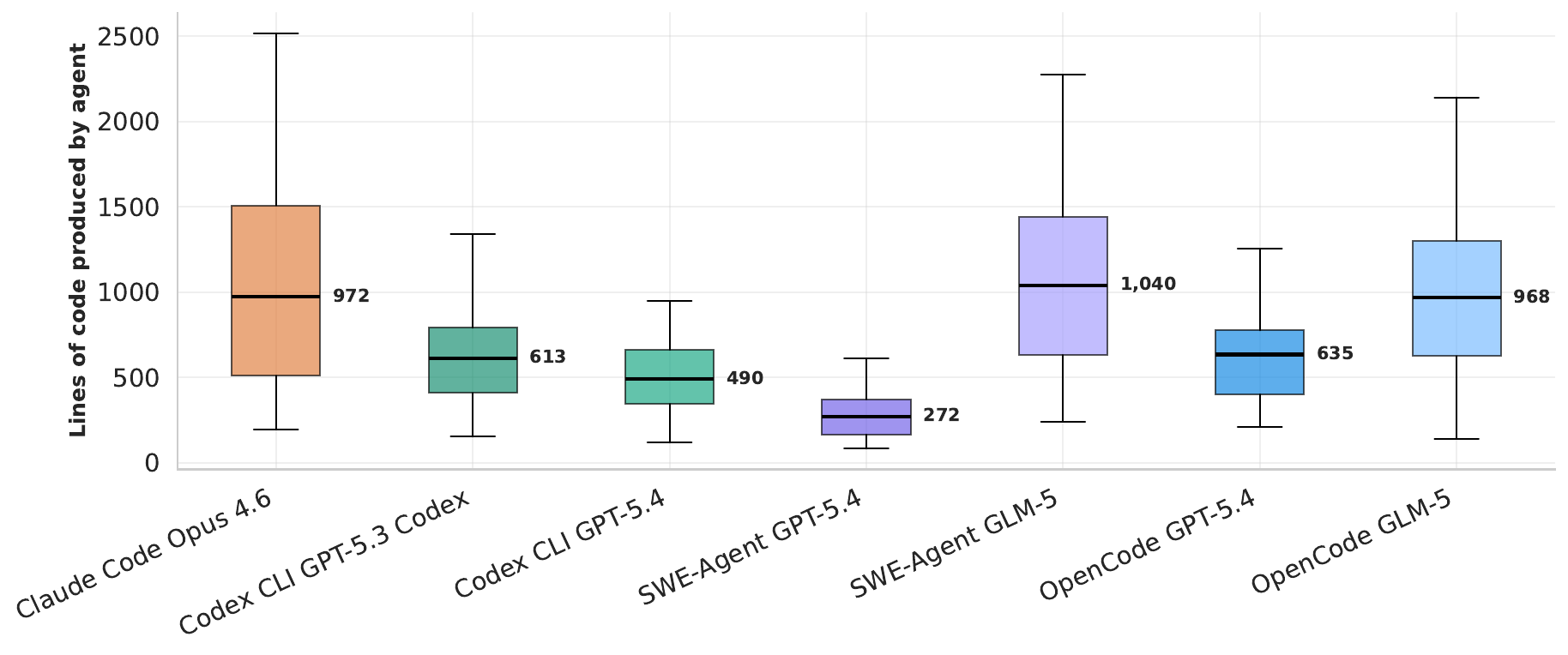}
    \caption{\textbf{Distribution of lines of codes produced by agents.} Box-and-whisker plots summarizing the number of  lines of code produced by each agent system across runs.}
    \label{fig:app-loc-agents}
\end{figure}

\clearpage 

\subsection{Error source analysis}
\label{sec:app-error-source}

% \begin{figure}[ht]
%     \centering
%     \begin{subfigure}[t]{0.48\textwidth}
%         \centering
%         \includegraphics[width=\textwidth]{results/discrepancy_analysis/divergence_types.pdf}
%         \caption{Divergence types (S-codes)}
%     \end{subfigure}
%     \hfill
%     \begin{subfigure}[t]{0.48\textwidth}
%         \centering
%         \includegraphics[width=\textwidth]{results/discrepancy_analysis/root_causes_coarse.pdf}
%         \caption{Root cause categories}
%     \end{subfigure}
%     \caption{Discrepancy analysis. Left: distribution of divergence types by approach. Right: root cause attribution from the three-way consistency check (Agent~$\leftrightarrow$~Extracted Methods, Paper~$\leftrightarrow$~Extracted Methods, Paper~$\leftrightarrow$~Code).
%      % {\color{red} I think the right one is very interesting, showing that the agent is responsible for a quarter to a third of the discrepencies, but the paper are also often unclear. I think one takeaway from this is that if we would like to introduce automated science, we can not just write text and methods like we used to, but we have to be very specific. }
%     }
%     \label{fig:discrepancy_analysis}
% \end{figure}

\begin{figure}[ht]
    \centering
    \includegraphics[width=0.8\textwidth]{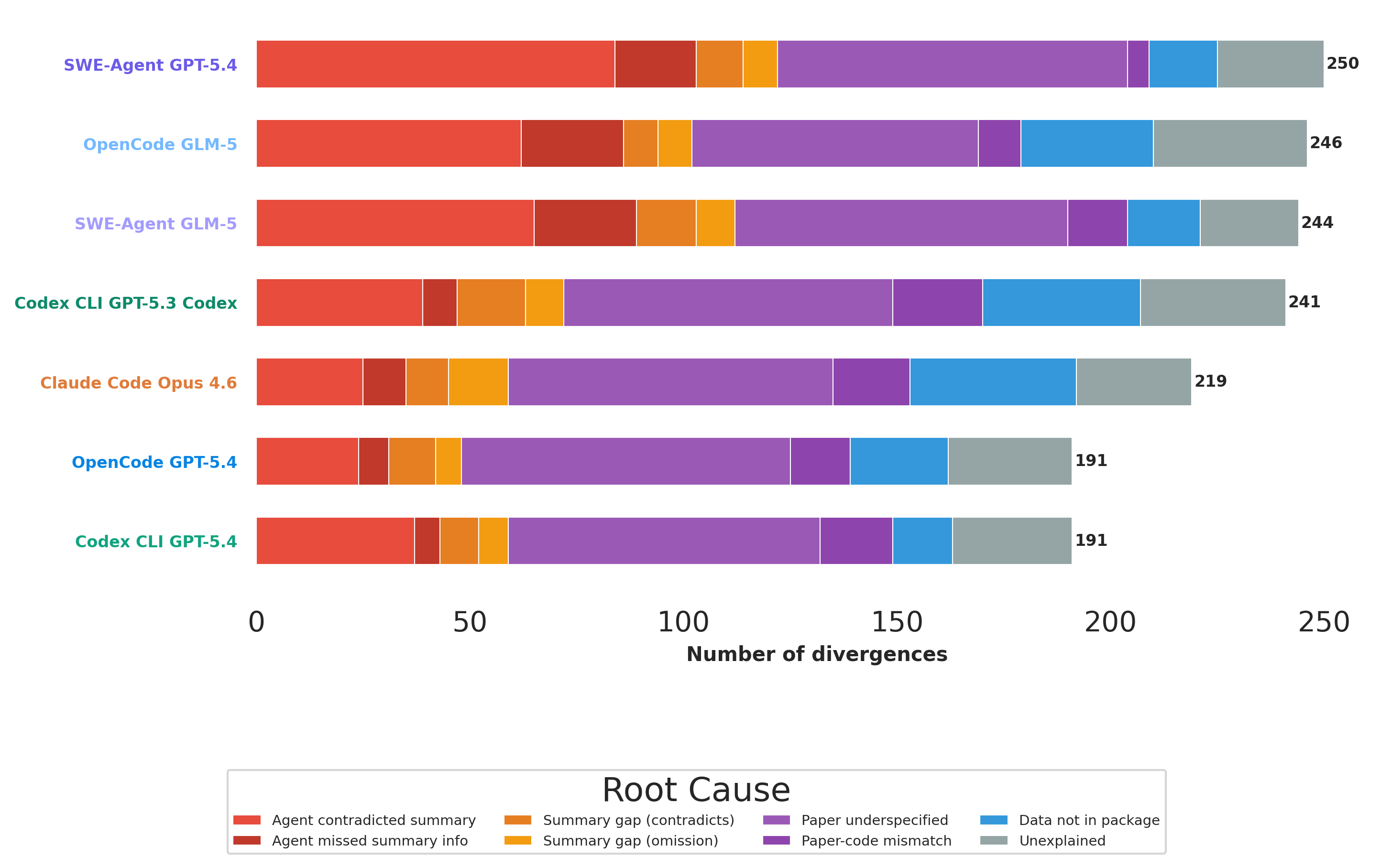}
    \caption{Comparison of error source attribution pipeline results between agents}
    \label{fig:app-divergenges-agent}
\end{figure}

\subsubsection{Human validation of error attribution}
\label{sec:app-error-source-human-audit}

We conducted a small-scale human audit of all 53 divergences associated with six tables in four purposively selected short papers. Reviewing whole papers rather than isolated records allowed annotators to reuse the relevant methodological and code context across related agent runs as reviewing these tasks requires substantial contextual knowledge. One assigned author-annotator reviewed each divergence in the primary sample. For every record, the annotator assessed whether the flagged divergence was meaningful and, if so, whether its assigned error source, divergence type, and severity were correct. A third author independently reviewed the same 27 divergences assigned to one of the original annotators in two of the four papers, providing a double-coded subset.

Table~\ref{tab:human-audit-validation} reports the validation results. Humans confirm 49 of 53 flagged divergences (92\%) as meaningful. Conditional on confirmation, 33 of 49 error-source labels (67\%) are correct, 10 (20\%) are unsure, and 6 (12\%) are labeled as incorrect. Divergence-type labels are substantially more reliable (96\% correct), as are severity labels (86\% correct). Source attribution is most reliable for agent errors and least reliable for other/unknown cases. Because the sample is purposive, clustered within four papers, and contains no data-missing cases, these rates validate the inspected records but should not be used to reweight the full-sample error-source distribution.

Table~\ref{tab:human-audit-agreement} reports agreement in the double-coded subset. Both annotators agree that every flagged case is a meaningful divergence, and they agree on the divergence type in 96\% of cases. Exact agreement is lower for severity (67\%) and especially error source (48\%), largely because one annotator uses the unsure category more frequently. We report exact agreement as the primary measure, Cohen's $\kappa$ for comparability, and Gwet's AC1 as a chance-adjusted statistic that is less sensitive to concentrated marginal distributions \citep{gwet2008agreement}. This matters here because $\kappa$ is undefined for divergence detection and equals zero for divergence type despite 96\% exact agreement; the corresponding AC1 values are 1.00 and 0.96. Together, the audit supports the pipeline's divergence detection and typing, while reinforcing that fine-grained source attribution is more uncertain and that the aggregate source shares should be read as indicative.

\begin{table*}[t]
\centering
\small
\caption{Human validation of the error-attribution pipeline.}
\label{tab:human-audit-validation}
\begin{tabular}{lrrrr}
\toprule
Audit target & Correct/yes & Unsure & Incorrect/no & $N$ \\
\midrule
\multicolumn{5}{l}{\textit{Panel A: Validation outcomes}} \\
Meaningful divergence & 49 (92\%) & 2 (4\%) & 2 (4\%) & 53 \\
Error source & 33 (67\%) & 10 (20\%) & 6 (12\%) & 49 \\
Divergence type & 47 (96\%) & 1 (2\%) & 1 (2\%) & 49 \\
Severity & 42 (86\%) & 7 (14\%) & 0 (0\%) & 49 \\
\midrule
\multicolumn{5}{l}{\textit{Panel B: Error-source validation by assigned source}} \\
Agent error & 11 (85\%) & 0 (0\%) & 2 (15\%) & 13 \\
Original error & 19 (66\%) & 8 (28\%) & 2 (7\%) & 29 \\
Extractor error & 3 (75\%) & 1 (25\%) & 0 (0\%) & 4 \\
Data missing & -- & -- & -- & 0 \\
Other / unknown & 0 (0\%) & 1 (33\%) & 2 (67\%) & 3 \\
\bottomrule
\end{tabular}
\vspace{2pt}
\begin{minipage}{0.92\textwidth}
\footnotesize \textit{Notes:} The primary sample contains 53 divergences from 27 table-runs spanning six tables in four purposively selected short papers. Panel A reports one assigned annotation per divergence. Error-source, divergence-type, and severity validation are conditional on the annotator confirming a meaningful divergence ($N=49$). Panel B maps the detailed pipeline labels to the five error-source groups reported in the main paper. The audit sample contains no data-missing cases. Percentages may not sum to 100 due to rounding.
\end{minipage}
\end{table*}

\begin{table*}[t]
\centering
\small
\caption{Inter-annotator agreement in the double-coded audit subset.}
\label{tab:human-audit-agreement}
\begin{tabular}{lrrrr}
\toprule
Audit target & Exact agreement & Cohen's $\kappa$ & Gwet's AC1 & $N$ \\
\midrule
Meaningful divergence & 27 (100\%) & -- & 1.00 & 27 \\
Error source & 13 (48\%) & -0.05 & 0.33 & 27 \\
Divergence type & 26 (96\%) & 0.00 & 0.96 & 27 \\
Severity & 18 (67\%) & 0.16 & 0.59 & 27 \\
\bottomrule
\end{tabular}
\vspace{2pt}
\begin{minipage}{0.92\textwidth}
\footnotesize \textit{Notes:} Two annotators independently rated the same 27 divergences from the runs from two papers. We treat the three responses for each target as nominal categories and report exact agreement as the primary measure. Cohen's kappa is undefined for meaningful-divergence detection because both annotators answered yes in every case; for divergence type, one annotator answered correct in all 27 cases while the other answered correct in 26 and unsure in one, yielding 96\% exact agreement but $\kappa=0$. Gwet's AC1 is less sensitive to such concentrated marginal distributions. Error-source disagreements mostly reflect different use of the ``unsure'' category.
\end{minipage}
\end{table*}

\subsubsection{Sensitivity to the attribution judge}
\label{sec:app-error-source-judge-sensitivity}

To test whether GPT-5.4 preferentially assigns fewer errors to outputs produced by GPT models, we repeated the source-attribution stage with Claude Code using Claude Opus 4.6. We held fixed the production pipeline's Step-1 divergences, final data-availability labels, source documents and code, and classification schema, and reran only the three source checks used to assign an error source. We used the ten-paper cohort selected ex ante for the extractor analysis within the stability sample and all seven model--scaffold arms in the release manifest, yielding 70 paper--agent runs and 334 matched divergences; nine papers contributed at least one divergence. This is therefore a matched sensitivity analysis of the attribution judge rather than a rerun of divergence discovery.

Table~\ref{tab:judge-sensitivity} reports the results. Fine-grained source labels agree in 59.3\% of cases ($\kappa=0.474$), the five reported source groups agree in 63.5\% ($\kappa=0.482$), and the binary distinction between agent and non-agent errors agrees in 77.8\% ($\kappa=0.540$). Claude assigns 44.6\% of divergences to agent error, compared with 32.6\% under GPT-5.4, a 12.0 percentage-point increase (paper-cluster bootstrap 95\% CI: 5.6 to 19.7; exact McNemar $p<0.001$). Absolute attribution shares are thus meaningfully judge-sensitive. The broad arm-level pattern nevertheless remains: OpenCode GPT-5.4 and Claude Code, two relatively strong arms, have the lowest agent-error shares under the alternate judge (20.0\% and 20.5\%), whereas the two lower-performing GLM-5 arms have substantially higher shares (56.2\% and 53.8\%). The Claude Code arm is also the only arm whose agent-error share falls under the Claude judge, so the arm-specific estimates should not be read as evidence that Claude is itself free of self-preference.

The shift does not, however, appear specific to GPT-produced outputs. Claude increases the agent-error share by 12.8 percentage points for GPT outputs and by 10.7 points for non-GPT outputs. The difference between these increases---our directional test of GPT-judge self-preference---is 2.1 percentage points (paper-cluster bootstrap 95\% CI: $-8.8$ to 14.1). Although the estimate is positive in the hypothesized direction, its interval is wide and spans zero. We therefore find no clear evidence that GPT-5.4 differentially favors GPT outputs in this subset, while recognizing that the comparison is based on only nine divergence-contributing papers and does not establish equivalence across judges.

\begin{table}[t]
\centering
\small
\caption{Matched sensitivity of error-source attribution to the judge model.}
\label{tab:judge-sensitivity}
\resizebox{\textwidth}{!}{%
\begin{tabular}{lrrrrr}
\toprule
Output-producing model and scaffold & $N$ & GPT-5.4 judge & Claude judge & $\Delta$ (pp) & Binary agreement \\
\midrule
\multicolumn{6}{l}{\textit{Panel A: Aggregate comparison}} \\
All outputs & 334 & 32.6\% & 44.6\% & 12.0 & 77.8\% \\
GPT-model outputs & 203 & 32.5\% & 45.3\% & 12.8 & 76.4\% \\
Non-GPT-model outputs & 131 & 32.8\% & 43.5\% & 10.7 & 80.2\% \\
\midrule
\multicolumn{6}{l}{\textit{Panel B: Model--scaffold arms}} \\
Claude Opus 4.6 / Claude Code & 44 & 22.7\% & 20.5\% & $-2.3$ & 88.6\% \\
GPT-5.3 / Codex & 56 & 23.2\% & 41.1\% & 17.9 & 67.9\% \\
GPT-5.4 / Codex & 31 & 38.7\% & 45.2\% & 6.5 & 74.2\% \\
GPT-5.4 / OpenCode & 50 & 16.0\% & 20.0\% & 4.0 & 88.0\% \\
GPT-5.4 / mini-SWE-agent & 66 & 50.0\% & 68.2\% & 18.2 & 75.8\% \\
GLM-5 / OpenCode & 48 & 39.6\% & 56.2\% & 16.7 & 70.8\% \\
GLM-5 / mini-SWE-agent & 39 & 35.9\% & 53.8\% & 17.9 & 82.1\% \\
\bottomrule
\end{tabular}}
\vspace{2pt}
\begin{minipage}{0.94\textwidth}
\footnotesize \textit{Notes:} Entries are shares of matched divergences classified as agent error by each judge. $\Delta$ is Claude minus GPT-5.4 in percentage points. Binary agreement compares agent-error versus all other source classifications. The GPT group contains the four GPT-5.3/GPT-5.4 arms; the non-GPT group contains the Claude Opus 4.6 and two GLM-5 arms. The GPT-versus-non-GPT contrast in $\Delta$ is 2.1 percentage points (paper-cluster bootstrap 95\% CI: $-8.8$ to 14.1). Three records with fixed missing-data classifications are included in the denominators but were not eligible for model judgment. Percentages and differences may differ slightly because of rounding.
\end{minipage}
\end{table}

% Add to your LaTeX preamble:
%   \usepackage{booktabs, longtable, colortbl, xcolor, pifont, textcomp, tikz}
%   \newcommand{\cmark}{\textcolor{green!60!black}{\ding{51}}}
%   \newcommand{\xmark}{\textcolor{red}{\ding{55}}}
%   \newcommand{\omark}{\textcolor{orange!80!black}{$\circ$}}
%   \setlength{\tabcolsep}{4pt}
%   % Pipeline stage badges (fail stage column)
%   \newcommand{\stageData}{%
%     \tikz[baseline=-0.6ex]\node[fill=gray!20,rounded corners=2pt,
%       inner xsep=3pt,inner ysep=1.5pt,font=\scriptsize\sffamily]{%
%       Data avail.};}
%   \newcommand{\stagePC}{%
%     \tikz[baseline=-0.6ex]\node[fill=blue!12,rounded corners=2pt,
%       inner xsep=3pt,inner ysep=1.5pt,font=\scriptsize\sffamily]{%
%       Code\,$\neq$\,Paper};}
%   \newcommand{\stagePS}{%
%     \tikz[baseline=-0.6ex]\node[fill=orange!18,rounded corners=2pt,
%       inner xsep=3pt,inner ysep=1.5pt,font=\scriptsize\sffamily]{%
%       Paper\,$\neq$\,Sum.};}
%   \newcommand{\stageSA}{%
%     \tikz[baseline=-0.6ex]\node[fill=green!15,rounded corners=2pt,
%       inner xsep=3pt,inner ysep=1.5pt,font=\scriptsize\sffamily]{%
%       Sum.\,$\neq$\,Agent};}

\small
\setlength{\tabcolsep}{3pt}
\begin{longtable}{p{1.4cm} c p{2.2cm} c p{0.9cm} p{5.5cm}}
\caption{Example for output of error source pipeline identifying divergences, assigning severity, a stage of failure and a type of error for paper 10.1086\_715252}\label{tab:errors_10_1086_715252}\\
\toprule
\textbf{Output} & \textbf{Sev.} & \textbf{Fail stage} & \textbf{Fail type} & \textit{Agent} & \textbf{Description} \\
\midrule
\endfirsthead
\multicolumn{6}{l}{\small\textit{(continued)}} \\
\toprule
\textbf{Output} & \textbf{Sev.} & \textbf{Fail stage} & \textbf{Fail type} & \textit{Agent} & \textbf{Description} \\
\midrule
\endhead
\midrule \multicolumn{6}{r}{\small\textit{continued on next page}} \\
\endfoot
\bottomrule
\multicolumn{6}{p{\linewidth}}{\smallskip\noindent{\footnotesize\textit{Notes:} Fail type: \xmark~contradicts (direct conflict); \omark~omission (upstream specifies, downstream silent); ?~unclear. Agents: Claude = Claude Code Opus~4.6; Codex~5.3/5.4 = Codex CLI; OC = OpenCode; SWE = SWE-Agent; GLM = GLM-5.}} \\
\endlastfoot
  Table 1 & \cellcolor{red!15}critical & \stageSA & \xmark & \textit{Claude} & Agent uses simple AMCE-difference approach instead of the parallel design causal mediation estimator (med.para) with covariates, and reports two-sided instead of one-sided p-values. \\
  Table 1 & \cellcolor{red!15}critical & \stagePC & \omark & \textit{Codex 5.3} & The Python replicator substitutes the original parallel-design mediation/randomization-inference procedure with a pooled round-level product-of-coefficients OLS routine, so the indirect-effect estimates and their reported p-values are not computed the way the R code does. \\
  Table 1 & \cellcolor{red!15}critical & \stageSA & \xmark & \textit{Codex 5.4} & The agent maps the mediator-specific natural and controlled columns to the wrong experiment indicators, so the natural columns use `exp\_id\_*\_con` and the controlled columns use `exp\_id\_*\_nat`. \\
  Table 1 & \cellcolor{red!15}critical & \stagePC & \omark & \textit{OC 5.4} & The Python replicator replaces the original covariate-adjusted `med.para.ri()` mediation analysis on `exp\_id\_* == 1` versus `== 2` experiment arms with round-based treatment-control mean differences and separate within-round permutation tests. \\
  Table 1 & \cellcolor{red!15}critical & \stagePC & \omark & \textit{SWE 5.4} & The Python replicator does not implement the original mediation routine for Table 1 and instead fills every indirect-effect, parenthesized uncertainty, and total-effect cell from a simple punishment mean-difference plus bootstrap computed on hand-picked round subsets. \\
  Table 1 & \cellcolor{orange!20}medium & \stagePC & \omark & \textit{OC GLM} & The Python code reports HC1 standard errors in the inference rows, while the original R code reports randomization-inference tail probabilities (`indirect.greater` and `total.greater`). \\
  Table 1 & \cellcolor{red!15}critical & \stagePC & \omark & \textit{SWE GLM} & The Python replicator does not implement the original `med.para.ri` parallel-design mediation procedure for Table 1; instead it builds columns from ad hoc round filters and estimates indirect effects with two OLS regressions on `rate\_moral`/`rate\_competence`, omitting the original experiment-ID split, covariate set, and permutation-based inference. \\
  \midrule[0.3pt]
  Table 1 & \cellcolor{orange!20}medium & \stagePC & \omark & \textit{Codex 5.3} & For total effects, the Python code estimates a simple treatment coefficient on pooled round subsets, while the R code estimates the total effect only on the treatment-only experiment arm selected by each `exp\_id\_*` variable and includes the specified conjoint covariates. \\
  Table 1 & \cellcolor{red!15}critical & \stagePC & \omark & \textit{Codex 5.4} & The agent does not implement the paper's mediation estimator with covariate adjustment and randomization-inference p-values; it instead fits a pooled `vote \textasciitilde{} punishment * group` regression and reports cluster-robust normal p-values. \\
  Table 1 & \cellcolor{orange!20}medium & \stagePC & \omark & \textit{OC GLM} & The Python replicator replaces the mediation estimator with a simple difference between treatment coefficients from experiment arms, omitting the covariate-adjusted total-effect regression and the mediator-level weighted direct-effect calculation used in the R code. \\
\end{longtable}

% Note: the longtable above already has its own \caption and \label
%       (tab:errors_10_1086_715252); the orphan \caption/\label that were
%       here caused "\caption outside float" errors and have been removed.

\clearpage 

\subsection{Code snippets from reproduction agents}

\begin{figure}[ht]
    \centering
\begin{subfigure}[t]{0.49\textwidth}
    \includegraphics[width=\textwidth]{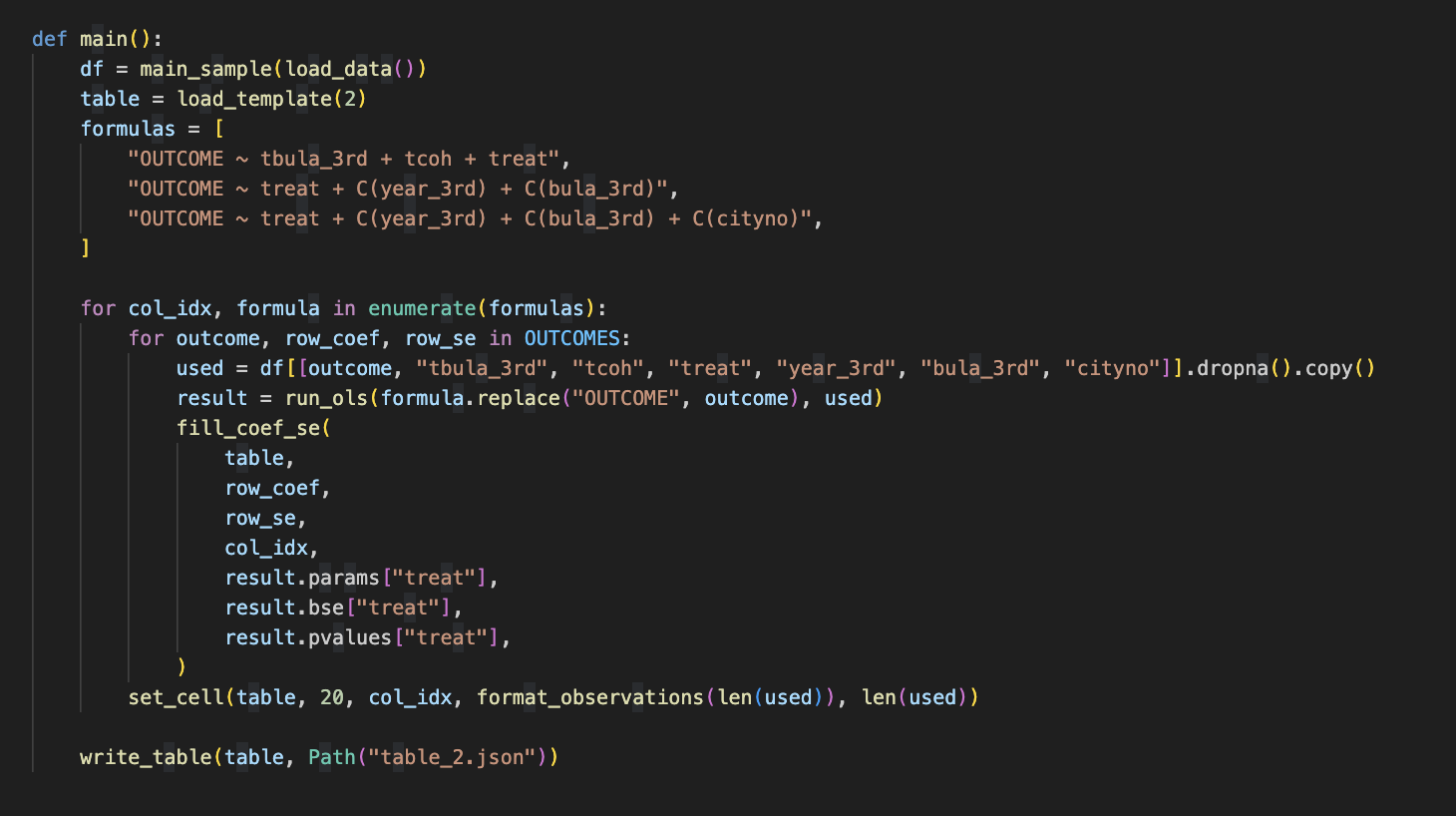}
    \caption{Codex 5.4 --- \textit{table\_2.py}}
\end{subfigure}
\begin{subfigure}[t]{0.49\textwidth}
    \includegraphics[width=\textwidth]{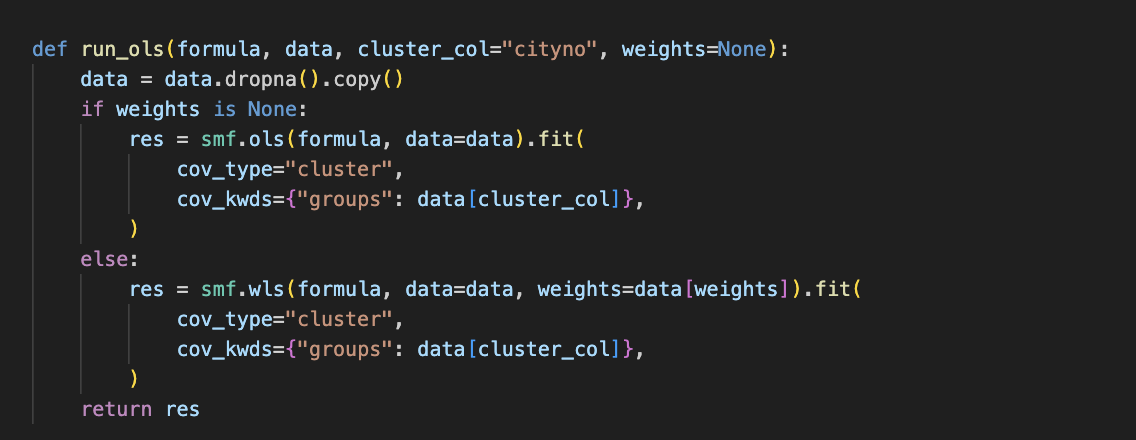}
    \caption{Codex 5.4 --- \textit{replication\_utils.py}}
\end{subfigure}
\begin{subfigure}[t]{0.48\textwidth}
    \includegraphics[width=\textwidth]{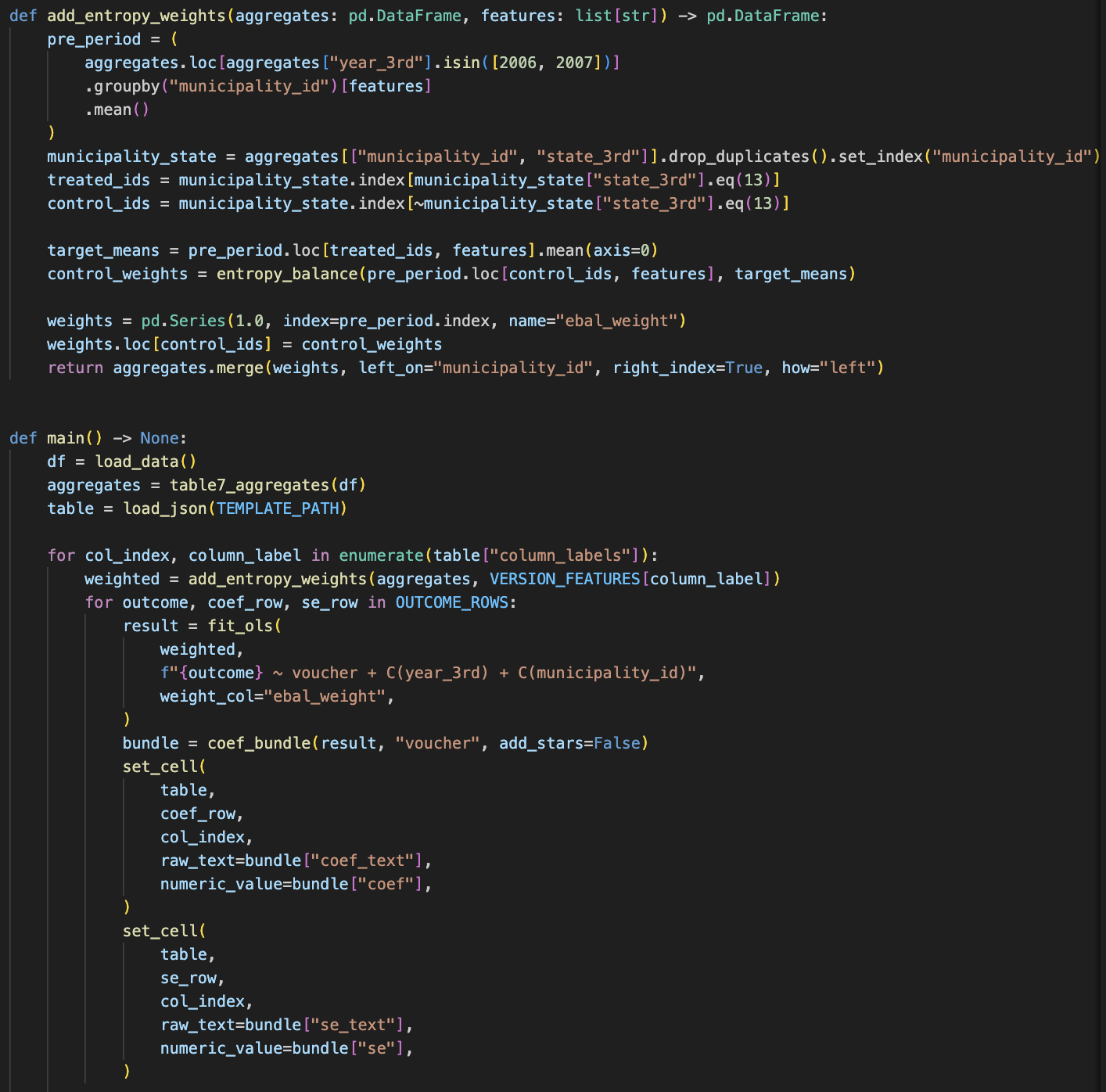}
    \caption{OpenCode GPT 5.4 --- \textit{table\_7.py}}
\end{subfigure}
\begin{subfigure}[t]{0.48\textwidth}
    \includegraphics[width=\textwidth]{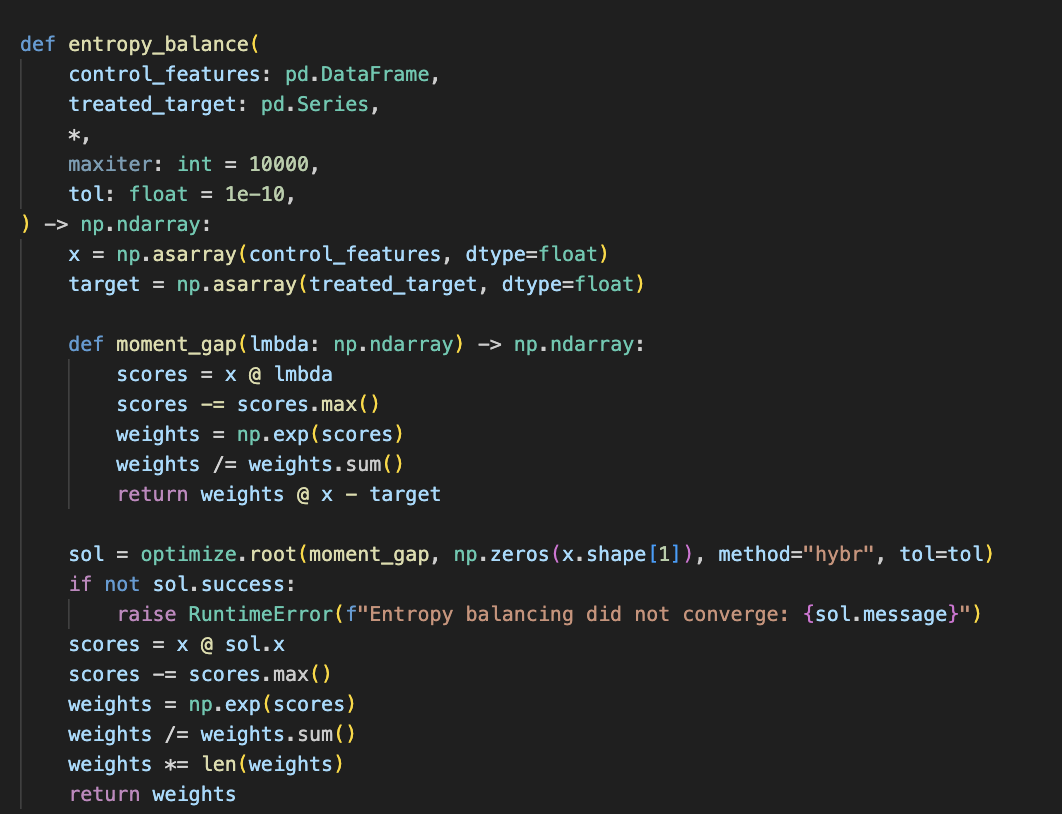}
    \caption{OpenCode GPT 5.4 --- \textit{replication\_utils.py}}
\end{subfigure}

\caption{\textbf{Example screenshots from agents reproducing standard econometric methods}. All screenshots are from reproduction attempts of the paper "The Long-Run Effects of Sports Club Vouchers for Primary School Children" (Marcus, Siedler \& Ziebarth - AEJ Economic Policy (2022)). The screenshots show segments of the reproduced code for this paper. Panel (a) shows the main regression using a TWFE model for DiD analysis. Functions like \textit{run\_ols}, \textit{format\_observation}, \textit{write\_table} are created as helper functions in a separate file. Panel (b) shows such a function to be used in multiple table creation files. Panel (c) shows the method of an alternative empirical approach via an entropy-balanced synthetic control, applying a shared function shown in Panel (d).}
\end{figure}
\label{fig:app-screenshots}

\clearpage

\section{Trace analysis and guardrails}

\label{app:guardrails_and_model_trace_audits}

\subsection{Setup}
 
Each agent operates in a workspace containing: \texttt{TASK.md} (replication instructions extracted from the paper), \texttt{methodology\_summary.json} (structured methodology context), \texttt{table\_templates/*.json} (expected output structure), a symlink to the original dataset, and agent-specific instruction files. These files contain paper-derived content by design. The original replication package and paper PDF reside on the same machine but outside the workspace boundary. Web access was restricted using the agent scaffold's configuration to the extent possible.

\subsection{Trace analysis}

\label{app:trace-analysis}

Runs are analyzed by parsing execution logs and extracting tool calls. The agents differ in how they expose file operations: Claude Code and OpenCode provide structured \texttt{Read}/\texttt{Write}/\texttt{Edit} tools alongside bash, Codex exposes bash together with a programmatic file editor, and SWE-Agent operates primarily through bash.

To normalize across these interfaces, each bash command is split into sub-commands at common boundaries (e.g., \texttt{\&\&} chaining), while preserving heredoc bodies as atomic units. Each sub-command is then classified by its first token into one of:
\begin{itemize}
    \item \texttt{exec}: executing code (e.g., \texttt{python}, \texttt{Rscript}, \texttt{make})
    \item \texttt{read}: reading files (e.g., \texttt{cat}, \texttt{head}, \texttt{sed})
    \item \texttt{navigation}: filesystem navigation (e.g., \texttt{ls}, \texttt{cd}, \texttt{pwd})
    \item \texttt{search}: searching (e.g., \texttt{grep}, \texttt{find}, \texttt{rg})
    \item \texttt{write}: writing files (e.g., \texttt{echo}, \texttt{cp}, \texttt{tee}, \texttt{cat} with \texttt{>})
    \item \texttt{other}: everything else
\end{itemize}

For each run we measure two quantities: the \emph{action count}, i.e.\ the number of tool calls (or sub-commands in the case of bash), and the \emph{tool-call character count}, i.e.\ the number of characters the LLM generated per tool call as illustrated in Figure~\ref{fig:trace-analysis} in the main body.

\subsection{Guardrail audit}

For the replication runs we extract the event trace (all tool calls with inputs and truncated outputs) and the workspace file tree and contents. Guardrail adherence is assessed via two
independent signals:
 
\paragraph{Regex scan.} We extract all absolute file paths and URLs from the raw event trace using regular expressions. Each path is classified as: \emph{allowed\_workspace} (inside the workspace), \emph{allowed\_data} (the mounted dataset symlink), \emph{forbidden\_replication\_package} (inside \texttt{papers/<id>/replication\_package/}),
\emph{forbidden\_paper\_pdf} (the paper file), or \emph{forbidden\_external} (any other path outside the workspace). URLs and web-access keywords (\texttt{curl}, \texttt{wget}, \texttt{requests}) are flagged separately. This provides a deterministic baseline independent of
LLM judgment.
 
\paragraph{LLM review.} In an effort to flag additional runs for potential violations we prompt GPT-5.4-mini with the normalized event trace and workspace artifacts to identify access breaches. Each breach is classified as:
\begin{itemize}
    \item \emph{external\_result\_lookup}: fetching results from the web, APIs, or forbidden files outside the workspace;
    \item \emph{forbidden\_paper\_access}: reading the published paper or prior replication results;
    \item \emph{forbidden\_code\_access}: reading original replication scripts or code from outside the workspace.
\end{itemize}
Each breach includes the artifact and line range, severity (low/medium/high), confidence (low/medium/high), and an evidence chain. The overall run assessment ranges from \emph{clean} to \emph{severe\_violation}, with \emph{insufficient\_evidence} for runs where logs are missing or truncated. The model is explicitly instructed that all workspace files are allowed inputs and that incorrect replications are not breaches. In practice we find the LLM based review to be over-inclusive; severe suspected cases are manually inspected and rerun if breaches were found. 
 
\subsection{Hardcoding audit}
 
\begin{figure}[ht]
    \centering
    \includegraphics[width=0.7\textwidth, trim=0 0 0 30, clip]{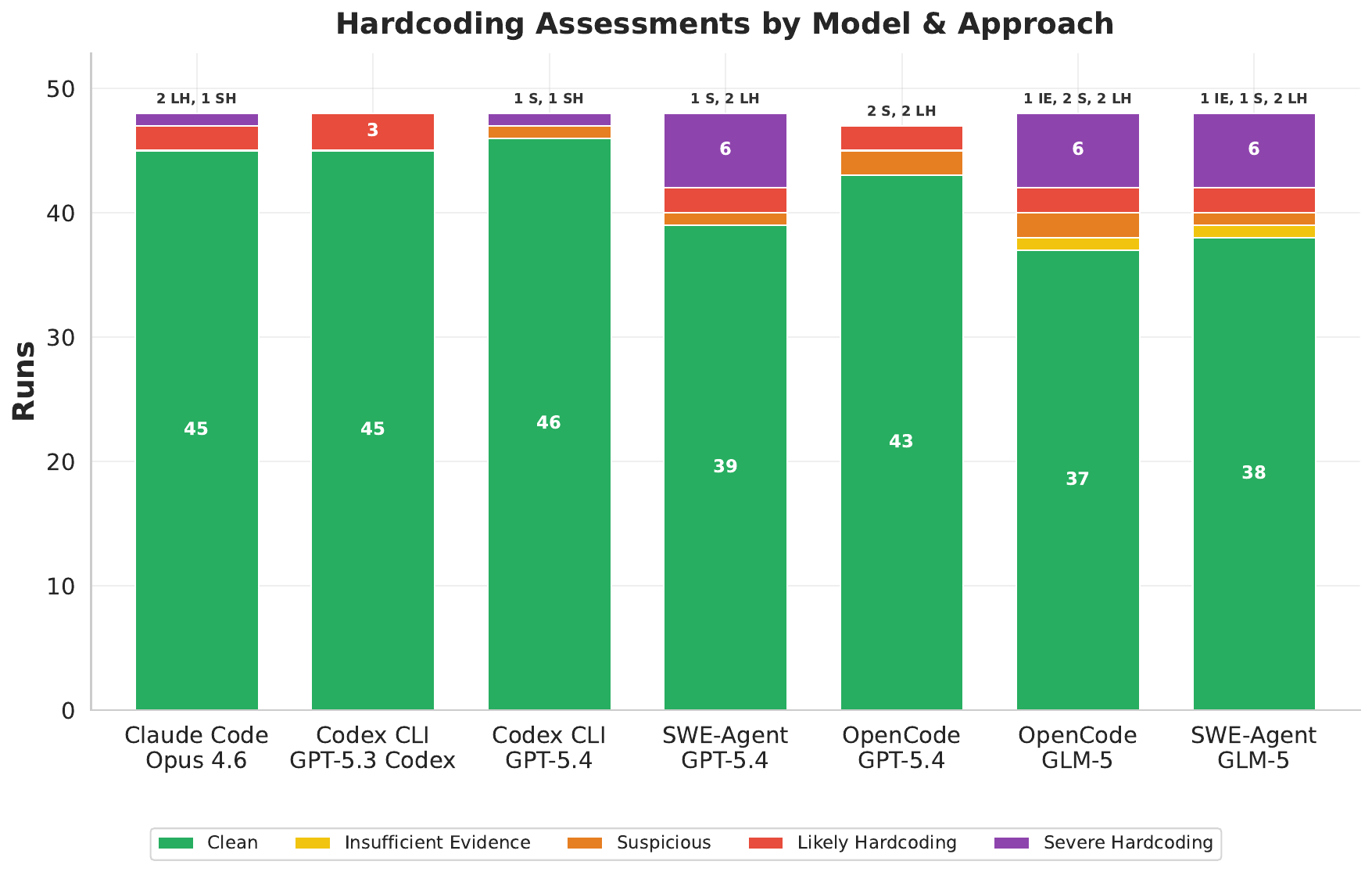}
    \caption{Hardcoding assessment outcomes by model and agent scaffold. Each stacked bar shows the number of runs classified from \emph{clean} to \emph{severe hardcoding}.}
    \label{fig:app-hardcoding_assessments_by_model_approach}
\end{figure}

For each run, we provide GPT-5.4-mini with all generated Python scripts from the workspace alongside the methodology summary and table templates for context. The model identifies numeric literals that appear in the script's output as statistical results but have no computation path from the dataset. Each run is classified from \emph{clean} to \emph{severe\_hardcoding}, with \emph{insufficient\_evidence} when Python files are missing or incomplete. In Figure~\ref{fig:app-hardcoding_assessments_by_model_approach} we classify the runs per agent scaffolding and model into their respective hardcoding categories. To assess whether hardcoding inflates replication grades, we compare runs classified as clean ($n = 293$) against all non-clean runs ($n = 42$). Clean runs achieve a mean grade of 3.03 ($\approx$ C) compared to 2.55 (between C and D) for non-clean runs. Non-clean runs perform no better—and if anything worse—than clean runs in this descriptive comparison; this does not establish whether hardcoding affects individual reproduction outcomes.

\clearpage

\section{Prompts}\label{app:prompts}
\begin{prompt}[title={C1. Extraction Prompt}, label=prompt:extraction]
\begin{alltt}\ttfamily\scriptsize
You are a methodological extraction specialist for social science research papers.

Your task: extract the methodology, structure, and specifications from academic papers so that
a replicator who cannot see the paper can reproduce every table and figure. Extract NO results.

## 1. What to extract

- Research questions
- Data description, source, sample size, time period
- Data processing steps: BOTH general steps (applied to all analyses) AND per-table/per-figure
  steps (sample restrictions, variable construction, or filtering specific to one table or figure)
- Per-table: exact column headers, exact row labels, panel structure, caption, notes,
  regression specifications, data source (if different tables use different datasets)
- Per-figure: full caption, plot type, axis labels, legend/series entries, subplot structure,
  visual details (approximate axis ranges, reference lines, line styles, color conventions)
- Per-regression-spec: model type, dependent variable, independent variables, controls,
  fixed effects, clustering, sample restrictions, equation, variable definitions

Extract ALL tables and ALL data-based figures. Skip only purely conceptual visualizations
(flow diagrams, frameworks, screenshots, photos, maps). Do not skip tables because they seem
simple - summary statistics, balance tables, and cross-tabulations must all be extracted.

## 2. What NOT to extract (results)

Never include: regression coefficients, standard errors, t-statistics, p-values, significance
stars, point estimates, confidence intervals, effect sizes, or descriptions of direction/magnitude.

DO include: table/figure structure (headers, labels, panels), design parameters (sample sizes,
thresholds, variable names), and descriptive labels ("N", "R-squared", "Controls", "Yes"/"No").

## 3. Data processing and cleaning steps

This is critical for replication. Extract every step needed to go from the raw data files to
each analysis-ready sample. Separate these into:

**General steps** (applied before any specific analysis):
- File loading, merging, and reshaping (which files, which keys)
- Variable construction (new columns derived from raw data)
- Data cleaning: missing value treatment, outlier removal, winsorization
- Sample filtering: time period restrictions, geographic restrictions, demographic restrictions
- Transformations: log transforms, standardization, encoding

**Per-table / per-figure steps** (use `sample_restrictions` on each table/figure spec):
- Additional filtering specific to one table (e.g., "Table 3 restricts to males aged 25-64")
- Subsample definitions for different columns (e.g., "Columns 1-3 use the full sample,
  columns 4-6 restrict to the treatment group")
- Variable construction specific to one analysis (e.g., "For Figure 2, compute rolling 30-day
  averages of daily returns")

Use the actual variable names from the dataset where possible.

## 4. Regression specifications

Attach each regression spec to the table or figure that displays its results. Include one spec
per distinct model variant (e.g., different controls or subsamples across columns).

For each spec:
- Use actual variable names from the paper, not generic descriptions.
  Write `dependent_var: "log(wage)"`, not `dependent_var: "outcome variable"`.
- `controls`: enumerate every control variable individually. Never write aggregate descriptions
  like "socio-demographic controls"--- list each variable: ["age", "female", "education_level", ...].
- `equation_latex`: copy the exact equation from the paper in LaTeX notation, preserving all
  subscripts, Greek letters, and notation. If no explicit equation is given, write one from the
  prose description. Example:
  `Y_i = \textbackslash{}\textbackslash{}beta_0 + \textbackslash{}\textbackslash{}beta_1 X_i + \textbackslash{}\textbackslash{}gamma Z_i + \textbackslash{}\textbackslash{}delta_j + \textbackslash{}\textbackslash{}epsilon_i`
- `variable_definitions`: define every symbol verbally, separated by semicolons. Include all
  information the paper provides: definition, construction, coding, unit, sign convention,
  omitted categories, data source. Only include what is explicitly stated--- do not infer.
  Example: `A_i^T: acceptance of targeted tax (1 if not "No", as defined in Section 2.3);
  c_i: income-threshold fixed effects (4 levels: bottom 20/30/40/50 percentile)`
- `sample_restrictions`: extract the exact condition including any mathematical formulation
  and expected sample size.
  BAD: "Among invalidated respondents"
  GOOD: "Among invalidated respondents, defined as sgn(g_i) != sgn(gamma_hat_i) (Section 4.1). N = 1,365."

## 5. Cross-reference resolution

When anything references content elsewhere ("see Appendix H", "controls listed in Table A3",
"as defined in Section 2"), go to that location and extract the actual content. Never leave
unresolved references--- the replicator cannot see the paper. This is especially important for
control variable lists: find the appendix and enumerate every variable individually.

## 6. Variable completeness

Extract ALL variables: dependent, independent, controls, instruments, weights, and constructed
variables. For each, combine information from all locations in the paper (methodology section,
table notes, appendices). The goal: a replicator should be able to construct every variable
exactly as the authors did based solely on your extraction.

## 7. Table and figure precision

- Copy EXACT column headers and row labels as printed. Count columns and rows precisely.
- Include panel structure (Panel A / Panel B) when present.
- Copy full captions and table notes (excluding notes about specific coefficient values).
- For figures: note approximate axis ranges, reference lines, and line style conventions.
- If different tables use different datasets, specify the data source per table."""


EXTRACTION_USER_PROMPT = """Extract the methodology from this paper. Follow the system instructions.

## Detected Table Captions:
\{table_captions\}

## Detected Figure Captions:
\{figure_captions\}

## Paper Text:
\{paper_text\}"""


TEMPLATE_GENERATION_SYSTEM_PROMPT = """You are a structural template specialist for academic paper tables and figures.

Generate templates that faithfully reproduce the exact layout of tables and figures from a paper.

## Table templates

- EXACTLY the same number of columns and rows as the original table.
- Use the EXACT column headers and row labels from the paper.
- Cell content rules (check the ORIGINAL TABLE in the paper for each cell):
  - **XXX**: computed result (coefficient, statistic, count, mean, etc.)
  - **(XXX)**: standard error in parentheses
  - **Empty**: leave blank if blank in the original (do NOT use --- or placeholders)
  - **Literal text**: copy "Yes", "No", checkmarks, or labels as-is
- Include panel headers (Panel A, Panel B) as spanning rows when present.
- Include the full caption above the table.

## Figure templates

- Matplotlib code skeleton that makes the intended visual style unambiguous.
- Use the correct plot function: `ax.plot()`, `ax.bar()`, `ax.hist()`, `ax.scatter()`, etc.
- Set line styles, markers, colors for each series (use `tab:blue`, `tab:orange`, etc.).
- Set approximate axis ranges from the paper (`ax.set_xlim()`, `ax.set_ylim()`).
- Add reference lines (axvline, axhline) if the paper has them.
- Use actual series names / legend labels from the paper.
- Include the full caption as the title.
- Set up subplots correctly if the figure has panels.
- NO actual data arrays--- use comments like `# TODO: fill with data`.
- The skeleton should be runnable (producing an empty styled plot) even without data."""


TEMPLATE_GENERATION_USER_PROMPT = """Generate structural templates for each table and figure.
Respond with valid JSON.

Go back to the ORIGINAL TABLE in the paper text and reproduce its structure EXACTLY.

Example regression table template (note how Variable C only appears in columns (3)-(4)):

| | (1) | (2) | (3) | (4) |
|---|---|---|---|---|
| Variable A | XXX | XXX | XXX | XXX |
| | (XXX) | (XXX) | (XXX) | (XXX) |
| Variable B | XXX | XXX | XXX | XXX |
| | (XXX) | (XXX) | (XXX) | (XXX) |
| Variable C | | | XXX | XXX |
| | | | (XXX) | (XXX) |
| Controls | No | Yes | No | Yes |
| Observations | XXX | XXX | XXX | XXX |

Example figure skeleton:

```python
import matplotlib.pyplot as plt
fig, ax = plt.subplots(1, 1, figsize=(10, 6))
# x_obj, density_obj = ... # TODO: compute KDE from data
# ax.plot(x_obj, density_obj, color="tab:blue", linestyle="-", linewidth=2, label="Objective")
ax.set_xlabel("Net gain (EUR/year)", fontsize=12)
ax.set_ylabel("Density", fontsize=12)
ax.set_xlim(-1500, 1500)
ax.axvline(x=0, color="black", linestyle=":", linewidth=0.8)
ax.legend()
plt.tight_layout()
plt.savefig("figure_X.png", dpi=150)
```

## Paper Text (excerpt):
\{paper_text\}

## Extracted Tables:
\{table_specs_json\}

## Extracted Figures:
\{figure_specs_json\}"""


# ---------------------------------------------------------------------------
# Non-replicable figure types (code-level safety net)
# ---------------------------------------------------------------------------

NON_REPLICABLE_KEYWORDS = [
    "flow diagram", "flow chart", "flowchart", "conceptual framework",
    "conceptual diagram", "diagram", "schematic", "screenshot", "photo",
    "photograph", "timeline",
]
\end{alltt}
\end{prompt}
\begin{prompt}[title={C1b. Markdown Methods-Document Extraction Prompt (variant)}, label=prompt:extraction-markdown]
\begin{alltt}\ttfamily\scriptsize
You are a methodological extraction specialist for
social science research papers.

Your task: write a self-contained METHODS DOCUMENT in Markdown describing how the analyses
in this paper were conducted, so that a replicator who cannot see the paper can reproduce
every table. Write it the way a careful, well-organized methods section reads:
prose and ordered steps, not exhaustive specification lists. Extract NO results.

\#\# Output format

Produce a single Markdown document with exactly these sections:

\# Methods Summary

\#\# 0. Research Question
Provide the main research question(s) and the context in 3-5 sentences.

\#\# 1. Data
For each data source a brief overview with unit of observation, time period covered, and how
the raw files relate to each other (merge keys, level of aggregation).

\#\# 2. Data Preparation
An ordered list of the processing steps that take the raw files to the analysis sample(s):
merging of datasets, variable construction, cleaning (missing values, outliers,
winsorization), sample filters, and transformations. Name the variables being constructed
and the conditions being applied. If different analyses use different samples, define each
sample once here and refer to it by name later.

\#\# 3. Analyses
Describe each analysis the paper runs, grouped the way the paper groups them (one
subsection per table or family of related tables). For each: the model type and
estimator, dependent variable, key independent variable(s), control variables, fixed
effects, standard error treatment (clustering level), weights, and estimation sample and
everything else necessary to construct the dataset.
Describe variation across columns compactly ("columns 2 and 4 add controls XY") rather
than one specification per column. Include the estimating equation when the paper states
one explicitly.

\#\# 4. Outputs to Reproduce
A complete inventory of every table in paper order. For each:
- Its identifier and full caption as printed (e.g., "Table 2: The Effect of X on Y").
- 1-3 sentences on its contents: what the rows and columns represent, panel structure,
  and which analysis from Section 3 it reports.

\#\# Level of detail

Target the detail a diligent reader would carry away from a close reading and what is
necessary for a replication without unnecessary details or length. Specifically:
- Resolve cross-references: if the paper defers content to an appendix or another section
  ("controls listed in Table A3", "as defined in Section 2"), fetch that content and state
  it here. The replicator cannot see the paper.
- Name actual variables ("log hourly wage"), not placeholders ("the outcome").
- Control sets: list the variables fully and compactly (one comma-separated line). If the paper
  itself only names a category ("standard demographic controls") and nowhere enumerates
  it, report the category as the paper gives it.
- Do not reproduce exact row/column header text, table notes, or visual styling details;
  the description of contents in Section 4 is sufficient.

\#\# What NOT to include (results)

Never include any numeric value a replicator should compute independently:
- Regression coefficients, standard errors, t-statistics, p-values, significance stars
- Point estimates, confidence intervals, effect sizes, means/medians/SDs of outcomes
- Test statistics (F, chi-squared, Wald) and their p-values
- R-squared, log-likelihood, information criteria
- Calibrated or estimated parameter values (even for structural/calibration tables)
- Counts or percentages that are computed results (e.g., "N=1,234" as a regression output)

Never describe result direction or magnitude ("positive and significant", "large negative
effect", "coefficient of 0.42"). If a label or caption embeds a result, keep only the
structural part.
\end{alltt}
\end{prompt}

\begin{prompt}[title={C2. Reproduction Prompt}, label=prompt:replication]
\begin{alltt}\ttfamily\scriptsize
# Replication Task

You are given a methodological summary of a research paper and its associated
dataset. Your goal is to replicate the paper's empirical results using only
the methodology description below and the data. You do not have access to the
original paper, its results, or any replication code.

## Data

The dataset is located at: `\{data_filename\}`

## Methodological Summary

**Paper**: \{title\} (ID: \{paper_id\})

**Research Questions**:
\{research_questions\}

**Data Description**: \{data_description\}

**Data Context**: \{data_context\}

\{data_source\}\{time_period\}

**Data Processing Steps**:
\{processing_steps\}

\{items_section\}

## Constraints

You are in an isolated workspace for fair benchmarking.

1. **Workspace only.** Only read and write files inside this workspace directory.
   Do not access files outside of it or navigate to parent directories.
   **Exception:** the `data/` directory may be a symlink pointing outside the
   workspace -- this is intentional and you should use it freely as your dataset.

2. **No searching for the paper.** Do not search the internet for this paper,
   its authors, its published results, or any replication code or packages.
   Do not search for prior replication attempts.

3. **No searching for results.** Do not look up expected coefficients, effect
   sizes, tables, or figures from this paper. Your replication must be derived
   entirely from the methodology summary and the data provided.

4. **Allowed web use.** You may search for Python library documentation
   (e.g. statsmodels, pandas, matplotlib) and general statistical methods.

5. **Work independently.** Base your replication only on the methodology
   description in this file and the dataset.

## Instructions

1. **Quick data check**: Inspect the data files to confirm column names and
   basic structure. The methodology summary above describes the variables in
   detail, so a brief check should suffice.

2. **Write `prepare_data.py`**: Load and clean the data following the processing
   steps described above. All scripts can import from this module.

3. **Write and execute one script at a time**: For each item:
   a. Write the script (see output filename specified for each item above)
   b. Execute it
   c. Fix any errors immediately
   d. Move on to the next item once the output file is verified

\{table_instructions\}\{figure_instructions\}

4. **R packages**: If the paper's methodology requires R-specific packages
   (e.g. for specialized estimators), you can call R from Python using `rpy2`.

5. Execute every script and verify the output file exists.

6. Save all outputs in the current working directory.

**Reasonable assumptions.** Where the methodology description is incomplete or
ambiguous, you are free to make reasonable assumptions based on common practice
in the field. Document your assumptions briefly in comments.

Focus on substance and accuracy. Match the described methodology as closely as
possible, including sample restrictions, variable transformations, and
statistical specifications.
\end{alltt}
\end{prompt}
Note that, while Prompt~\ref{prompt:replication} specifies that web use is allowed for researching ``Python library documentation'' and ``general statistical methods,'' the benchmark configuration disallowed web access; compliance was assessed by the guardrail audit.

\begin{prompt}[title={C3. Explanation - Extraction of Divergences Prompt}, label=prompt:explanations-div]
\begin{alltt}\ttfamily\scriptsize
    You are an expert at tracing numerical replication failures back to their root code discrepancies.

Your working directory contains two subfolders:
  original_code/   -- original \{original_language\} replication package (code + data directory)
  agent_code/      -- agent-generated Python replicator

## Target failure
Item: \{item_id\}  |  Grade: \{grade\}  |  \{n_failed\} of \{n_total\} cells failed

\{already_attributed_block\}Failed cells still to explain (sorted by |% difference|):
\{cell_table\}

## Your tasks

1. DATA AVAILABILITY --- Check whether the data files required to produce \{item_id\} exist in
   the data directory inside original_code/. If any required input file is absent from the
   replication package, set data_available = "missing" and describe the missing file(s).
   Otherwise set data_available = "available".

2. AGENT CODE --- In agent_code/*.py, find the code that computes values for the REMAINING
   cells listed above (ignore already-attributed sections).
   If no such code exists, set agent_behavior = "<description of what is missing>".

3. ORIGINAL CODE --- In original_code/\{original_file_glob\}, find the equivalent \{original_language\} code.

4. DISCREPANCIES --- For the REMAINING cells only, identify all distinct root causes.
   Most outputs have one, but multi-panel tables with different specifications may have
   several (e.g. columns 1--3 OLS wrong clustering; columns 4--6 IV wrong instrument).
   If all remaining cells are already explained, return an empty divergences array.
   List one entry per distinct root cause --- do NOT split trivial variations.

5. SECTION MAPPING --- For each discrepancy, list which rows/columns/panels it explains
   (e.g. "All columns", "Columns 1--3 (OLS)", "Panel B"). Maps many cells to one cause.
   Also list each specific affected cell from the cell table above as
   \{\{"item_id": "\{item_id\}", "row_label": "...", "column_label": "..."\}\} using the exact
   labels shown. Include all cells you believe this discrepancy explains.

6. ALSO EXPLAINS --- For each discrepancy, list which other failures it also explains.
   Each entry is either:
     - A plain string "<item_id>" if the discrepancy explains the entire other item, OR
     - An object \{\{"item_id": "<item_id>", "sections": "<which part>"\}\} if only partial.
   Be conservative: only include items you are confident share the exact same root cause.

## Other unresolved failures
\{remaining_block\}

## Divergence taxonomy (use exactly one code)
  S1  Wrong model specification      -- incorrect FE, clustering level, or SE type
  S2  Wrong estimator / inference    -- wrong estimator (OLS vs IV) or missing inference step
  S3  Data source substitution       -- proxy used / required dataset absent from package
  S4  Wrong sample restriction       -- filter missing, wrong condition, or wrong subset
  S5  Wrong variable construction    -- outcome/predictor coded differently than reference
  S6  Missing analysis component     -- required step entirely omitted
  S8  Wrong merge / transform logic  -- wrong join type, key, duplicate handling, or reshape
  S9  Wrong sequencing               -- steps in wrong order, changing results
  S0  Other                          -- does not fit any category above

## Severity
  minor    -- unlikely to materially affect point estimates or conclusions
  medium   -- could shift estimates noticeably; sign/significance probably stable
  critical -- likely changes sign, significance, or core conclusion of a main result

## Output
Write a JSON object with a "divergences" array to \{output_path\}.
One entry per distinct root cause found in \{item_id\} (usually 1, occasionally 2--3).
Write ONLY valid JSON --- no markdown fences, no preamble, no comments.
Escape all special characters properly inside string values.

\{\{
  "divergences": [
    \{\{
      "id": \{divergence_id\},   (use sequential integers starting here for additional entries)
      "output": "\{item_id\}",
      "description": "<one sentence: what is different>",
      "original_behavior": "<what the original \{original_language\} code does on this specific point>",
      "agent_behavior": "<what Python does, or the ABSENT description>",
      "original_proof": "<short verbatim \{original_proof_label\}>",
      "agent_proof": "<short verbatim Python snippet, or ABSENT>",
      "original_location": \{\{"file": "<\{original_file_ext\}>", "line": "<line or range>"\}\},
      "agent_location": \{\{"file": "<filename.py or ABSENT>", "line": "<line or ABSENT>"\}\},
      "divergence_type": "<S0|S1|S2|S3|S4|S5|S6|S8|S9>",
      "severity": "<minor|medium|critical>",
      "data_available": "<available|missing>",
      "data_available_note": "<one sentence naming specific file(s) checked or found absent>",
      "explains_sections": ["<e.g. 'All columns', 'Columns 1-3 (OLS)', 'Panel B'>"],
      "affected_cells": [
        \{\{"item_id": "\{item_id\}", "row_label": "<exact row label from the cell table>", "column_label": "<exact column label>"\}\}
      ],
      "also_explains": ["<item_id>", \{\{"item_id": "<item_id>", "sections": "<which part>"\}\}]
    \}\}
  ]
\}\}
\end{alltt}
\end{prompt}
\begin{prompt}[title={C4.1. Explanation - Error source Original Paper vs Code}, label=prompt:explainer-paper-code]
\begin{alltt}\ttfamily\scriptsize

You are verifying whether the paper supports the original \{original_language\} code's behavior
for a set of replication divergences.

Your working directory contains:
  paper.pdf              --- the published paper
  original_code_files/   --- the original \{original_language\} replication code (for reference only)

For each divergence you are given:
  - `original_behavior`: what the \{original_language\} code does for this analysis step
  - `original_proof`: the exact \{original_language\} code snippet implementing this behavior
  - `original_location`: the file and line number in the original code

The original code is already provided --- you do NOT need to re-read the code files.
Read paper.pdf to determine whether it explicitly states, implies, or is silent
about the behavior described in `original_behavior`.  Classify using exactly one
of these four verdicts:

  consistent   = paper and original code explicitly agree on this specific point
  contradicts  = direct contradiction: the paper explicitly says X AND the original
                 code explicitly does Y, where X \ensuremath{\neq} Y
  omission     = the original code implements X, but the paper does not mention X at
                 all --- no contradiction, the paper simply doesn't cover this detail

Rules:
- "contradicts" requires both documents to explicitly address the same point with
  different answers.  The paper being silent on an implementation detail is
  NEVER sufficient for "contradicts" --- use "omission" instead.
- "omission" means the original code (the upstream document here) specifies something
  the paper (the downstream document) doesn't address.  Example: code computes
  monthly standardized values and averages them; paper describes the monthly formula
  but says nothing about aggregation \ensuremath{\rightarrow} omission.
- "consistent" if both agree; prefer "omission" when one side is explicit and the other is silent.
- Your note must cite specific evidence: variable name, section heading, line
  number, or direct quote.
- You MUST produce exactly one of these three verdicts for every divergence id. Do not abstain or invent new labels.

EFFICIENCY: Do NOT read the entire paper exhaustively.  Use the divergence
description to identify the relevant section, check that section, form a
well-founded hypothesis, cite the evidence, and move on.

\end{alltt}
\end{prompt}
\begin{prompt}[title={C4.2. Explanation - Error source Original Paper vs Extractor}, label=prompt:explainer-paper-extractor]
\begin{alltt}\ttfamily\scriptsize
You are verifying whether the methodology summary accurately represents what
the paper says, for a set of replication divergences.

Your working directory contains:
  paper.pdf                  --- the published paper
  methodology_summary.json   --- the summary passed to the replicator agent

For each divergence you are given:
  - `original_behavior`: what the original code does for this analysis step
  - `original_proof`: the exact code snippet (for context on what to look for)
  - `original_location`: the file and line number in the original code

The original code is already provided --- you do NOT need to read the code files.
Read paper.pdf and methodology_summary.json to determine whether the paper states
this behavior and whether the summary faithfully represents it.  Classify using
exactly one of these four verdicts:

  consistent   = summary faithfully represents the paper on this point, OR both
                 are silent (summary correctly captured the paper's silence)
  contradicts  = direct contradiction: paper explicitly says X AND summary
                 explicitly says Y \ensuremath{\neq} X
  omission     = paper explicitly says X, but the summary does not mention X at
                 all --- the summary dropped information the paper provided

Rules:
- "omission" means the paper (upstream) says something the summary (downstream)
  dropped.  This is the most common failure mode for this check.
- "contradicts" requires both documents to explicitly address the same point
  differently --- not just one being silent where the other speaks.
- If neither the paper nor the summary addresses the point, mark "consistent"
  (they agree by silence --- the break is not here).
- The summary is NOT expected to be more detailed than the paper.  If the paper
  states X at a conceptual level (e.g. "cluster at neighborhood level") and the
  summary conveys the same concept (e.g. "use neighborhood cluster identifiers"),
  mark "consistent" --- even if the summary does not reproduce the exact variable
  names or implementation details that appear in the original code but not in the
  paper.  The standard is semantic faithfulness to what the paper says, not
  completeness of implementation detail.
- To distinguish "consistent" from "omission" at the conceptual level, ask:
  "Would a careful agent reading only the summary know that they should implement X
  (the concept the paper describes)?"  If yes \ensuremath{\rightarrow} "consistent" (the concept was
  conveyed, even if less specific).  If no, because the summary says nothing about
  X at all \ensuremath{\rightarrow} "omission".  Note: "omission" does not require that the agent would
  actively do the wrong thing; it is sufficient that the summary dropped a concept
  the paper stated, leaving the agent without guidance on that point.
- Your note must cite specific evidence from both documents.
- You MUST produce exactly one of these three verdicts for every divergence id. Do not abstain or invent new labels.

EFFICIENCY: Do NOT read documents exhaustively.  Use the divergence description
to locate the relevant section in each document, check it, form a well-founded
hypothesis, cite the evidence, and move on.
\end{alltt}
\end{prompt}
\begin{prompt}[title={C4.3. Explanation - Error source Extractor vs Agent}, label=prompt:explainer-extractor-agent]
\begin{alltt}\ttfamily\scriptsize
You are verifying whether the agent's Python code implements what the
methodology summary instructs, for a set of replication divergences.

Your working directory contains:
  methodology_summary.json   --- the summary passed to the replicator agent
  agent_code/                --- the agent's Python replication code (for reference only)

For each divergence you are given:
  - `agent_behavior`: what the Python code actually does for this analysis step
  - `agent_proof`: the exact Python code snippet implementing this behavior
  - `agent_location`: the file and line number in agent_code/

The Python code is already provided --- you do NOT need to re-read agent_code/.
Read methodology_summary.json to determine whether it explicitly instructs,
implies, or is silent about the behavior described in `agent_behavior`.  Classify
using exactly one of these four verdicts:

  consistent   = agent follows the summary on this point, OR both are silent
                 (agent correctly followed the summary's omission)
  contradicts  = direct contradiction: summary explicitly says X AND agent code
                 explicitly does Y \ensuremath{\neq} X
  omission     = summary explicitly instructs X, but the agent does not implement
                 X at all --- agent omitted something the summary described

Rules:
- Each divergence also includes a `description` field explaining what the step
  is about.  Use this to find the relevant section in methodology_summary.json
  when agent_behavior is "ABSENT" or otherwise unclear.
- "omission" covers agent_behavior = "ABSENT": use `description` to identify
  what analysis step was supposed to be implemented, look it up in the summary,
  and if the summary describes it \ensuremath{\rightarrow} omission.  Only mark "consistent" if the
  summary is also silent on that step.
- "contradicts" requires the summary to explicitly say X AND the agent to do
  something explicitly different --- not just the agent being silent.
- IMPORTANT: if the summary does NOT mention this specific detail at all
  (e.g. the summary is silent about weighting, file choice, variable variant,
  clustering method, or any other implementation detail), mark "consistent" ---
  the agent cannot be expected to follow guidance that was not provided.
  The agent making its own reasonable choice on an unspecified detail is NOT
  a contradiction or omission.
- Your note must cite specific evidence from both the summary and the Python code.
- You MUST produce exactly one of these three verdicts for every divergence id. Do not abstain or invent new labels.

EFFICIENCY: Work from the provided code snippets and the summary text.  Do NOT
exhaustively search through all files.  Form a well-founded hypothesis based on
the evidence given, cite it, and move on.
\end{alltt}
\end{prompt}
\begin{prompt}[title={C4.4. Explanation - Error source Data Missing}, label=prompt:explainer-data]
\begin{alltt}\ttfamily\scriptsize
You are checking whether the data required by the original \{original_language\} code is
available to a Python replicator, for a set of divergences.

Your working directory is the replication data directory.  The original
\{original_code_description\} are in \{code_files_path\}.

## Step 1 --- identify what the original code loads

For each divergence, read the original code files and find the data file(s)
loaded for that analysis step.

## Step 2 --- distinguish raw inputs from code-constructed intermediates

Before checking whether a file exists, determine whether it is:

  (a) A RAW SOURCE FILE --- brought in from outside (survey data, official
      statistics, downloaded datasets).  These must exist as-is for any
      replicator.

  (b) A CODE-CONSTRUCTED INTERMEDIATE --- a file that is itself created by
      the original code in the same replication package.  A Python
      replicator would construct this from its own upstream steps, not load a
      pre-built file.

You can identify constructed intermediates by searching the original code for
save/export commands that write that filename.

## Step 3 --- check availability

  • If the required file is a RAW SOURCE FILE: check whether it exists in the
    current directory.  If yes \ensuremath{\rightarrow} available; if no \ensuremath{\rightarrow} missing.

  • If the required file is a CODE-CONSTRUCTED INTERMEDIATE: instead check
    whether the raw source file(s) that feed into its construction are present.
    If those raw sources exist \ensuremath{\rightarrow} available (the agent can construct the
    intermediate from them); if the raw sources are also absent \ensuremath{\rightarrow} missing.

  available = the required data (raw or constructable from present sources)
              was accessible; the agent had what it needed and chose incorrectly
  missing   = the required raw data is absent; the agent could not have
              implemented this correctly regardless of effort

Every divergence must receive `available` or `missing` --- no other values.
In your note, name the file(s) you checked and state whether each is a raw
source or a constructed intermediate.
\textbackslash{}end\{lstlisting\}
\end{alltt}
\end{prompt}

\end{document}